\documentclass[final,5p,times,twocolumn]{elsarticle}

\usepackage{graphicx}
\usepackage{url}
\usepackage{subcaption}
\usepackage{float}

\usepackage{booktabs}
\usepackage{multirow}
\usepackage{array}
\usepackage{tabularx}
\usepackage{makecell}
\usepackage{adjustbox}
\usepackage{rotating}
\usepackage{longtable}

\usepackage{amsmath}
\usepackage{amssymb}

\usepackage{caption}
\usepackage{setspace}

\usepackage{placeins}

\usepackage{xcolor}

\usepackage{booktabs}
\usepackage{pdflscape}   
\usepackage{ltablex}
\usepackage{graphicx}
\usepackage{subcaption}
\usepackage{caption}
\usepackage{makecell}
\keepXColumns
\usepackage{array}
\usepackage{ragged2e}
\newcolumntype{Y}{>{\RaggedRight\arraybackslash}X}
\usepackage{booktabs}  
\usepackage{tabularx}
\usepackage{mathrsfs}%
\usepackage{array}
\usepackage{pdflscape} 
\usepackage{newunicodechar}
\newunicodechar{Δ}{$\Delta$}
\usepackage[utf8]{inputenc}
\usepackage{longtable}  
\usepackage{array}
\usepackage{graphicx}%
\usepackage{multirow}%
\usepackage{amsmath,amssymb,amsfonts}%
\usepackage{amsthm}%
\usepackage{mathrsfs}%
\usepackage{xcolor}%
\usepackage{textcomp}%
\usepackage{manyfoot}%
\usepackage{booktabs}%
\usepackage{algorithm}%
\usepackage{algorithmicx}%
\usepackage{algpseudocode}%
\usepackage{listings}%
\usepackage{comment}

\usepackage{amssymb}
\usepackage{amsmath}

\journal{Medical Image Analysis}

\begin{document}

\begin{frontmatter}



\title{MS-DKC: A Dataset Knowledge Card Framework for Designing and Adapting Medical Image Segmentation Models} 


\author[aff1,aff7]{Tariq M. Khan\corref{cor1}}
\ead{tariq045@gmail.com}
\author[aff2]{Syed Saud Naqvi}
\ead{saud_naqvi@comsats.edu.pk}
\author[aff1]{Thantrira Porntaveetus}
\ead{Thantrira.P@chula.ac.th}
\author[aff3,aff4]{Hamid Alinejad-Rokny}
\ead{h.alinejad@unsw.edu.au}
\author[aff5]{Shahzaib Iqbal}
\ead{shahzaib.iqbal91@gmail.com}
\author[aff6]{Imran Razzak}
\ead{imran.razzak@mbzuai.ac.ae}
\author[aff2]{Mohammad AU Khan}
\ead{masmatkhan@psu.edu.sa}

\cortext[cor1]{Corresponding author.}
\address[aff1]{Center of Excellence in Precision Medicine and Digital Health, Faculty of Dentistry, Chulalongkorn University, Bangkok, Thailand}
\address[aff2] {Department of Computer Engineering, COMSATS University Islamabad, Islamabad, Pakistan}
\address[aff3]{School of Biomedical Engineering, UNSW, Sydney, NSW, Australia}
\address[aff4]{Visiting Scholar (Collaborative Projects), Center of Excellence in Precision Medicine and Digital Health, Chulalongkorn University, Bangkok, Thailand}
\address[aff5]{Department of Computing, Abasyn University Islamabad Campus (AUIC), Islamabad, Pakistan}
\address[aff6]{Mohamed bin Zayed University of Artificial Intelligence, Abu Dhabi, United Arab Emirates}
\address[aff7]{College of Computer and Information Sciences, prince Sultan University, Riyadh, SAudi Arabia}

\begin{abstract}
Medical image segmentation is often framed as a search for stronger architectures, but this can obscure a more fundamental question: what does the dataset require from the model? In medical imaging, this requirement is shaped by foreground occupancy, morphology, boundary ambiguity, topology sensitivity, annotation quality, acquisition variation, and operating point.

This paper introduces the Medical Segmentation Dataset Knowledge Card (MS-DKC), a framework for making these factors explicit. MS-DKC records dataset evidence through image/acquisition, morphology, supervision, context-dependence, and deployment-risk descriptors. These descriptors are mapped to failure modes, design priors, and risk-aligned criteria, making segmentation design more traceable than architecture-first comparison.

We evaluate MS-DKC on DRIVE, ISIC2018, and ACDC, representing distinct regimes. DRIVE contains sparse, thin, branching vessels, favoring detail-preserving models, sensitivity-aware optimization, threshold analysis, and topology-aware metrics. DKC-TNet-v2 achieved Dice 0.8044 and IoU 0.6730 with 35103 parameters, while SA-UNetv2-DKC-AmbRef reached Dice 0.8141, IoU 0.6865, sensitivity 0.8265, specificity 0.9804, and AUC 0.9853. ISIC2018 involves compact but appearance-variable lesions; validation-constrained score-function selection on Att-Next-Topo/ATTNext produced MS-DKC-AttNextTopo-VCSF-NoAug with Dice 0.8872, IoU 0.8214, precision 0.9173, Boundary F1 0.4878, and ASSD 4.13, while plausible additions failed to improve the risk-aligned profile. ACDC provides a multi-class cardiac case, where MS-DKC recommends four-class softmax segmentation, class-balanced Dice/CE supervision, and class-wise surface evaluation. Overall, the results support dataset-conditioned design: different datasets require different priors, operating points, and evidence before a model can be judged appropriate.
\end{abstract}



\begin{keyword}



\end{keyword}

\end{frontmatter}


\section{Introduction}

Medical image segmentation is a central task in medical image analysis. It supports organ delineation, lesion quantification, treatment planning, radiotherapy support, image-guided intervention, and longitudinal disease monitoring. Deep learning made this task practical on a scale, and the U-Net architecture remains one of the most influential examples. Its contracting path captures the semantic context, while the expanding path and skip connections recover spatial detail from limited annotated data \cite{ronneberger2015unet}. The same encoder-decoder idea was later extended to volumetric learning through V-Net and 3D U-Net, where dense voxel-wise prediction is required in three-dimensional scans \cite{milletari2016vnet,cicek20163dunet}.

Much of the progress that followed refined this template. Attention U-Net used attention gates to suppress irrelevant skip information and highlight task-relevant regions \cite{oktay2018attentionunet}. UNet++ redesigned the skip pathways with nested and dense connections to reduce the semantic gap between the encoder and decoder features \cite{zhou2020unetplusplus}. nnU-Net then showed that the careful configuration of the specific dataset of preprocessing, training, architecture settings and post-processing can be as important as the choice of the backbone itself \cite{isensee2021nnunet}. More recent work has further widened the design space. UNETR, Swin-Unet, nnFormer, TransBTS, TransFuse, and TransUNet use Transformer or hybrid CNN-Transformer components to capture broader spatial context \cite{hatamizadeh2022unetr,cao2022swinunet,wang2022nnformer,wang2021transbts,zhang2021transfuse,chen2024transunet}. State-space models such as SegMamba and related Mamba-based methods provide another route to long-range modeling \cite{xing2024segmamba,dang2024logvmamba}. Foundation-model approaches, including MedSAM and MedSAM2, have also brought prompt and transferable segmentation priors to medical imaging \cite{ma2024medsam,zhu2024medsam2}.

Related medical image analysis and segmentation studies also show that lightweight, multi-kernel, complexity-aware, self-supervised, fusion-based, and edge-oriented designs remain highly relevant when model choices must be aligned with data characteristics and deployment constraints \cite{khan2022leveraging,khan2022mkis,iqbal2023fusion,iqbal2023ldmres,qayyum2023two,javed2024advancing,iqbal2025tbconvl,khan2024esdmr,xu2025edge,safdar2025focal,khan2026edge}.

These developments have produced a rich architectural toolbox, but they also reveal a limitation in how medical segmentation studies are often framed. New methods are commonly introduced as stronger backbones, larger context models, more expressive attention blocks, or more transferable pretrained systems. This architecture-first framing is useful, but it does not fully explain why a particular design is suitable for a particular dataset. A retinal vessel dataset, a dermoscopic lesion dataset, and a cardiac MRI dataset may all be binary or low-cardinality segmentation problems, yet they do not have the same risks. Thin branching vessels are sensitive to missed terminal structures and connectivity breaks. Skin lesions are compact regions, but their borders can be fuzzy, low-contrast, and strongly affected by the variation of the appearance. The cardiac structures are larger and more spatially coherent, with different failure modes again. A model that is well matched to one of these regimes may be unnecessary, insufficient, or even poorly calibrated for another.

This distinction matters because many medical datasets are not difficult in the same way as natural-image recognition datasets. Natural-image difficulty is often associated with broad semantic diversity, large category vocabularies, and large-scale annotated corpora. In medical segmentation, the difficulty is often tied to foreground occupancy, target morphology, topology sensitivity, downsampling risk, anisotropy, annotation uncertainty, acquisition variation, and clinical operating point. These properties affect practical design choices: how deep the encoder should downsample, how skip features should be fused, whether boundary or topology losses are justified, whether a compact model is preferable to a large one, which threshold should be selected, and which metrics should be reported.

Existing work already points toward this problem. The nnU-Net demonstrated that the configuration of the data set can be more important than the architectural novelty in many settings \cite{isensee2021nnunet}. Hybrid CNN-Transformer and state-space models are often motivated by the need for a broader context when purely local convolution is not enough \cite{hatamizadeh2022unetr,cao2022swinunet,wang2021transbts,xing2024segmamba}. Foundation-model studies show that transferable priors can be useful, but still require domain adaptation, appropriate prompting, calibration, and careful evaluation \cite{ma2024medsam,zhu2024medsam2,efficientmedsams2024}. What is still missing is a general design record that begins with the dataset, states the evidence, identifies the expected risks, and then justifies the modeling and evaluation choices.

In this paper, we formulate medical image segmentation as a \emph{dataset-conditioned design and adaptation problem}. We introduce the \emph{Medical Segmentation Dataset Knowledge Card} (MS-DKC), a structured framework that records dataset descriptors, descriptor confidence, anticipated risks, ranked design priors, and risk-aligned evaluation criteria. MS-DKC is not proposed as a single architecture. Instead, it is a way to make the reasoning behind architecture selection, model capacity, loss design, thresholding, refinement, and metric choice explicit and auditable.

This distinction also defines the scope of the experiments. The paper is not intended to be an unrestricted search for a new state-of-the-art architecture on every dataset. Instead, it asks whether dataset knowledge can help us decide when an existing model should be accepted, adapted, or rejected. Therefore, the experiments are organized as models and decision-analysis studies. In DRIVE, MS-DKC is used more extensively because DRIVE has the strongest risk profile in this work: only 20 training images and 20 test images, sparse vessels, thin terminal branches, and topology-sensitive errors. In ISIC and ACDC, the role is supporting and diagnostic: these datasets show that a different target morphology leads to different design priorities and that the same intervention should not be expected to work everywhere.

We evaluate this idea in three representative segmentation regimes. DRIVE is used as the primary thin-structure case. Its sparse, narrow branching vessels motivate a detail-preserving design, an imbalance-sensitive optimization, sensitivity-aware threshold selection, and a topology-aware evaluation. In this profile, MS-DKC is instantiated both as a compact model, DKC-TNet-v2, and as an adaptation pathway for the vessel-specific SA-UNetv2 backbone. ISIC2018 is used as a compact-region lesion case. Its profile is different: lesions are usually single connected regions, but show a large variation in size, color, texture, and boundary appearance. The revised ISIC branch shows that the strongest current intervention is not a larger module stack, but validation-constrained operating-point selection on a stable AttNextTopo backbone, supported by boundary- and size-stratified evaluation. ACDC is retained as a multi-class cardiac-anatomy case, where the MS-DKC profile emphasizes shape consistency, myocardium boundary sensitivity, background dominance, and class-wise surface evaluation.

The new ISIC results are important for the argument. Att-Next-Topo/ATTNext is a strong skin-lesion backbone on this split, and the no-augmentation diagnostic branch shows that validation-constrained score-function selection can improve the overlap-boundary trade-off without adding architectural capacity. The retained MS-DKC-AttNextTopo-VCSF-NoAug candidate improves Dice, IoU, precision, boundary F1 and ASSD relative to the no-augmentation baseline, while several plausible architecture additions fail to improve the full risk-aligned profile. This is exactly the point of MS-DKC: a dataset-aware workflow should identify which intervention is justified by the measured risk profile and which attractive interventions should be rejected.

Table~\ref{tab:intro_positioning} summarizes how the proposed framework differs from neighboring lines of work. Unlike architecture surveys, MS-DKC does not primarily organize models by family. Unlike self-configuring pipelines, it records an explicit design rationale rather than only producing a configuration. Unlike foundation-model studies, it treats transferable priors as one possible intervention within a dataset-specific design space.

\begin{table*}[htbp!]
\centering
\caption{Positioning of the proposed dataset-centered framework relative to neighboring lines of work in medical image segmentation.}
\label{tab:intro_positioning}
\begin{tabular}{p{2.2cm}p{4.0cm}p{4.0cm}p{6.0cm}}
\toprule
\textbf{Category} & \textbf{Primary focus} & \textbf{Typical output} & \textbf{Relation to dataset-centered design} \\
\midrule

Architecture surveys and model reviews
& Organize the field by architecture family, modality, or training paradigm
& Taxonomies of CNNs, Transformers, hybrid models, state-space models, or foundation-model adaptations
& Provide useful methodological context, but offer limited guidance about which measurable dataset properties should justify specific modeling, optimization, and evaluation choices \\

nnU-Net and self-configuring pipelines
& Dataset-specific adaptation through empirical fingerprints and automated configuration
& Strong standardized segmentation pipelines with task-aware preprocessing, training, and post-processing
& Demonstrate the value of dataset-specific configuration, but do not provide a general design record linking dataset characterization to anticipated risks, ranked interventions, and evaluation rationale \\

Generalist and foundation-model segmentation studies
& Transferability, prompting, large-scale priors, and cross-task coverage
& Broad segmentation backbones, promptable models, and comparative performance analyses
& Highlight the promise of transferable priors, but give limited account of when particular model assumptions, capacity regimes, or evaluation strategies are justified by a specific dataset profile \\

\textbf{This paper}
& Dataset-conditioned design and adaptation reasoning
& A framework that links dataset descriptors to anticipated risks, design priors, empirical interventions, and risk-aligned evaluation
& Provides an explicit and traceable basis for designing compact models and adapting existing backbones according to measurable dataset structure rather than architecture-driven defaults \\

\bottomrule
\end{tabular}
\end{table*}

The main contributions of this paper are as follows:
\begin{itemize}
    \item We formulate medical image segmentation as a \emph{dataset-conditioned design and adaptation problem}, where measurable dataset properties guide choices about architecture family, model capacity, downsampling, supervision, loss design, threshold selection, post-processing, and evaluation.

    \item We introduce the \emph{Medical Segmentation Dataset Knowledge Card} (MS-DKC), a structured framework that represents segmentation datasets through image/acquisition, target morphology, supervision, context-dependence, and deployment-risk descriptors.

    \item We define a \emph{measurement--risk--intervention} workflow that links dataset descriptors and descriptor confidence to anticipated failure modes, ranked design priors, empirical interventions, and risk-aligned evaluation.

    \item We instantiate MS-DKC on DRIVE by designing a compact T-Net-inspired DKC-TNet-v2 model and by adapting the existing SA-UNetv2 retinal vessel backbone through Dice/MCC/BCE loss calibration, validation-selected thresholding, and ambiguous-pixel refinement.

    \item We show that MS-DKC-guided reasoning improves DRIVE vessel segmentation without relying on model size alone. DKC-TNet-v2 provides a strong accuracy--efficiency trade-off with only 35,103 parameters, while SA-UNetv2-DKC-AmbRef achieves the strongest DRIVE result in our controlled experiments.

    \item We extend the experimental analysis to ISIC2018 and ACDC. On ISIC2018, the Att-Next-Topo/ATTNext backbone is already highly competitive, and the new MS-DKC-AttNextTopo-VCSF-NoAug diagnostic candidate improves the overlap-boundary trade-off without increasing the model size. The ISIC ablations further show that DiceBoost, FES boundary boosting, focal modulation, and PolySF are not automatically beneficial. In ACDC, the task is a multi-class cardiac-anatomy problem with shape constraints, small foreground occupancy, and high myocardium boundary sensitivity. These findings support the broader claim that the preferred model, threshold, and metric panel should follow the profile of the dataset.
\end{itemize}

The central message is therefore simple: medical segmentation should not begin with the assumption that one architecture family is universally best. It should begin with the measured structure of the dataset, the risks implied by that structure, and the evaluation criteria needed to verify that the chosen design has addressed those risks.

The remainder of the paper is organized as follows. We first explain why natural-image design defaults transfer imperfectly to medical segmentation. We then define MS-DKC, describe its measurement layers, and show how those measurements are translated into risks and design priors. Next, we revisit U-Net-like design, model capacity, transfer learning, and overlap-centric evaluation from a dataset-conditioned viewpoint. We then present the DRIVE, ISIC2018, and ACDC experiments before discussing limitations and future work.

\section{Why Natural-Image Design Defaults Transfer Imperfectly}

Many influential design choices in medical image segmentation were inherited from the broader computer vision literature. Encoder-decoder architectures, residual backbones, multiscale feature hierarchies, large-scale pretraining, and more recent transformer-based designs were first developed largely in natural-image settings before being adapted to biomedical images. In many cases, this transfer has been highly beneficial: borrowed priors have improved optimization, accelerated convergence, and provided strong baselines under limited supervision \cite{ronneberger2015unet,milletari2016vnet,cicek20163dunet,isensee2021nnunet,hatamizadeh2022unetr}. However, their success should not obscure a central point: many medical segmentation tasks differ materially from the natural-image problems that originally motivated these defaults.

The first mismatch relates to \emph{task semantics}. Natural-image recognition is generally driven by high semantic variety and extensive vocabularies of categories, whereas many medical segmentation problems utilize either one foreground structure or a limited selection of tissue or pathologic areas. In such scenarios, the problem is often less from coarse-grained class separation but from fine-grained boundaries underclass imbalance, boundary ambiguity, topological sensitivity, or restricted supervision. As a consequence, encoder width, stage depth, and semantic priors learned from large-scale natural-image models are not inherently aligned with the main difficulty of the segmentation task \cite{chen2024transunet,hatamizadeh2022unetr,cao2022swinunet,wang2022nnformer,pei2022vtunet}.

The second mismatch is about \emph{the annotation regime}. Medical segmentation usually suffers from small datasets, costly annotation at the pixel or voxel level, and non-trivial inter-rater variability. Performance in such cases may rely as much on the dataset-specific setup and technical discipline as on the notional complexity of the backbone. Standardization, preprocessing, and task-specific flexibility may be as essential as architectural innovation, as shown in systems such as nnU-Net, No New-Net, and NiftyNet \cite{isensee2021nnunet,isensee2018nonewnet,gibson2018niftynet}.

The third mismatch involves \emph{structure-sensitive difficulty}.  A significant number of clinically relevant targets are narrow, sparse, branching, low contrast, topologically constrained, or intensely anisotropic. Examples include nuclei boundaries, tumor margins, membranes, arteries, ducts, and tiny lesions. In such tasks, repeated downsampling can eliminate signals that are necessary for clinically meaningful delineation, and errors that appear minor under overlap measures can still disrupt connectivity or distort topology. Methods that emphasize spatial fidelity, boundary preservation, and topology-aware supervision therefore show that architectural defaults from generic dense prediction may need to be re-examined when the target structure is sparse, thin, boundary-critical, or topology-sensitive \cite{zhou2020unetplusplus,huang2020unet3plus,valanarasu2020kiunet,kervadec2019boundary,shit2021cldice}.

The fourth mismatch concerns \emph{modality-specific image formation}. CT, MRI, ultrasound, OCT, microscopy, fundus imaging, and histopathology have distinct contrast mechanisms, noise characteristics, artifact profiles, and spatial organization. As a result, the transferred visual priors may only partially match the imaging evidence that is important for the segmentation task. The notion of useful context also varies by modality and task: some datasets require local boundary precision, others require volumetric continuity or broad anatomical context, and still others require robustness to scanner- or stain-induced variation \cite{milletari2016vnet,cicek20163dunet,myronenko2018automated,wang2021transbts,zhang2021transfuse,hatamizadeh2022swinunetr}.

The fifth mismatch involves \emph{dimensionality and anisotropy}. Many medical segmentation tasks are fundamentally volumetric, yet volumes are often anisotropic, incompletely sampled, or too large for simple full-resolution processing. As a consequence, the choice between 2D, 2.5D, anisotropic 3D, or isotropic 3D modeling is a dataset-dependent design decision rather than a routine implementation detail. More recent transformer- and state-space-based approaches can be interpreted partly as responses to these broader context- and dependency challenges. However, the deeper question is whether the chosen context mechanism is appropriate for the spatial structure of the dataset \cite{hatamizadeh2022unetr,wang2022nnformer,pei2022vtunet,ma2024umamba,xing2024segmamba,dang2024logvmamba}.

The final mismatch concerns \emph{evaluation assumptions}. Medical segmentation is often summarized using region-overlap metrics such as Dice or IoU, even when clinically significant failures are boundary-sensitive, topology-sensitive, uncertainty-sensitive, or subgroup-specific. The increasing use of boundary- and topology-based criteria shows that the assumptions inherited from generic dense prediction do not transfer cleanly to all medical scenarios \cite{kervadec2019boundary,shit2021cldice,wasserthal2023totalsegmentator}.

Taken together, these mismatches do not imply that natural-image defaults should be discarded. Rather, they often remain strong baselines. The problem is that they are too often treated as neutral starting points rather than as hypotheses whose suitability depends on the measurable properties of the dataset. This motivates a framework in which dataset properties are characterized first and architectural, supervisory, and evaluation choices are treated as responses to the risks implied by those measurements. In the next section, we formalize this view through the Medical Segmentation Dataset Knowledge Card (MS-DKC).

\section{Medical Image Segmentation as a Dataset-Conditioned Design Problem}

The preceding discussion suggests that many central design decisions in medical image segmentation cannot be justified by architectural popularity alone. If segmentation datasets differ systematically in acquisition conditions, target morphology, supervision quality, context dependence, and deployment exposure, then model design should begin with an explicit characterization of those factors rather than with the uncritical adoption of an inherited backbone. We therefore treat medical image segmentation as a \emph{dataset-conditioned design problem}.

In this view, the primary object of analysis is not the architecture in isolation but the dataset together with the risks it induces. The model design then becomes a response to those risks. This implies a shift from an \emph{architecture-first} workflow to a \emph{dataset-first} workflow. In an architecture-first workflow, a practitioner chooses a familiar model family, such as a standard U-Net, a transformer-enhanced encoder-decoder, or a foundation-model adaptation, and then adapts preprocessing, losses, and training details around that initial decision \cite{ronneberger2015unet,isensee2021nnunet,hatamizadeh2022unetr,ma2024medsam}. In a dataset-first workflow, the process begins by asking which properties of the dataset are measurable, which risks these properties imply, and which design choices are therefore justified as starting hypotheses.

To operationalize this perspective, we introduce a \emph{Medical Segmentation Dataset Knowledge Card} (MS-DKC). The MS-DKC is a structured and auditable record that links dataset characterization to design reasoning. Formally, for a segmentation dataset $D$, we define
\begin{equation}
\mathrm{MS\mbox{-}DKC}(D)=
\bigl\langle
\mathcal{M}(D),\mathcal{Q}(D),\mathcal{R}(D),\Pi(D),\mathcal{E}(D)
\bigr\rangle,
\label{eq:msdkc_main}
\end{equation}
where:
\begin{itemize}
    \item $\mathcal{M}(D)$ is the set of measured dataset descriptors,
    \item $\mathcal{Q}(D)$ is the associated descriptor-quality record,
    \item $\mathcal{R}(D)$ is the severity-scored risk profile inferred from the descriptor profile,
    \item $\Pi(D)$ is the set of ranked design priors, and
    \item $\mathcal{E}(D)$ is the risk-aligned evaluation suite.
\end{itemize}

This definition makes explicit that MS-DKC is not merely a summary of dataset metadata. It is a design-oriented artifact whose purpose is to record not only \emph{what} has been measured but also \emph{how confidently} it is known, \emph{which failure modes} it suggests, \emph{which design interventions} should be considered, and \emph{which evaluation choices} are needed to verify the relevant risks. In this formulation, $\Pi(D)$ contains design priors only, while the evaluation choices are kept in the separate component $\mathcal{E}(D)$ so that design recommendations and evaluation criteria do not overlap.

The core workflow of the framework is
\begin{equation}
\begin{aligned}
\text{dataset descriptors}
&\rightarrow \text{anticipated risks} \\
&\rightarrow \text{ranked design priors} \\
&\rightarrow \text{risk-aligned evaluation}.
\end{aligned}
\end{equation}
We express the risk inference stage as
\begin{equation}
\mathcal{R}(D)=\Psi\bigl(\mathcal{M}(D),\mathcal{Q}(D)\bigr)
=
\bigl\{(r,s_r): r\in\mathfrak{R},\ s_r\in[0,1]\bigr\},
\label{eq:risk_inference}
\end{equation}
where each risk $r$ has a severity score $s_r$, and $\Psi(\cdot)$ maps the measured descriptors and their confidence levels to plausible failure mechanisms such as loss of structural information, optimization bias, capacity mismatch, supervision mismatch, domain fragility, or evaluation mismatch. The resulting risk profile is then mapped to a ranked set of design priors,
\begin{equation}
\begin{aligned}
\Pi(D)&=\Gamma\bigl(\mathcal{R}(D),\mathcal{Q}(D)\bigr)
=\bigl(\pi_1(D),\ldots,\pi_K(D)\bigr),\\
\sigma_{\pi_1}&\succeq \cdots \succeq \sigma_{\pi_K}.
\end{aligned}
\label{eq:prior_set}
\end{equation}
where $\succeq$ denotes the ordering induced by recommendation strength. Each prior is recorded as
\begin{equation}
\begin{aligned}
\pi_k &= \bigl\langle T_k,\rho_k,\delta_k,\sigma_k\bigr\rangle,\\
\sigma_k &= \operatorname{strength}\Bigl(
\operatorname{severity}(\rho_k),\mathcal{Q}(T_k)
\Bigr),\\
\sigma_k &\in \{\mathrm{strong},\mathrm{moderate},\mathrm{exploratory}\}.
\end{aligned}
\label{eq:prior_record}
\end{equation}
Here $T_k\subseteq\bigcup_x\mathcal{M}_x(D)$ is the descriptor trigger set, $\rho_k\subseteq\mathcal{R}(D)$ is the subset of risks mitigated by the prior, and $\delta_k\in\Delta_{\mathrm{design}}$ identifies the design domain, such as architecture, capacity, resolution, supervision, loss, pretraining, or post-processing. Evaluation choices are not included in $\Pi(D)$; they are represented separately by $\mathcal{E}(D)$. The mappings $\Psi$ and $\Gamma$ are not assumed to be universal closed-form functions. In the proposed framework, they represent a structured rule- and evidence-guided process that is made explicit through the MS-DKC record.

\begin{figure*}[t]
    \centering
    \includegraphics[width=\textwidth]{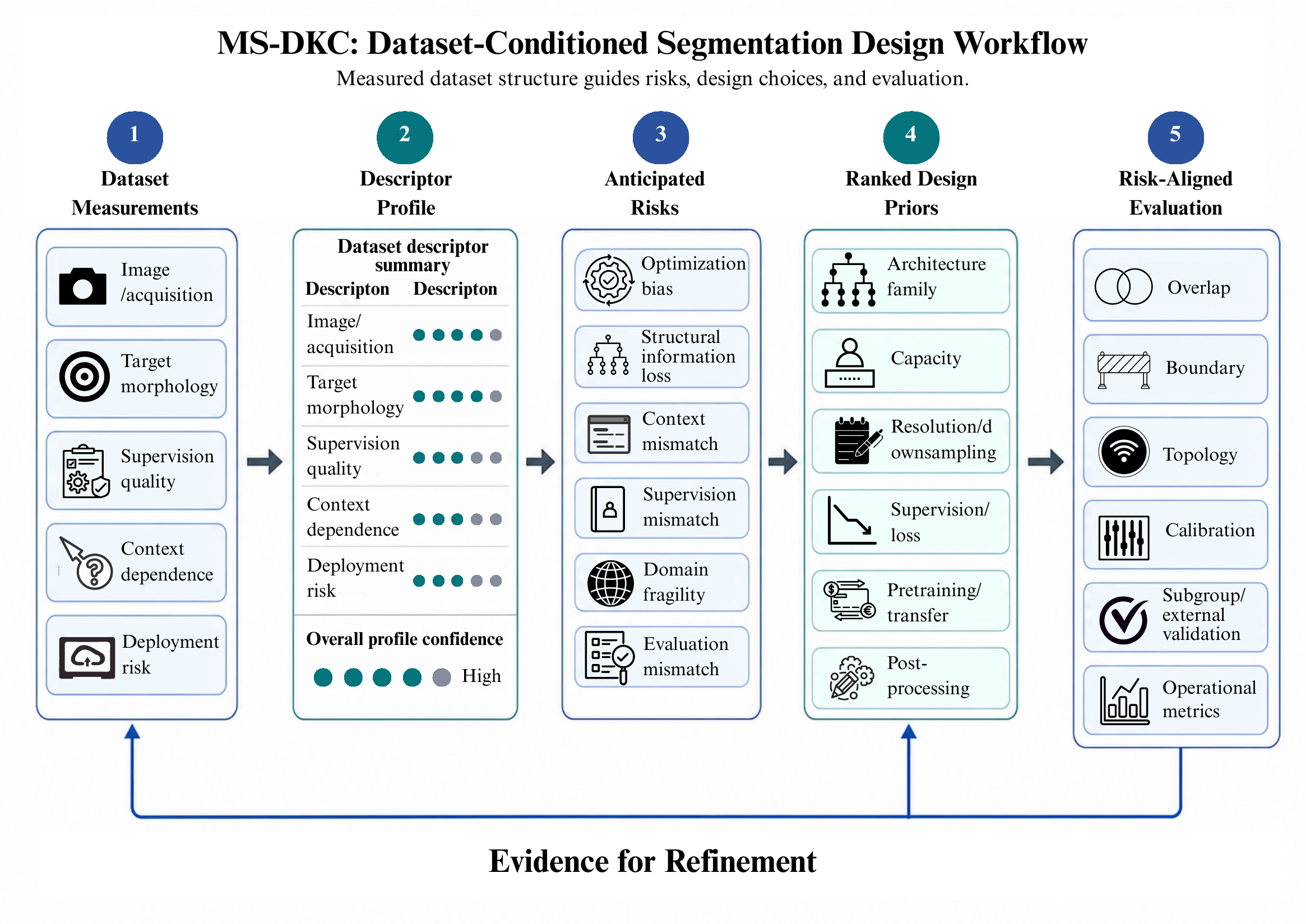}
    \caption{MS-DKC dataset-conditioned segmentation design workflow. Measured dataset structure is translated into descriptor profiles, anticipated risks, ranked design priors, and risk-aligned evaluation choices.}
    \label{fig:msdkc_workflow}
\end{figure*}

Figure~\ref{fig:msdkc_workflow} summarizes the overall MS-DKC workflow. The framework begins with dataset measurements, converts these measurements into a confidence information descriptor profile, maps the profile to anticipated risks, and then derives ranked design priors and risk-aligned evaluation choices. The feedback loop emphasizes that the evaluation results can refine both the profile of the dataset and subsequent design decisions.

The treatment of $\Pi(D)$ as a \emph{ranked set} rather than a singular deterministic prescription is a critical component of this formulation. The objective of MS-DKC is not to eradicate experimentation or to assert that an optimal architecture can be inferred from a closed form. Instead, it restricts the feasible design space and elucidates the justification for potential alternatives. In this regard, the framework is more similar to a structured design that existed prior to the implementation of an automatic solver.

The framework also distinguishes between \emph{task complexity} and \emph{architectural habit}. In a significant portion of the segmentation literature, decisions regarding depth, breadth, downsampling, supervision, or evaluation are inherited from previous models and are only loosely associated with the dataset being examined. Instead, MS-DKC poses the question of which manifestations of complexity are genuinely present. A low-cardinality segmentation task may still be challenging due to excessive sparsity, topology sensitivity, anisotropy, or uncertainty. In contrast, a large-capacity model may be inadequately justified if the data set demonstrates a modest dependency structure and a relatively stable appearance. The framework prevents the conflation of "larger" or "more recent" with "more appropriate" by explicitly defining these distinctions.

The proposed perspective is complementary to existing practice rather than opposed to it. Strong baselines such as U-Net, nnU-Net, transformer-based systems, and foundation-model adaptations can all perform extremely well \cite{ronneberger2015unet,isensee2021nnunet,hatamizadeh2022unetr,ma2024medsam}. The contribution of MS-DKC is not to displace these models, but to provide a principled basis for deciding when and why particular assumptions about capacity, context modeling, spatial fidelity, supervision, and evaluation are justified for a given dataset. The empirical sections of this paper use this framework to examine whether dataset-conditioned design choices provide a more appropriate basis for modeling and evaluation across different segmentation regimes.

The next two sections develop the framework in detail. We first define the descriptor layers that make up $\mathcal{M}(D)$, and then show how those measurements are translated into the anticipated risks and the priors of the ranked design.

\section{Measurement Layers of the Medical Segmentation Dataset Knowledge Card}

The MS-DKC is useful only if the properties of a segmentation dataset are represented in a structured and reproducible form. To this end, we organize the characterization of the dataset into five descriptor layers. These layers are not intended as rigid ontological categories, but as a practical decomposition of the information required to move from dataset profiling to design reasoning. The resulting set of descriptors is
\begin{equation}
\mathcal{M}(D)=
\bigl\langle
\mathcal{M}_{\mathrm{img}}(D),\mathcal{M}_{\mathrm{str}}(D),
\mathcal{M}_{\mathrm{sup}}(D),\mathcal{M}_{\mathrm{ctx}}(D),
\mathcal{M}_{\mathrm{deploy}}(D)
\bigr\rangle,
\label{eq:descriptor_layers}
\end{equation}
where the five components correspond to image/acquisition descriptors, target-structure descriptors, supervision descriptors, context-dependence descriptors, and deployment-risk descriptors.

This decomposition is intentionally designed to avoid reducing segmentation difficulty to a single scalar. A dataset may be simple in one sense and challenging in another: for example, a task may have only one foreground class, yet it may still be difficult because of extreme sparsity, boundary ambiguity, or topology sensitivity. The role of the descriptor layers is therefore to preserve the structure of the dataset profile rather than to collapse it prematurely.

\subsection{Image and Acquisition Descriptors}

The first layer records the properties of the imaging process itself. Medical images are shaped by modality-specific acquisition physics, scanner protocols, reconstruction pipelines, resolution limits, and artifact sources. Consequently, datasets that appear similar at the task level can differ substantially in intensity statistics, spatial sampling, anisotropy, noise characteristics, or cross-site heterogeneity.

Representative descriptors include modality, image dimensionality, voxel spacing, anisotropy ratio, field of view, intensity range, contrast variation, signal-to-noise characteristics, scanner or site information, and known acquisition artifacts. These descriptors are closely related to the normalization strategy, the selection of dimensions, the design of the augmentation, the requirements of the receptive field, and the expectations of robustness. They also provide an early indication of domain-shift exposure: a dataset acquired under a single protocol poses a different generalization challenge than a dataset aggregated across scanners, institutions, or staining processes.

\subsection{Target-Structure Descriptors}

The second layer records the characteristics of the anatomical or pathological targets. This layer is critical because many important segmentation errors arise from a mismatch between model behavior and target morphology rather than from a general semantic limitation.

Representative descriptors include foreground occupancy, object count per image or volume, size distribution, aspect-ratio variation, boundary thinness, connectivity, branching complexity, shape irregularity, and topology criticality. Some targets are large, compact, and region-dominant, whereas others are sparse, filamentary, or highly sensitive to small contour errors. These differences directly affect the risks associated with common design decisions, such as repeated pooling, coarse patching, and decoder resolution.

\subsection{Supervision Descriptors}

The third layer records the properties of the available supervision. In medical segmentation, labels are often expensive to obtain and may be uncertain, partial, inconsistent between raters, or protocol dependent. These differences should not be treated as hidden nuisances, because they materially affect which losses, regularizers, and evaluation criteria are appropriate.

Representative descriptors include class count, foreground imbalance, rare-label frequency, proportion of partially labeled data, annotation protocol variation, inter-rater agreement, boundary ambiguity, weak-label availability, and systematic label noise. These descriptors influence not only model training, but also how confidently performance should be interpreted.

\subsection{Context-Dependence Descriptors}

The fourth layer records the extent and type of context required to solve the segmentation problem. This layer replaces a vague notion of ``representation difficulty'' with a more operational question: how much local, multiscale, long-range, or volumetric dependency appears necessary for reliable delineation?

Representative descriptors include anisotropy-driven context asymmetry, target extent, multiplicity, dependence on wider anatomical layout, sensitivity to resolution loss, foreground-background appearance overlap, and evidence of long-range spatial interaction. These descriptors are directly relevant to the choice among local convolutional designs, multiscale hybrids, transformer-based systems, or other long-context architectures. Importantly, they characterize the dataset rather than presuppose a particular learned representation.

\subsection{Deployment-Risk Descriptors}

The fifth layer records properties related to the environment in which the segmentation model is expected to operate. A model that performs well in retrospective benchmarking may still be poorly matched to deployment if it is fragile under site shifts, poorly calibrated, too slow, or prone to clinically severe rare errors.

Representative descriptors include expected latency and memory constraints, edge versus server deployment, calibration requirements, expected out-of-distribution exposure, subgroup imbalance, robustness demands, human-in-the-loop review assumptions, and the downstream severity of false positives and false negatives. These descriptors are included because model suitability cannot be judged only from image statistics and label structure; it also depends on operational and clinical context.

\subsection{Descriptor Quality and Measurement Confidence}

Not all descriptors are equally direct or equally reliable. For this reason, the MS-DKC records not only descriptor values but also the quality of those values. For each descriptor $m_j \in \mathcal{M}(D)$, we record
\begin{equation}
\mathcal{Q}(D)=
\bigl\{q_j:m_j\in\bigcup_x\mathcal{M}_x(D)\bigr\},
\qquad
q_j=\bigl\langle\mathrm{source}_j,\mathrm{method}_j,\mathrm{confidence}_j\bigr\rangle .
\label{eq:descriptor_quality_layer}
\end{equation}
where $\mathrm{source}_j$ specifies whether the descriptor was obtained from raw images, annotations, metadata, protocol documentation, or expert judgment; $\mathrm{method}_j$ indicates whether it was measured directly, estimated by proxy, or assigned qualitatively; and $\mathrm{confidence}_j$ summarizes how strongly it should influence downstream reasoning.

This distinction is important because some descriptors, such as voxel spacing or foreground occupancy, are straightforward to compute, whereas others, such as topology criticality or clinically meaningful boundary tolerance, may depend partly on expert interpretation. Recommendations should therefore be stronger when supported by multiple high-confidence descriptors and more cautious when they arise from uncertain or proxy-derived measurements.

\begin{table*}[htbp!]
\centering
\caption{Core descriptor schema of the Medical Segmentation Dataset Knowledge Card (MS-DKC). Required descriptors are expected for most segmentation datasets; optional descriptors are task- or deployment-dependent.}
\label{tab:msdkc_schema}
\begin{tabular}{p{2.6cm}p{5.2cm}p{2.0cm}p{2.0cm}p{2.0cm}p{2.3cm}}
\toprule
\textbf{Layer} & \textbf{Representative descriptors} & \textbf{Type} & \textbf{Source} & \textbf{Status} & \textbf{Typical linked risks} \\
\midrule
Image/acquisition
& Modality, dimensionality, voxel spacing, anisotropy, field of view, contrast variation, scanner/site heterogeneity
& Num./cat.
& Image, metadata
& Required
& Context mismatch, domain fragility \\

Target structure
& Foreground occupancy, object count, size distribution, boundary thinness, branching complexity, topology criticality
& Num./cat.
& Annotation, image
& Required/ optional
& Optimization bias, structural information loss \\

Supervision
& Class imbalance, partial labels, inter-rater agreement, boundary ambiguity, weak labels, label noise
& Num./cat.
& Annotation, protocol
& Required/ optional
& Supervision mismatch, miscalibration \\

Context dependence
& Long-range dependency, appearance overlap, anisotropy-driven context asymmetry, sensitivity to resolution loss
& Num./cat.
& Image, annotation, proxy
& Optional
& Context mismatch, capacity mismatch \\

Deployment risk
& Site-shift exposure, subgroup imbalance, latency/memory constraints, failure severity, calibration requirement
& Cat.
& Metadata, protocol, expert
& Optional
& Domain fragility, evaluation mismatch \\
\bottomrule
\end{tabular}
\end{table*}

\section{From Dataset Measurements to Design Priors}

The layered structure of MS-DKC becomes practically useful only when dataset descriptors are translated into concrete design implications. Building on the formulation above, we use the term \emph{design prior} to denote a traceable starting hypothesis about architecture, capacity, preprocessing, supervision, loss, pretraining, or post-processing that is supported by one or more measured dataset descriptors. A design prior is not a fixed rule and is not a substitute for empirical validation. Rather, it is a ranked recommendation that narrows the plausible design space and makes explicit why particular interventions should be considered more seriously than others.

The central inference step of MS-DKC is the transformation from the descriptor profile to the anticipated risk profile,
\begin{equation}
\mathcal{R}(D)=\Psi\bigl(\mathcal{M}(D),\mathcal{Q}(D)\bigr)
=\bigl\{(r,s_r):r\in\mathfrak{R},\ s_r\in[0,1]\bigr\},
\end{equation}
followed by the transformation from anticipated risks to ranked design priors,
\begin{equation}
\begin{aligned}
\Pi(D)&=\Gamma\bigl(\mathcal{R}(D),\mathcal{Q}(D)\bigr)
=\bigl(\pi_1(D),\ldots,\pi_K(D)\bigr),\\
\sigma_{\pi_1}&\succeq\cdots\succeq\sigma_{\pi_K}.
\end{aligned}
\end{equation}
Each prior $\pi_k(D)$ is recorded together with (i) the descriptor or descriptors that triggered it, (ii) the risk or risks it is intended to mitigate, (iii) the design domain to which it belongs, and (iv) a recommendation-strength label such as \emph{strong}, \emph{moderate}, or \emph{exploratory}. This structure is important because it prevents the framework from becoming a vague checklist: a recommendation is useful only if it is traceable to measured evidence.

\subsection{Foreground Sparsity, Class Imbalance, and Optimization Bias}

The extreme foreground-background imbalance is one of the most common descriptor patterns in medical segmentation. In many tasks, the clinically relevant target occupies only a small fraction of the image or volume, whereas the background dominates the optimization signal. The resulting risk is \emph{optimization bias}: a model may achieve apparently strong average overlap while under-segmenting rare or clinically important structures.

This descriptor pattern supports design priors that involve imbalance-based losses, foreground-based cropping, hard-example sampling, and evaluation protocols that explicitly report recall of minority-structures or lesion-level sensitivity. In a workflow conditioned by the dataset, such interventions should not be treated as optional enhancements; they are direct responses to a measured occupancy pattern.

\subsection{Thin Structures, Boundary Sensitivity, and Structural Information Loss}

When the target is thin, branching, poorly contrasted, or topologically fragile, the main danger is \emph{structural information loss} caused by repeated downsampling, coarse patching, or inadequate decoder resolution. This pattern is common in vascular segmentation, duct or catheter segmentation, membrane delineation, boundary-critical organ segmentation, and many small-lesion contexts.

In these cases, MS-DKC supports stronger priors for delayed downsampling, high-resolution feature pathways, multiscale fusion, boundary-aware supervision, and topology-sensitive regularization \cite{zhou2020unetplusplus,huang2020unet3plus,valanarasu2020kiunet,kervadec2019boundary,shit2021cldice}. The same descriptor profile should also influence the evaluation: region-overlap metrics alone are often insufficient when connectivity or boundary fidelity is clinically meaningful.

\subsection{Low Label Cardinality and Capacity Mismatch}

Many medical segmentation tasks involve one foreground class or only a small number of clinically relevant labels. This does not imply low difficulty, but it changes the nature of the representational burden. In such cases, difficulty often arises from morphology, ambiguity, sparsity, or context dependence rather than broad semantic separation. The associated risk is \emph{capacity mismatch}: model scale may be inherited from architectures optimized for very different semantic regimes.

To emphasize that capacity should be justified rather than inherited, we write the appropriate capacity prior conceptually as
\begin{equation}
c^{*}(D)\in\Pi(D),
\qquad
c^{*}(D)=f_c\!\left(s(D),\,t(D),\,u(D),\,m(D),\,d(D),\,\kappa(D)\right),
\label{eq:capacity_prior}
\end{equation}
where the scalar summaries are projections of the descriptor layers: $s(D)$ and $t(D)$ describe sparsity/scale and topology/boundary properties from $\mathcal{M}_{\mathrm{str}}$, $u(D)$ captures supervision uncertainty from $\mathcal{M}_{\mathrm{sup}}$, $m(D)$ captures modality and acquisition from $\mathcal{M}_{\mathrm{img}}$, $d(D)$ captures context dependence from $\mathcal{M}_{\mathrm{ctx}}$, and $\kappa(D)$ captures deployment constraints from $\mathcal{M}_{\mathrm{deploy}}$. This expression is not intended as a calibrated estimator of optimal capacity. It is a compact way to state that the capacity should depend on the properties of the data set rather than the architectural habit.

\subsection{Context Dependence, Anisotropy, and Architecture Family}

Some segmentation challenges are predominantly determined by local edge and texture indicators, while others are dependent on broader anatomical architecture, multiscale interaction, or volumetric continuity. Descriptors such as anisotropy, target extent, multiplicity, and evidence of long-range dependence reflect these requirements. The resulting risk is \emph{context mismatch}: the model either fails to capture dependencies that are significant or wastes modeling complexity on dependencies that are not.

At this stage, the architecture family becomes a conditional design choice. For local, detail-sensitive tasks with modest context demands, carefully configured U-Net-like models may remain appropriate. However, anisotropic volumetric datasets or problems with stronger long-range interactions may justify hybrid, transformer-based, or other long-context architectures \cite{chen2021transunet,hatamizadeh2022unetr,cao2022swinunet,hatamizadeh2022swinunetr,deng2021missformer,ma2024umamba,xing2024segmamba}. The key point is not that one architecture family prevails in general, but that the architectural decision should be based on the measured dependence structure of the dataset.

\subsection{Annotation Uncertainty, Supervision Mismatch, and Evaluation Design}

When boundaries are ambiguous, raters disagree, or some labels are missing, deterministic supervision may promote overfitting to annotation artifacts and produce misleadingly confident predictions. The resulting risk is \emph{supervision mismatch}: both the training objective and the evaluation process may assume a level of label certainty that the dataset does not support.

This profile supports the use of soft-label formulations, uncertainty-aware learning, confidence calibration analysis, robust losses, and explicit rater-variation modeling. It also means that evaluation should not be based exclusively on single-number overlap summaries since such summaries do not reflect annotation variability. When supervision is uncertain, the problem is not only how the model is trained but also how its performance is interpreted against imperfect labels.

\subsection{Modality Heterogeneity, Site Shift, and Domain Fragility}

Variation in scanner, protocol, reconstruction, staining, or institution introduces a direct risk of \emph{domain fragility}. A model may perform well under the training distribution but degrade substantially when these acquisition factors change. The MS-DKC framework includes measurements that support modality-aware normalization, realistic augmentation, domain-adaptive pretraining, subgroup-wise evaluation, calibration checks, and stronger external validation requirements \cite{ma2024medsam,cheng2023sammed2d,wu2023medicalsamadapter,ma2025medsam2}.

More generally, these descriptors caution against interpreting pretrained representations as universally transferable. A transferred encoder may still be useful, but its effectiveness depends on how well the source prior matches the imaging domain and the structural properties that define the segmentation task.

\subsection{Conflict Resolution and Recommendation Strength}

In practice, descriptor signals may conflict. For example, a dataset may require both high spatial preservation and wide context modeling. The purpose of MS-DKC is not to remove such tension, but to make it explicit. The elements of $\Pi(D)$ should, therefore, be treated as ranked hypotheses rather than as immutable rules. Some recommendations will be strong because they are supported by multiple high-confidence descriptors, whereas others will be moderate or exploratory because the evidence is weaker or partly contradictory.

This reflects how segmentation systems are developed in practice: not through unconstrained architectural search, but through controlled and interpretable comparison of a narrower set of viable interventions.

\subsection{From Design Priors to Reproducible Segmentation Workflows}

The practical significance of MS-DKC is not that it removes experimentation but that it changes how experimentation is organized. Instead of treating pipeline design as a largely architecture-centered search problem, the framework encourages researchers to begin with an explicit dataset profile, derive a risk profile from that characterization, and then compare a smaller set of descriptor-justified interventions.

The evaluation protocol should also be selected as a function of the inferred risk profile:
\begin{equation}
\begin{aligned}
\mathcal{E}(D)
&=g\bigl(\mathcal{R}(D),\mathcal{Q}(D)\bigr)
=\bigl(\epsilon_1(D),\ldots,\epsilon_L(D)\bigr),\\
\delta_{\epsilon}&\in\Delta_{\mathrm{eval}},
\qquad
\Delta_{\mathrm{design}}\cap\Delta_{\mathrm{eval}}=\varnothing .
\end{aligned}
\label{eq:eval_from_risk_main}
\end{equation}
Here $\mathcal{E}(D)$ denotes the ranked metric and validation choices most appropriate for the identified risks. As with the design-prior mapping, this expression denotes a structured selection process rather than a closed-form rule. Boundary-sensitive, topology-aware, uncertainty-aware, subgroup-specific, calibration-oriented, or external-validation metrics should therefore be emphasized whenever the corresponding risks are present in the MS-DKC. The explicit separation between $\Pi(D)$ and $\mathcal{E}(D)$ prevents design recommendations and evaluation criteria from being conflated.
\begin{equation}
\bigl(\mathcal{M}^{(t+1)},\mathcal{Q}^{(t+1)}\bigr)
=
\operatorname{Refine}\!\left(
\mathcal{M}^{(t)},\mathcal{Q}^{(t)},
\mathrm{evidence\ from\ }\mathcal{E}^{(t)}
\right),
\label{eq:refinement_loop}
\end{equation}
after which $\mathcal{R}$, $\Pi$, and $\mathcal{E}$ are recomputed for iteration $t+1$. This closes the design loop rather than treating the workflow as a single pass.

This perspective provides the basis for the critique developed in the next section. If design priors should emerge from measurable dataset structure, then dominant defaults such as U-Net-like pipelines must be examined not only in terms of empirical convenience, but also in terms of whether their assumptions about downsampling, capacity, context modeling, supervision, and evaluation are justified for the dataset at hand.

\begin{table*}[htbp!]
\centering
\caption{Representative measurement-to-risk-to-intervention mappings in the MS-DKC framework.}
\label{tab:risk_mapping}
\begin{tabular}{p{3.8cm}p{2.0cm}p{2.5cm}p{4.8cm}p{2.8cm}}
\toprule
\textbf{Descriptor pattern} & \textbf{Anticipated risk} & \textbf{Affected design domain} & \textbf{Representative interventions} & \textbf{Evaluation implications} \\
\midrule
Extreme foreground sparsity, severe class imbalance
& Optimization bias
& Loss, sampling, training distribution
& Dice-, Tversky-, or focal-style objectives; foreground-aware cropping; hard-example mining
& Lesion-level sensitivity; minority-structure recall \\

Thin, branching, topology-sensitive targets
& Structural information loss
& Resolution policy, supervision, decoder design
& Delayed downsampling; high-resolution pathways; boundary-aware or topology-aware losses
& Boundary and connectivity metrics \\

Low label cardinality with modest semantic diversity but high morphology sensitivity
& Capacity mismatch
& Encoder width/depth, pretraining, architecture scale
& Lighter backbones; stronger spatial preservation; selective transfer
& Capacity ablations rather than scale-by-default comparisons \\

High anisotropy, long-range dependency, wide anatomical context
& Context mismatch
& Architecture family, dimensionality
& 2.5D, anisotropic 3D, or hybrid long-context models
& Cross-slice and volumetric consistency analysis \\

High inter-rater disagreement, partial labels, boundary ambiguity
& Supervision mismatch
& Loss design, uncertainty modeling, calibration
& Soft labels; robust objectives; rater-aware training
& Uncertainty-aware and agreement-aware evaluation \\

Scanner/site/stain heterogeneity, deployment-shift exposure
& Domain fragility
& Normalization, augmentation, validation strategy
& Modality-aware preprocessing; realistic augmentation; external validation
& Subgroup analysis; calibration; external-site validation \\
\bottomrule
\end{tabular}
\end{table*}



\subsection{Worked Example: A Thin-Structure, Sparse-Target Dataset}

To illustrate the intended use of MS-DKC, consider a retinal vessel segmentation dataset based on fundus images. Suppose that the measured profile includes: (i) extremely low foreground occupancy, (ii) thin and branching target morphology, (iii) high sensitivity to connectivity errors, (iv) moderate image-quality variation across devices, and (v) limited annotation uncertainty. In MS-DKC form, the dominant descriptor pattern is sparse occupancy, topology-critical thin structure, and mild acquisition heterogeneity.

This profile supports three anticipated risks: optimization bias toward background prediction, structural information loss under aggressive downsampling, and moderate domain fragility across imaging devices. The resulting ranked design priors would prioritize (1) imbalance-aware loss design or foreground-aware sampling, (2) spatial preservation through delayed downsampling or higher-resolution pathways, (3) boundary-aware or topology-aware supervision, and (4) evaluation beyond region overlap, including connectivity or topology-sensitive criteria.

This example does not imply that one design is generally ideal for vessel segmentation. Rather, it shows how a dataset profile can reduce the plausible design space and make the rationale for those decisions explicit. A different dataset, such as large-organ CT segmentation with broader targets and lower topology sensitivity, would produce a different MS-DKC profile and therefore a different set of initial design priors.

\section{Rethinking U-Net as a Default Design Prior}

The purpose of MS-DKC is not to dismiss historically successful architectures, but to make their assumptions visible and examine whether those assumptions are supported by the measured structure of the dataset. This perspective is especially relevant for U-Net, whose influence on medical image segmentation has become so widespread that it often serves not only as a strong baseline, but also as an unquestioned default.

\subsection{Why U-Net Became the Dominant Baseline}

The historical importance of U-Net in medical image segmentation is well justified. Its influence rests on practical and methodological advantages that remain relevant today. First, U-Net can be effectively trained under limited data, making it attractive in medical settings where dense annotation is expensive and the size of the data set is often modest \cite{ronneberger2015unet}. Second, its encoder--decoder structure with skip connections offers an effective compromise between contextual aggregation and spatial detail preservation. Third, the architecture is simple, modular, and adaptable in modalities and tasks, which has facilitated its use in organ segmentation, lesion delineation, microscopy, retinal imaging, and pathology.

Subsequent variants reinforced this position while preserving the basic encoder--decoder prior. UNet++ redesigned the skip pathways to reduce the semantic gap between the encoder and decoder features \cite{zhou2020unetplusplus}. Attention U-Net introduced gating mechanisms to improve focus on task-relevant regions \cite{oktay2018attentionunet}. UNet 3+ emphasized dense multiscale aggregation and deep supervision \cite{huang2020unet3plus}. Most importantly, nnU-Net showed that a large fraction of segmentation performance depends on dataset-aware configuration of preprocessing, training, and architecture, while remaining strongly aligned with a U-Net-like design template \cite{isensee2021nnunet}. Together, these developments made U-Net not only a successful architecture, but also a reference architecture for medical segmentation.

\subsection{A Baseline Is Not the Same as a Universally Justified Default}

Any critique of U-Net must begin from a simple observation: for many medical segmentation tasks, a well-configured U-Net remains an excellent baseline. U-Net-like models frequently provide a robust equilibrium of interpretability, computational efficiency, performance, and implementation maturity in problems with moderate anatomical scale, reasonably stable appearance, limited data, and context requirements that can be captured by hierarchical local aggregation \cite{ronneberger2015unet,isensee2021nnunet}. Their implementation ecosystem is well-established across 2D, 2.5D, and 3D variations, and their multiscale structure supports contextual aggregation. In addition, their skip connections help conserve spatial detail.

The problem is therefore not that U-Net is weak. The problem is that it is often treated as a \emph{default assumption} rather than as a \emph{dataset-tested baseline}. This distinction is central to a dataset-conditioned design framework. A baseline is a candidate model class whose suitability should be evaluated against the structure of the dataset. In contrast, a default is a starting assumption that may escape scrutiny precisely because it has become familiar. The purpose of this section is therefore not to displace U-Net from medical segmentation, but to make its assumptions visible again.

\subsection{Which U-Net Assumptions Are Actually Being Inherited?}

From a dataset-conditioned perspective, the key question is not whether one uses a U-Net-like architecture, but which design assumptions are being inherited when U-Net is chosen as the starting point. Several assumptions recur across practice.

\paragraph{Downsampling as a benign operation.}
Standard U-Net reasoning often treats repeated pooling or stride-based reduction as a routine mechanism for expanding the receptive field and aggregating context. This can be effective when target structures are sufficiently large and spatial detail is recoverable through skip connections. However, for many medical segmentation tasks, the main challenge lies precisely in preserving fine-scale evidence: thin vessels, membranes, small lesions, catheters, or fragile boundaries. In such settings, aggressive downsampling can remove the information needed for clinically meaningful delineation before the decoder has the opportunity to recover it. The issue is therefore not whether U-Net contains a recovery pathway, but whether the dataset profile justifies the degree of resolution loss assumed by the architecture \cite{ronneberger2015unet,huang2020unet3plus,valanarasu2020kiunet}.

\paragraph{Monotonic depth-wise capacity growth.}
U-Net variants typically allocate increasing channel width with depth, reflecting a familiar vision-design logic in which progressively abstract representations require progressively richer feature banks. This is often sensible, especially when the task requires a broad contextual integration. However, it is not automatically justified in low-cardinality segmentation tasks where the decisive difficulty may lie in boundary fidelity, topology preservation, supervision uncertainty, or deployment efficiency rather than semantic abstraction. In such cases, inherited width schedules may allocate capacity to aspects of the problem that are less relevant than spatial precision or robustness. The question is therefore not simply whether the model is large or small, but whether its capacity profile is aligned with the representational burden implied by the dataset.

\paragraph{Local hierarchical aggregation as sufficient context modeling.}
U-Net-like architectures assume that local convolutional aggregation combined with hierarchical pooling provides the relevant context for prediction. For many tasks, this assumption is well supported. However, some segmentation problems depend on broader anatomical relationships, multiscale interaction, long-range volumetric dependence, or anisotropy-aware context. In such cases, the appropriate design question is not whether U-Net is obsolete but whether the dataset requires a context mechanism beyond what the standard local encoder--decoder structure provides by default.

\paragraph{Transferred encoder priors as broadly appropriate.}
In many practical pipelines, especially those influenced by natural-image segmentation and pretrained backbones, encoder design inherits assumptions from large-scale visual recognition. These transferred priors may improve the optimization and sample efficiency, but they also reflect forms of semantic richness that are not always the dominant source of difficulty in medical segmentation. A dataset with one foreground class may still be difficult because of sparsity, topology, uncertainty, or acquisition shift rather than broad category-level variation. The relevant question is therefore not whether the transfer is useful in general, but whether the transferred prior addresses the measured source of difficulty in the dataset.

\begin{table*}[htbp!]
\centering
\caption{Common inherited U-Net assumptions viewed through the MS-DKC framework.}
\label{tab:unet_assumptions}
\begin{tabular}{p{2.49cm}p{3.0cm}p{3.99cm}p{3.6cm}p{2.9cm}}
\toprule
\textbf{Inherited assumption} & \textbf{When often defensible} & \textbf{When it becomes risky} & \textbf{MS-DKC descriptors to inspect} & \textbf{Possible MS-DKC decision} \\
\midrule
Repeated downsampling is benign
& Moderately sized structures, stable appearance, limited topology sensitivity
& Thin, sparse, branching, or boundary-critical targets
& Foreground occupancy, boundary thinness, topology criticality, target scale distribution
& Accept, delay, or redesign downsampling \\

Monotonic depth-wise capacity growth is beneficial
& Tasks requiring progressively richer abstraction and broad contextual integration
& Low-cardinality tasks dominated by boundary precision, sparsity, uncertainty, or deployment constraints
& Label cardinality, morphology complexity, supervision quality, deployment constraints
& Retain, reduce, or ablate capacity schedule \\

Local hierarchical aggregation provides sufficient context
& Local texture and boundary cues dominate; context needs are moderate
& Strong long-range dependence, broad anatomical interaction, or anisotropic volumetric context
& Context dependence, anisotropy, target extent, multiplicity, field of view
& Retain U-Net, adapt context path, or compare long-context models \\

Transferred encoder priors are broadly appropriate
& Source and target domains are reasonably aligned; data are scarce but stable
& Difficulty is driven by acquisition shift, topology sensitivity, annotation uncertainty, limited semantic diversity, or strong source-domain mismatch
& Acquisition heterogeneity, domain-shift exposure, annotation uncertainty, target morphology
& Use transfer, replace with in-domain pretraining, or compare with training from scratch \\
\bottomrule
\end{tabular}
\end{table*}

Table~\ref{tab:unet_assumptions} summarizes these assumptions in the language of MS-DKC and highlights the descriptor patterns that should be examined before a U-Net prior is accepted as justified.

\subsection{How a U-Net Prior Is Justified in MS-DKC}

In the MS-DKC framework, U-Net is treated as a conditional design prior rather than a universal starting point. A prior U-Net is justified when the measured dataset profile corroborates its assumptions about resolution loss, hierarchical context aggregation, capacity allocation, supervision reliability, and evaluation adequacy. This implies that the decision to employ a standard or minimally adapted U-Net should be associated with explicit descriptors rather than inherited from convention.

Three potential outcomes can be used to understand the justification process. Initially, the U-Net prior may be \emph{accepted} if the dataset contains moderately sized structures, manageable class imbalance, stable acquisition conditions, modest topology sensitivity, and context requirements that can be conveyed through local hierarchical aggregation. Second, the U-Net prior may be \emph{adapted} when the dataset supports the general encoder-decoder structure but exposes risks that require modification, such as delayed downsampling for thin structures, high-resolution pathways for boundary-sensitive targets, uncertainty-aware supervision for ambiguous labels, or stronger augmentation for acquisition heterogeneity. Third, the U-Net prior may be \emph{challenged} when the dataset profile indicates that standard local encoder--decoder assumptions are likely to be insufficient, for example, under strong long-range volumetric dependence, severe anisotropy, substantial domain shift, or deployment constraints that make inherited capacity schedules inappropriate.

This does not imply that U-Net cannot be used in difficult settings. Rather, it means that its use should be accompanied by a clear account of which dataset descriptors support the selected configuration and which risks require adaptation or comparison against alternatives. In this sense, MS-DKC keeps U-Net central as a strong baseline while preventing its assumptions from being accepted without dataset-conditioned justification. The empirical analysis later in this paper uses this distinction to compare standard U-Net-style baselines with dataset-conditioned adaptations selected from the MS-DKC profile.

\begin{figure*}[t]
    \centering
    \includegraphics[width=\textwidth]{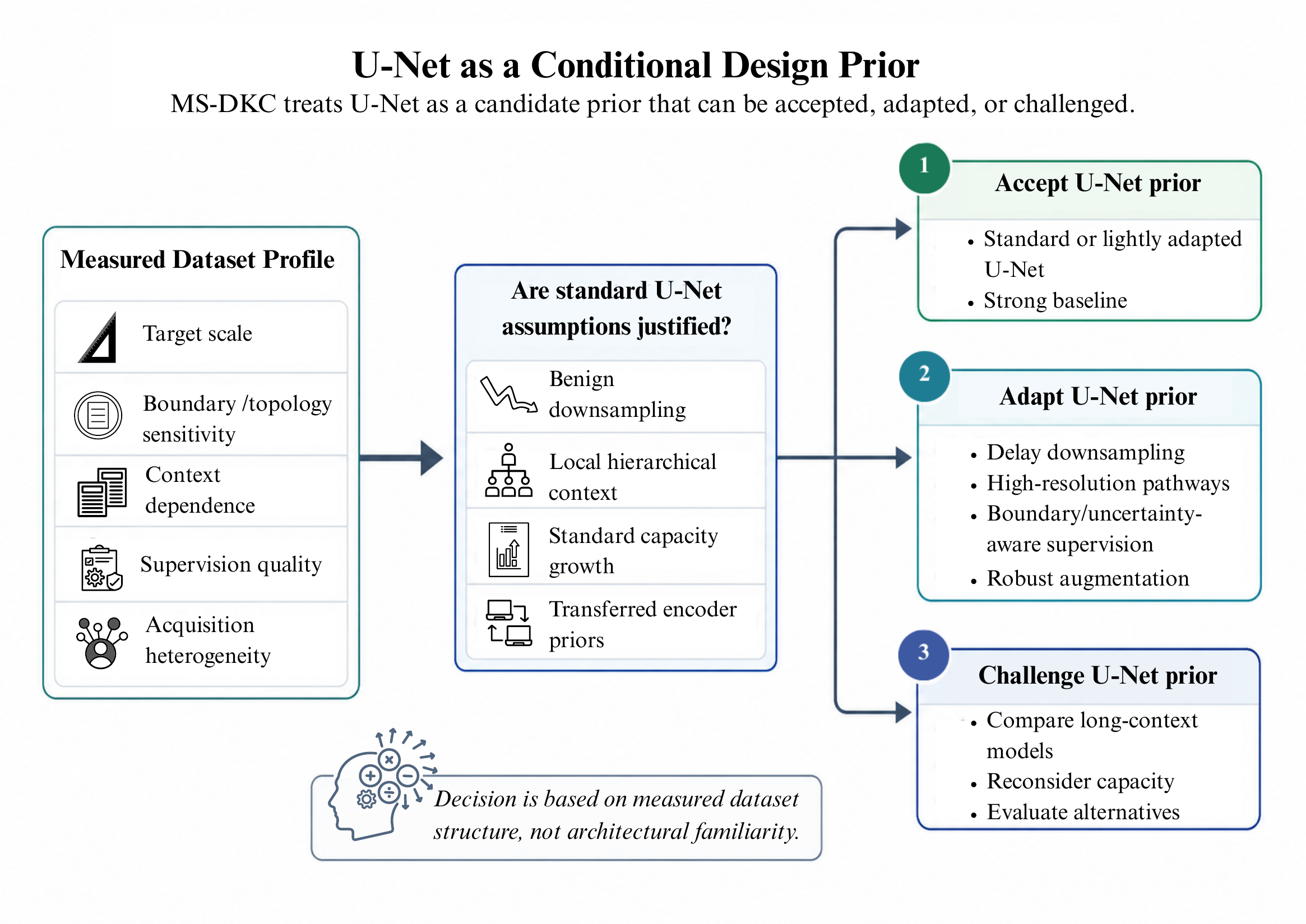}
    \caption{U-Net as a conditional design prior in the MS-DKC framework. The dataset profile determines whether a standard U-Net prior should be accepted, adapted, or challenged.}
    \label{fig:unet_conditional_prior}
\end{figure*}

Figure~\ref{fig:unet_conditional_prior} illustrates how MS-DKC treats U-Net as a conditional design prior. A measured dataset profile is first used to examine whether standard U-Net assumptions are justified. Depending on the descriptor pattern, the U-Net prior may be accepted, adapted, or challenged through comparison with alternative architectures or configurations.

\subsection{What Recent Model Families Actually Reveal}

The emergence of transformer-based, state-space, and foundation-model approaches makes inherited U-Net assumptions easier to see. Transformer-based systems such as TransUNet, UNETR, Swin UNET, Swin UNETR, and MISSFormer emphasize richer long-range interaction and multiscale dependency modeling when a broader context is important \cite{chen2021transunet,hatamizadeh2022unetr,cao2022swinunet,hatamizadeh2022swinunetr,deng2021missformer}. State-space approaches such as U-Mamba and SegMamba revisit related questions through efficient long-range sequence modeling in high-dimensional medical data \cite{ma2024umamba,xing2024segmamba}. Promptable and foundation-model-based systems such as MedSAM, SAM-Med2D, Medical SAM Adapter, and MedSAM2 introduce another possibility: that broad segmentation priors can be transferred between tasks, provided that adaptation accounts for domain mismatch and clinical constraints \cite{ma2024medsam,cheng2023sammed2d,wu2023medicalsamadapter,zhu2024medsam2}.

These developments should not be interpreted as evidence that U-Net has failed. Rather, they reinforce a broader lesson: no single architecture family should be treated as universally privileged independently of dataset structure. Some tasks may benefit from deeper global interaction or transferable priors, whereas others may benefit from simpler, more detail-preserving, or more explicitly dataset-constrained solutions. Architectural novelty is therefore not the central issue. What matters is the compatibility between model assumptions and measurable dataset attributes.

\subsection{When a U-Net Prior Is Actually Well Supported}

A dataset-conditioned framework should be able to justify U-Net when appropriate. In general, a standard or lightly adapted U-Net prior is most defensible when target structures are moderately sized, spatially coherent, and not overly sparse; when local texture and boundary cues are informative; when the required context can be captured through hierarchical local aggregation; when topology sensitivity is modest; and when available supervision is limited but reasonably reliable. Under such conditions, a conventional encoder-decoder baseline or an nnU-Net-style self-configuring system may offer an effective balance of performance, efficiency, and reproducibility \cite{ronneberger2015unet,isensee2021nnunet}.

In the MS-DKC language, this corresponds to a dataset profile with moderate class imbalance, modest topology sensitivity, limited anisotropy, manageable context dependence, relatively stable acquisition conditions, and supervision that is sufficiently reliable for deterministic training and evaluation. In such cases, the value of dataset-conditioned reasoning is not that it rejects U-Net, but that it justifies the U-Net prior explicitly rather than treating it as an implicit default.

\subsection{When U-Net Should No Longer Be Treated as Sufficient by Default}

By contrast, several descriptor patterns suggest that a standard U-Net should not be treated as sufficient without a more careful redesign. Extremely thin or branching targets may justify delayed downsampling or stronger high-resolution pathways. Strong anisotropy or long-range volumetric dependence may motivate 2.5D, 3D, hybrid, or longer-context architectures. Severe annotation uncertainty may justify changes in supervision and evaluation more than changes in raw capacity. Very low label cardinality combined with limited structural diversity may suggest that the inherited encoder width is excessive, especially under deployment constraints. Strong acquisition heterogeneity or site shift may require robustness-oriented interventions that benchmark-style U-Net pipelines do not naturally prioritize.

The key issue is therefore not whether U-Net can be modified to handle these cases, since it often can, but whether the need for modification is inferred from measured dataset structure or discovered only through ad hoc trial and error. MS-DKC provides a language for making that distinction explicit. Rather than beginning with a default U-Net and progressively adding modules until performance improves, one can begin by asking which descriptor patterns already indicate that changes in downsampling, capacity, context mechanism, supervision, or evaluation are likely to be necessary.

\subsection{From U-Net Critique to Capacity and Transfer}

The broader lesson of this section is that U-Net should remain central to medical segmentation, but it should not remain invisible as a design assumption. Its historical success makes it a powerful baseline, not a universal answer. Once the dataset-conditioned perspective is adopted, the more precise question becomes: which components of the U-Net design are genuinely supported by the dataset, and which persist mainly because they are inherited defaults?

This question leads directly to the next issue addressed in the paper: capacity, pretraining, and transfer. If standard encoder widths, depth patterns, and inherited natural-image priors are not automatically justified for every medical segmentation problem, then a more explicit account of how dataset structure should inform capacity allocation and the use of transferred representations is needed.

\section{Capacity, Pretraining, and Transfer Under Low-Cardinality Supervision}

The critique of inherited U-Net defaults raises a broader question: how much model capacity is genuinely warranted for a given medical segmentation task, and under what conditions should transferred priors from natural-image or large-scale segmentation models be used? These two issues are often intertwined in contemporary practice. Large encoders are frequently adopted because pretraining is accessible and is generally assumed to be helpful when labeled medical data is limited. Both assumptions can be reasonable, but neither should be treated as universally valid. The essential question is not whether capacity and transfer are beneficial in general, but whether they align with the measurable structure of the segmentation dataset.

As in the previous section, MS-DKC provides the relevant framework: capacity and transfer should be viewed as \emph{dataset-conditioned design priors} rather than inherited defaults. This means that the choice to use a large backbone, a pretrained encoder, an in-domain self-supervised model or a foundation-model adaptation should be justified by measurable dataset attributes rather than by architectural familiarity or benchmark momentum.

\subsection{Low Label Cardinality Changes the Interpretation of Capacity}

One reason why this issue is often overlooked is that medical segmentation inherits a notion of model strength from natural-image learning that is closely tied to semantic breadth. Large backbones, high-dimensional representations, and expressive context mechanisms were originally motivated by tasks requiring discrimination among many object categories in highly variable natural scenes. Medical segmentation often differs in an important way: the number of clinically relevant labels is usually limited, sometimes consisting of a single foreground class or only a few anatomical or pathological structures. This does not make such tasks easy, but it changes the nature of their difficulty.

In many medical datasets, the dominant burden is not broad semantic discrimination, but preservation of fine boundaries, topology, sparsity, weak contrast, acquisition heterogeneity, or long-range anatomical consistency. A large encoder may still be useful, but its usefulness cannot be inferred simply from the success of high-capacity models in natural-image benchmarks. Low label cardinality therefore does not imply that little capacity is needed; rather, it weakens the assumption that generic semantic capacity is automatically the right kind of capacity.

\subsection{Capacity Mismatch as a Dataset-Conditioned Risk}

This distinction motivates the notion of \emph{capacity mismatch} as a dataset-conditioned risk. Capacity mismatch occurs when a model is either under-specified or over-specified relative to the structural and representational demands of the dataset. An under-specified model may lack the receptive-field range, feature diversity, or spatial precision needed to separate foreground from background under challenging appearance variation. An over-specified model, by contrast, may allocate substantial representational resources to a task whose primary difficulty lies elsewhere, for example, in annotation noise, calibration, boundary ambiguity, domain fragility, or deployment constraints.

The practical consequences extend beyond wasted computation. Capacity mismatch can also lead to unstable optimization, overfitting under limited supervision, poorly calibrated confidence, and misleading explanations of why a model succeeds or fails. These costs are especially important in medical segmentation, where annotation is expensive and deployment conditions may be restrictive. In the MS-DKC framework, capacity is therefore not a fixed architectural attribute to be inherited passively, but a design variable that should be matched to the measured structure of the dataset.

\begin{table*}[htbp!]
\centering
\caption{Capacity and transfer choices viewed through the MS-DKC framework.}
\label{tab:capacity_transfer}
\begin{tabular}{p{2.5cm}p{4.5cm}p{4.9cm}p{5.1cm}}
\toprule
\textbf{Design choice} & \textbf{When often justified} & \textbf{When it becomes risky} & \textbf{MS-DKC descriptors to inspect} \\
\midrule
Large model capacity
& Strong appearance variation, long-range dependence, heterogeneous context, multimodal complexity
& Low-cardinality tasks dominated by sparsity, topology, boundary detail, or severe deployment constraints
& Context dependence, acquisition heterogeneity, morphology complexity, deployment limits \\

Generic natural-image pretraining
& Small datasets, moderate source--target compatibility, optimization support needed
& Strong modality mismatch, boundary-critical detail, weak semantic overlap with source domain
& Modality, acquisition physics, domain-shift exposure, target morphology \\

In-domain self-supervised or medical-domain pretraining
& Strong modality specificity, large unlabeled in-domain data, source mismatch to natural images
& Limited unlabeled data or weak evidence that domain-specific features address the main bottleneck
& Data availability, modality specificity, acquisition heterogeneity, supervision scarcity \\

Foundation-model or promptable adaptation
& Broad transferable priors are useful and adaptation is compatible with task structure
& High domain mismatch, dominant fine spatial detail, or limited relevance of promptable interaction to deployment
& Context dependence, target scale, supervision quality, deployment workflow, domain fragility \\

High capacity at reduced resolution
& Large structures, moderate topology sensitivity, coarse context dominates
& Thin targets, small lesions, high-frequency boundary detail, strong downsampling sensitivity
& Target scale, topology criticality, sparsity, downsampling sensitivity \\
\bottomrule
\end{tabular}
\end{table*}

\begin{figure*}[t]
    \centering
    \includegraphics[width=\textwidth]{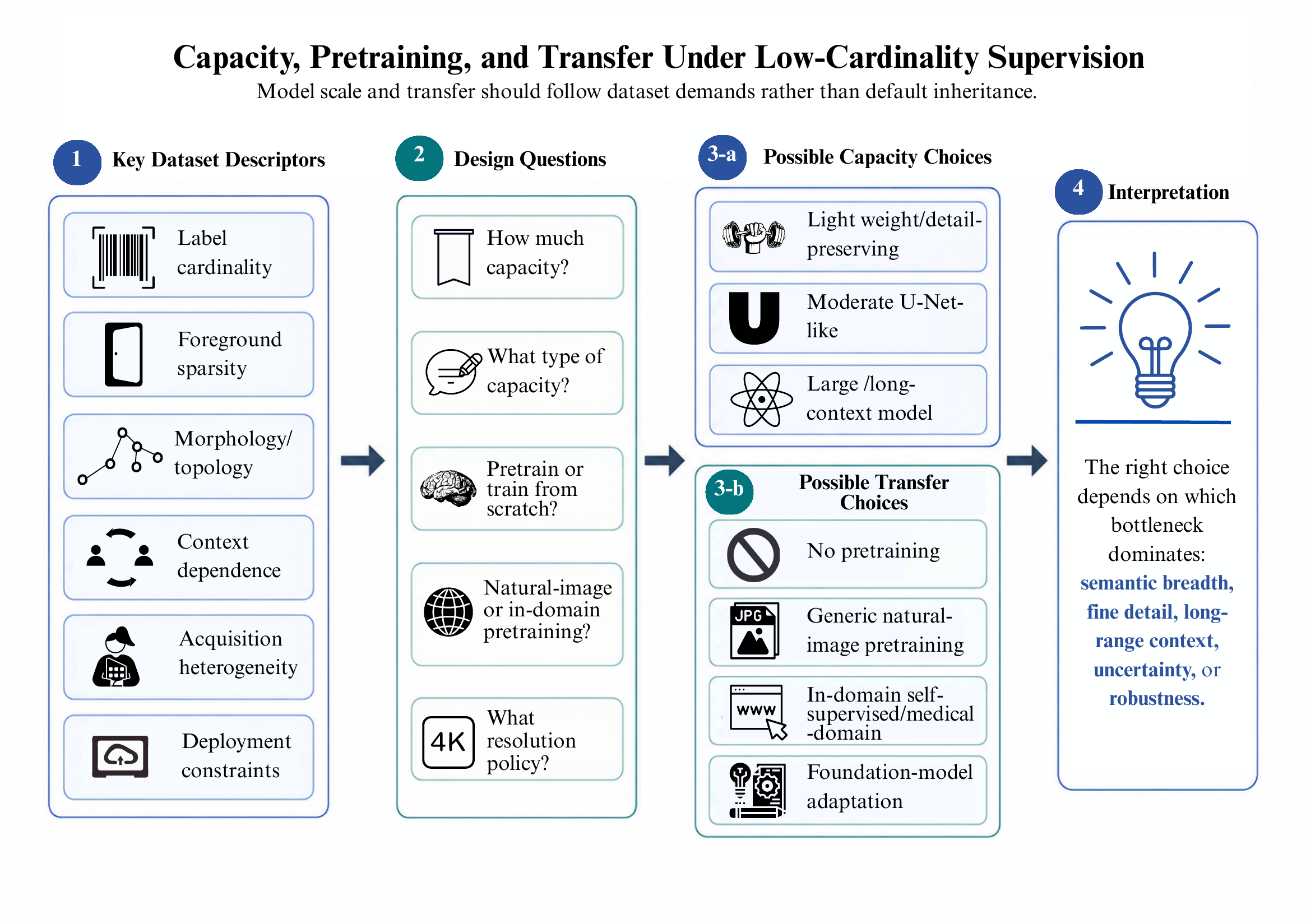}
    \caption{Dataset-conditioned reasoning for capacity, pretraining, and transfer. Model scale and transfer strategy are selected according to measured dataset demands rather than inherited architectural defaults.}
    \label{fig:capacity_transfer}
\end{figure*}

Table~\ref{tab:capacity_transfer} presents these options as dataset-conditioned priors rather than general enhancements. The purpose is not to rank capacity or transfer strategies universally, but to identify descriptor patterns that make particular choices more or less defensible. Figure~\ref{fig:capacity_transfer} summarizes how MS-DKC links dataset descriptors to capacity and transfer decisions. Rather than treating model scale or pretraining as generally beneficial, the framework asks which bottleneck dominates the dataset: semantic breadth, fine spatial detail, long-range context, uncertainty, acquisition shift, or deployment constraints.

\subsection{Which Dataset Properties Actually Matter for Capacity?}

Within the MS-DKC framework, capacity should be understood as a consequence of measurable dataset attributes rather than as a static architectural inheritance. Relevant descriptors include target size distribution, topology sensitivity, foreground sparsity, intra-class variation, inter-class overlap, acquisition heterogeneity, anisotropy, context dependence, and downsampling sensitivity. A task with a single foreground structure may still justify substantial capacity if the target is highly variable, poorly contrasted, embedded in ambiguous background context, or dependent on long-range volumetric cues. Conversely, a task with several labels may require only modest capacity if the structures are spatially stable and visually distinct.

In this framework, label count is only one signal among many, and often not the most important one. Capacity should be understood not only in terms of model size, but also in terms of what the model is being asked to represent: local boundary precision, multiscale contextual reasoning, robustness under acquisition shift, uncertainty tolerance, or some combination of these. Without explicit descriptor-based reasoning, model scale can easily become a surrogate for design confidence rather than a justified response to measured task demands.

\subsection{What Existing Segmentation Systems Already Suggest}

This point is already implicit in several influential segmentation systems. nnU-Net demonstrated that strong performance can be obtained not by maximizing raw model complexity, but by adapting architecture configuration, preprocessing, patching, and training to the fingerprint of the dataset \cite{isensee2021nnunet}. Transformer-based models such as TransUNet, UNETR, Swin-Unet, Swin UNETR, and MISSFormer suggest, in different ways, that some tasks benefit from richer context modeling than conventional convolutional encoders provide \cite{chen2021transunet,hatamizadeh2022unetr,cao2022swinunet,hatamizadeh2022swinunetr,deng2021missformer}. State-space models such as U-Mamba and SegMamba make a related claim through efficient long-range dependency modeling \cite{ma2024umamba,xing2024segmamba}. Together, these works suggest that the relevant design question is not whether larger or more expressive models are preferable in general, but what \emph{type} of capacity is appropriate for the dependency structure and scale complexity of the dataset.

This interpretation is more useful than treating newer model families as evidence of a universal trend toward scale. Their significance is that they expose multiple dimensions of capacity: local detail preservation, long-range interaction, efficient volumetric context modeling, and transferable priors. These dimensions may matter differently depending on the dataset profile.

\subsection{Pretraining as a Conditional Intervention Rather Than a Generic Improvement}

A similar argument applies to \emph{pretraining}. Generic pretraining remains attractive in medical segmentation for several legitimate reasons. It can improve optimization stability, accelerate convergence, and sometimes compensate for limited annotated data. In some tasks, especially when structures are large and appearance statistics are not too distant from the source domain, pretrained encoders can provide a strong starting point. Recent systems such as MedSAM, SAM-Med2D, Medical SAM Adapter, and MedSAM2 also show that transferred segmentation priors can be useful when adapted carefully to the medical setting \cite{ma2024medsam,cheng2023sammed2d,wu2023medicalsamadapter,zhu2024medsam2}.

However, transfer is not a neutral design choice. It brings assumptions about useful features, scale hierarchies, semantic organization, and context mechanisms that may or may not align with the target dataset. For this reason, the relevant question should not be framed simply as ``pretrained or not pretrained''. A more informative dataset-conditioned question is: \emph{pretrained on what, transferred how, and for which measured source of difficulty?} If the central challenge of the dataset is modality mismatch, scanner variation, or high-frequency boundary detail, then generic natural-image pretraining may be less informative than in-domain self-supervised learning or careful training from scratch with stronger spatial inductive bias. If the key issue is long-range anatomical context in a volumetric task, then the usefulness of transfer may depend more on the compatibility of the context mechanism than on the nominal size of the source model. If the dataset is low-cardinality but structurally demanding, then a very large encoder may improve optimization while still being poorly aligned in terms of capacity allocation.

\subsection{Recording and Justifying Transfer in the MS-DKC}

Within the MS-DKC framework, the pretraining strategy should be recorded and justified in the same way as any other design decision. The card should make explicit whether a model uses no pretraining, generic natural-image pretraining, medical-domain supervised pretraining, self-supervised in-domain pretraining, or adaptation from a large segmentation foundation model. More importantly, it should state why that choice is appropriate for the dataset. If pretraining is used primarily to stabilize optimization under a small sample size, that rationale differs from that of using it to import general shape priors, support promptable adaptation, or improve robustness under domain shift.

Making these distinctions explicit helps avoid the widespread but rarely examined assumption that larger transferred priors are always preferable. It also improves reproducibility: two studies using the same pretrained backbone may rely on it for very different reasons, and those reasons should be visible in the design record rather than left implicit.

\subsection{Capacity, Transfer, and Resolution Policy Belong to the Same Design Space}

Capacity and pretraining interact closely with \emph{resolution policy}. In many medical segmentation tasks, increasing model size is not the most effective response to difficulty if the key information is being removed by preprocessing or downsampling. A compact model that preserves native spatial detail may be more appropriate than a much larger model that operates only on heavily reduced inputs. This is particularly relevant in thin-structure and small-target regimes, where the effective bottleneck may be signal preservation rather than abstract representational power.

Accordingly, MS-DKC treats downsampling sensitivity and fine-detail preservation as part of the same design space as capacity. The relevant question is not merely how much capacity a model has, but whether that capacity is allocated to the right resolution and context regime. In some datasets, broader context is the missing ingredient; in others, preserving fine-scale evidence is more important than adding representational depth. Therefore, capacity choice, transfer choice, and resolution policy should be justified together rather than optimized independently.

\subsection{From Design Priors to Evaluation}

This section has a practical implication that capacity selection should be approached as a design prior, rather than a strictly empirical addendum. In a low-cardinality segmentation task with a sparse foreground, modest morphological diversity, and strict deployment constraints, a lighter encoder and narrower bottleneck may be more suitable prior than a large transferred backbone. When combined with robust regularization and uncertainty-aware evaluation, additional capacity may be justified in a heterogeneous multimodal tumor task with uncertain boundaries and strong context dependence. The primary concern in a 3D anisotropic dataset may not be the nominal model size, but rather the distribution of capacity across local and global context pathways. The primary requirement in each example is that the decision can be traced back to measurable attributes of the dataset, rather than architectural familiarity.

In addition, this framework specifies the manner in which empirical discoveries should be assessed. The fact that a pretrained or larger model outperforms a smaller baseline does not necessarily indicate that scale is the optimal design response; the advantage may be due to optimization convenience, superior source priors, or robustness benefits. In contrast, the success of a lighter model in a low-cardinality task does not necessarily suggest that semantic scope is irrelevant; it may suggest that the primary impediment is located elsewhere. Therefore, it is imperative that a dataset-conditioned assessment refrain from making simplistic assumptions, such as ``bigger is better'' or ``pretraining always helps''. The measured demands of the data set are addressed by capacity and transfer, which is the more specific claim.

Evaluation is a direct consequence of this argument. The evaluation must be associated with the risks identified by the dataset if the size, transfer, and resolution policy of the model is chosen accordingly. Therefore, medical image segmentation requires a more comprehensive and risk-aligned evaluation strategy than can be achieved solely through overlap metrics, in addition to a design that is aware of the dataset.
\section{Evaluation Beyond Dice: Toward Risk-Aligned Assessment}

If medical image segmentation is treated as a dataset-conditioned design problem, evaluation must also be treated as a dataset-conditioned choice rather than a fixed reporting routine. In much of the literature, model quality is still reported mainly through overlap-based measures such as Dice, IoU, or related region-similarity scores. These measures continue to be beneficial and are frequently required for comparison; however, they are inadequate to adequately describe numerous hazards that are significant in the context of medical segmentation. A model may achieve a high Dice score while still failing in clinically meaningful ways, such as missing thin branches, fracturing anatomical connectivity, seeping across critical boundaries, producing inadequately calibrated confidence, or degrading under scanner or site shift. In this regard, as with inherited architectural defaults, inherited evaluation defaults can obscure dataset-specific design constraints and failure modes.

This section underscores the necessity of ensuring that the evaluation is in accordance with the risks outlined in the Medical Segmentation Dataset Knowledge Card (MS-DKC). Those properties should be reflected in the evaluation protocol if a dataset contains thin structures, topology-sensitive anatomy, severe class imbalance, annotation ambiguity, or domain shift. The purpose of risk-aligned evaluation is to contextualize existing metrics, rather than to supplant them. Dice continues to be valuable; however, it should be reported in conjunction with metrics that more accurately reflect the task's structural, statistical, and deployment-related risks.

The loop is illustrated in Figure~\ref{fig:msdkc_workflow}, where dataset measurements inform design priors, design priors guide model development, and the resulting system is evaluated using metrics and validation settings that are aligned with the dataset's risks. In this regard, evaluation is not an additional reporting process that is implemented subsequent to model training. It is a component of the identical dataset-conditioned reasoning process.

\subsection{Why Overlap Metrics Alone Are Not Enough}

There are valid explanations why overlap metrics became the standard. They are straightforward, broadly comprehensible, and frequently offer a valuable first-order summary of segmentation quality. Dice and IoU can reasonably accurately represent coarse segmentation performance for numerous region-dominated tasks, including the segmentation of large organs or smooth anatomical compartments. Nevertheless, these measurements are frequently insensitive to a number of critical error types in medical imaging.

Dice may be minimally affected by boundary deviations when the target is considerably larger. Despite being clinically pertinent, the absence of narrow structures may also have a limited impact on overlap. Topological failures, including discontinuous branches, false gaps, or fractured connectivity, may be weakly penalized, even if they diminish downstream utility.. In the same way, average overlap can obscure systematic failure in out-of-distribution subgroups, minority classes, small lesions, or rare structures. This is not merely a reporting limitation in a dataset-conditioned framework; it is a disparity between the risk structure of the task and the evaluation protocol.

\subsection{Evaluation Mismatch as a Dataset-Conditioned Risk}

To clarify this problem, we define \emph{evaluation mismatch} as a dataset-conditioned risk. Evaluation mismatch occurs when the selected metric suite and validation setting do not reflect the primary failure modes of the dataset. A study may appear convincing under standard reporting while remaining poorly aligned with the structural, statistical, or deployment-relevant risks of the task.

This framing shifts the question from ``Which metrics are standard?'' to ``Which metrics are justified by the dataset's measured risks?'' In this respect, evaluation design should be treated in the same way as model design. Just as architecture family, capacity, supervision, and preprocessing should be guided by descriptor-triggered risks, the metric suite and validation strategy should also be selected according to the dataset profile.

\begin{table*}[htbp!]
\centering
\caption{Representative risk-aligned evaluation choices in the MS-DKC framework.}
\label{tab:eval_risk_mapping}
\begin{tabular}{p{2.5cm}p{3.5cm}p{4.9cm}p{5.0cm}}
\toprule
\textbf{Dominant dataset risk} & \textbf{When often present} & \textbf{Evaluation emphasis} & \textbf{MS-DKC descriptors to inspect} \\
\midrule
Boundary and structural error
& Thin, small, branching, or interface-critical targets
& Hausdorff distance, average surface distance, normalized surface Dice, boundary F-score, contour error summaries
& Boundary thinness, topology criticality, target scale distribution \\

Topology and connectivity failure
& Graph-like or continuity-sensitive anatomy
& Connectivity analysis, connected-component statistics, clDice or topology-aware metrics, branch preservation
& Branching complexity, connectedness, topology sensitivity \\

Imbalance and rare-event failure
& Small lesions, rare structures, screening-oriented tasks, sparse foreground
& Class-wise reporting, lesion-level sensitivity, object-level summaries, size-stratified analysis
& Foreground occupancy, rare-label frequency, target prevalence, class imbalance \\

Uncertainty and supervision mismatch
& Boundary ambiguity, inter-rater disagreement, partial or noisy labels
& Inter-rater comparison, calibration, uncertainty-error correlation, agreement-aware analysis
& Inter-rater agreement, boundary ambiguity, supervision quality \\

Domain fragility and distribution shift
& Multi-site, multi-scanner, protocol-heterogeneous, stain- or device-variable datasets
& External validation, subgroup analysis, cross-site testing, calibration under shift
& Acquisition heterogeneity, site variation, domain-shift exposure \\

Operational and deployment mismatch
& Resource-constrained settings, human-in-the-loop workflows, latency-sensitive deployment
& Latency, memory use, robustness, error severity, subgroup consistency, calibration
& Deployment constraints, workflow assumptions, failure severity \\
\bottomrule
\end{tabular}
\end{table*}

\begin{figure*}[t]
    \centering
    \includegraphics[width=\textwidth]{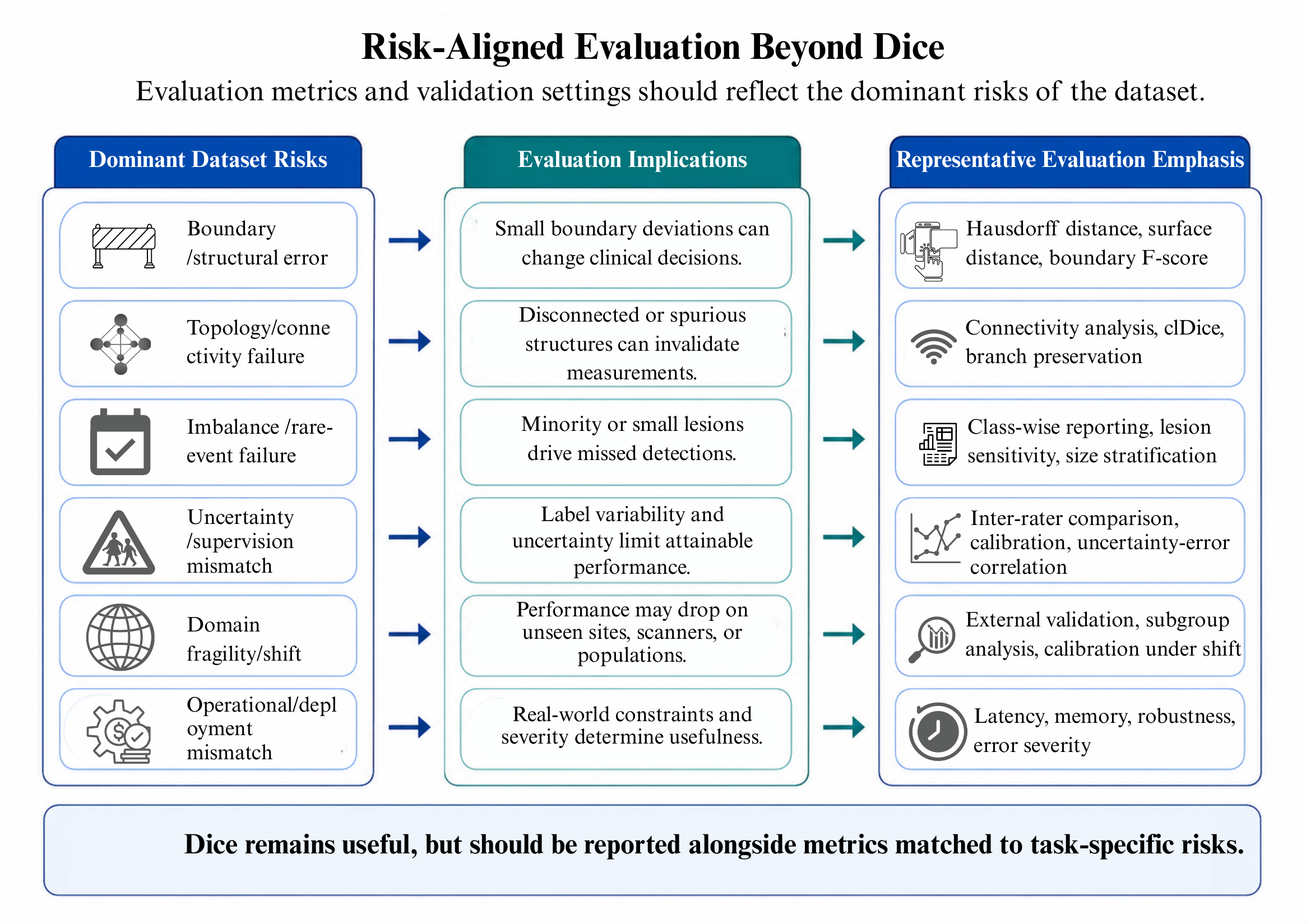}
    \caption{Risk-aligned evaluation beyond Dice. Dominant dataset risks are mapped to evaluation emphases that better reflect structural, statistical, and deployment-related failure modes.}
    \label{fig:risk_aligned_evaluation}
\end{figure*}

Table~\ref{tab:eval_risk_mapping} illustrates how dominant dataset risks can be translated into relevant metric families and validation settings. The goal is not to prescribe a uniform checklist, but to make evaluation decisions traceable to the underlying dataset profile. Figure~\ref{fig:risk_aligned_evaluation} illustrates the central idea of risk-aligned evaluation. Dominant dataset risks should determine which metric families and validation settings are emphasized. Dice remains useful, but it should be interpreted together with boundary, topology, uncertainty, subgroup, external-validation, and operational measures when the dataset profile indicates these risks.

\subsection{Boundary- and Topology-Sensitive Assessment}

One direct use of risk-aligned evaluation is in tasks involving thin structures, branching morphologies, or small anatomical interfaces. In such cases, boundary accuracy and connectivity may be as important as region overlap. This includes vessel segmentation, duct segmentation, membrane delineation, tumor margin analysis, and several pathology segmentation tasks. For these tasks, overlap measures should be supplemented by boundary-sensitive distances and topology-aware evaluations.

Boundary-oriented measures such as Hausdorff distance, average surface distance, normalized surface Dice, boundary F-score, and contour-based error summaries can provide a clearer picture of whether the model preserves clinically important interfaces. Similarly, topology-sensitive measures, connectivity analysis, branch preservation statistics, and connected-component studies become important when the target structure is graph-like or when broken continuity affects clinical interpretation. The general principle is that evaluation should respond directly to the \emph{morphological descriptors} recorded in the MS-DKC. If topology sensitivity is high, topology-aware evaluation should not be optional. If boundary thinness is central to the task, region overlap alone is inadequate.

\subsection{Imbalance, Rare Structures, and Heterogeneous Failure}

Another common evaluation problem occurs in highly imbalanced tasks. When foreground occupancy is very small, average overlap scores can be dominated by easy cases or by background-rich images in which the target is absent or only weakly represented. This is particularly problematic in lesion segmentation, small-structure delineation, and screening-oriented workflows, where clinically relevant events may be rare. Under these conditions, aggregate metrics should be supplemented by class-wise reporting, foreground-conditional reporting, lesion-level sensitivity, object-level summaries, or stratification by target size and prevalence.

The same principle applies to case-wise and slice-wise heterogeneity. In volumetric segmentation, some slices may be straightforward, while others are anatomically complex or uncertainty-dominated. Dataset-level averages may therefore conceal severe local failures. A risk-aligned evaluation protocol should include stratification across easy and hard subsets, foreground size bins, anatomical subregions, or uncertainty strata whenever the dataset profile indicates that these factors are clinically meaningful. The role of MS-DKC is to identify which forms of stratification are likely to matter.

\subsection{Annotation Uncertainty, Agreement, and Calibration}

Medical segmentation labels are often imperfect. Boundaries may be ambiguous, raters may disagree, and some tasks may admit multiple clinically acceptable delineations. In such settings, evaluating the model as though the annotation were a single perfect ground truth can create a false sense of certainty. This is especially problematic when systems are later used in settings that require confidence-based decision making or clinical review.

For this reason, the MS-DKC supervision layer should influence not only model design but also evaluation design. When annotation uncertainty is high, model performance should be interpreted relative to inter-rater variability where possible, and confidence or uncertainty estimates should be assessed explicitly. Calibration metrics, uncertainty-error correlations, risk-coverage analysis, and agreement-aware analyses may all be useful depending on the task. The essential principle is that label uncertainty should not be hidden during evaluation. If the dataset indicates uncertain supervision, the evaluation protocol should represent this ambiguity rather than compress it into a single deterministic score.

\subsection{External Validation and Distribution Shift}

A common limitation of segmentation studies is that evaluation is often restricted to in-distribution test splits closely related to the training data. However, acquisition and site heterogeneity are central descriptors in MS-DKC because segmentation models may degrade when scanner settings, reconstruction protocols, institutions, staining procedures, or patient populations vary. A model that performs well on an internal test set but fails under cross-site shift may still be unsuitable for practical use, especially in clinical settings where robustness is essential.

External validation should therefore be viewed as a risk-driven requirement rather than an optional addition when the dataset profile indicates domain-shift exposure. If the MS-DKC profile suggests multi-site aggregation, scanner diversity, substantial protocol dependence, or likely out-of-distribution deployment, subgroup analysis and external validation become central components of evaluation. Calibration should also be reassessed under shift whenever predictions are likely to influence downstream decisions. In this framework, the distribution shift is not merely a limitation that should be mentioned after reporting the results; it is a measurable deployment risk that should shape the evaluation from the beginning.

\subsection{Operational Relevance and Deployment Constraints}

Evaluation in medical segmentation should also take into account operational constraints. In some settings, inference time, memory footprint, and edge-deployment feasibility may be as important as raw segmentation accuracy. In others, the main concern may be whether the model supports human-in-the-loop review through uncertainty estimates, controlled false-negative behavior, or robust performance under limited computation. These issues are often overlooked in benchmark-style reporting, even though they may be central to the intended use of the system.

The deployment-risk layer of MS-DKC therefore extends the evaluation beyond predictive accuracy. It encourages reporting not only segmentation metrics but also latency, memory use, calibration, robustness, subgroup consistency, and the practical severity of different error types. This is especially important when comparing models with very different scales. A modest improvement in average Dice may not justify a large increase in computational cost if the deployment setting is constrained, whereas a lighter model may be preferable if it provides greater stability, calibration, or practical usability in the target environment.

\subsection{Toward a Justified Evaluation Protocol}

The key point of this section is that the evaluation should be justified in terms of the dataset profile. A dataset with smooth, large, and well-separated organs can appropriately prioritize overlap and surface distance. A vessel dataset should emphasize boundary quality and connectivity. A lesion dataset should prioritize sensitivity under imbalance and size stratification. A multi-site MRI dataset should emphasize external validation and calibration under shifts. A dataset with uncertain or inconsistent labels should include analysis that is aware of agreement or uncertainty. Evaluation should therefore be treated as an extension of dataset knowledge rather than as a fixed appendix to model training.

This does not mean that every study must report every possible metric. Rather, the evaluation suite should be \emph{justified}. MS-DKC helps make this reasoning explicit by linking each key metric family to a measured source of task risk. In practice, this can be implemented through a concise evaluation card or table that states which dataset descriptors motivated which evaluation choices. Such reporting would improve interpretability and comparability by allowing readers to understand not only what scores are reported, but also why those scores are meaningful for the task.

\subsection{From Risk-Aligned Evaluation to Segmentation Regimes}

The main limitation of current medical segmentation evaluation is not that standard metrics are useless, but that they are too often used without reference to the measured structure and risk profile of the dataset. A dataset-conditioned segmentation framework therefore requires a correspondingly dataset-conditioned evaluation framework. Once model design is linked to acquisition variability, target morphology, supervision uncertainty, and deployment risk, evaluation must be linked to those same properties. This closes the MS-DKC loop: dataset measurements motivate design priors, those priors guide model development, and the resulting system is evaluated using metrics and validation settings aligned with the risks the dataset presents.

The next section makes this perspective concrete across representative segmentation regimes. These examples illustrate how different medical segmentation settings can produce different MS-DKC profiles and therefore different justified choices in architecture, capacity, supervision, and evaluation.


The central claim of this paper is that medical image segmentation should be approached as a dataset-conditioned design problem. To make this claim concrete, we now consider several representative segmentation regimes in which the Medical Segmentation Dataset Knowledge Card (MS-DKC) would induce meaningfully different design priors. These examples are not intended as exhaustive benchmarks, nor do they imply that a single architecture is universally best within any one domain. Their purpose is instead to show how measurable differences in acquisition, target structure, supervision, context dependence, and deployment risk translate into different justified choices in architecture, capacity, preprocessing, supervision, and evaluation.

A useful role of case studies in this framework is that they shift the discussion away from abstract model-family comparison and toward design reasoning in context. A vessel-segmentation task, a volumetric organ-segmentation task, a brain-tumor segmentation problem, and a nuclei-segmentation problem may all fall under the broad heading of medical image segmentation, yet they differ substantially in target morphology, annotation uncertainty, dependency structure, and operational risk. The value of the MS-DKC lies precisely in making these differences explicit. In what follows, each case study is therefore organized in the same way: a dataset profile, the dominant risks implied by that profile, and the design priors that follow from those risks.

\section{Experimental Results and Dataset-Conditioned Adaptation Analysis}
\label{sec:experimental_results}

This section evaluates the proposed Medical Segmentation Dataset Knowledge Card (MS-DKC) framework across three representative medical image segmentation regimes. DRIVE is treated as the primary experimental dataset because it exposes the strongest dataset-conditioned design problem in this study: sparse foreground, thin branching structures, topology preservation, strong threshold dependence, and foreground--background imbalance. ISIC and ACDC are retained as comparative experiments to demonstrate that the preferred architecture changes when the dataset profile changes. ISIC is dominated by compact lesion regions with heterogeneous appearance and fuzzy boundaries, whereas ACDC contains larger, coherent cardiac structures where standard U-Net-style region modeling remains highly effective.

The experimental narrative follows the MS-DKC workflow introduced earlier in the paper. First, the dataset is profiled through measurable descriptors. Second, the descriptors are translated into expected failure modes and design risks. Third, model interventions are selected according to those risks. Fourth, models are evaluated using risk-aligned metrics rather than Dice and IoU alone. The goal is therefore not to claim that a single module is universally beneficial, but to show how dataset knowledge explains why a model improves some metrics while degrading others.

\subsection{Experimental Protocol and Reporting Strategy}
\label{subsec:experimental_protocol}

All experiments were organized to test dataset-conditioned design rather than architecture size alone. The study should therefore be read as an MS-DKC adaptation and evaluation study, not as a claim that we propose one universal segmentation backbone. Within each dataset-specific block, models were compared under matched preprocessing, input resolution, optimizer settings, and evaluation code as far as possible. When augmentation was used, it is explicitly stated. In the DRIVE final setting, controlled RotFlip500 augmentation was used. In the latest ISIC AttNextTopo diagnostic branch, no augmentation was used, so that the effect of the backbone, loss, threshold-selection rule, and diagnostic modules could be inspected without an additional augmentation variable.

DRIVE was selected as the primary full-scale case study because it is the most difficult dataset-conditioned setting in this work. It contains only 20 official training images and 20 official test images, and the target foreground is sparse, thin, branching, and topology-sensitive. For this reason, DRIVE receives the most extensive experimental treatment: compact model design, adaptation of an existing vessel-specific backbone, threshold studies, topology-aware evaluation, ablation analysis, and final three-seed testing where available. The DRIVE experiments should not be read as claiming that DKC-TNet-v2 or SA-UNetv2-DKC-AmbRef is a universal state-of-the-art architecture. Rather, they show how MS-DKC can take existing model directions--a compact T-Net-style design and the SA-UNetv2 vessel backbone--and improve their useful operating behavior under the measured risks of DRIVE.

The DRIVE experiments include two stages. The first is a development-stage analysis under a controlled $512 \times 512$ Fast1000 setting. These results include the compact DKC-TNet branch and early SA-UNetv2 adaptation variants such as Dice/MCC/BCE calibration and ambiguity-aware refinement. These experiments are retained because they document the MS-DKC development path. The second stage is the final reproducibility setting using $592 \times 592$ RGB images, RotFlip500 augmentation, no validation split, and three seeds. The final DRIVE model selection is based on this second setting, with seeds 42, 123, and 2025. Unless otherwise stated, final DRIVE results are reported at a fixed threshold of 0.65. Threshold sweeps are interpreted as operating-point analyses rather than as test-set model selection.

ISIC and ACDC are used as supporting diagnostic datasets. Their purpose is to demonstrate that the same MS-DKC workflow does not recommend the same intervention for every medical segmentation problem. ISIC is a compact lesion segmentation task dominated by scale variation, fuzzy boundaries, and appearance variation rather than vessel-like topology. ACDC contains compact but anatomically related cardiac structures and must be treated as a four-class softmax segmentation task. For ISIC, all rows are treated as controlled single-run diagnostic tests under the reported protocol. The ISIC module experiments are used to test which interventions are retained or rejected by the ISIC knowledge card, not to claim a universal final model.

Because the evaluated models are run under our controlled dataset splits, image sizes, optimizer choices, augmentation settings, threshold-selection procedures, hardware environment, and metric code, their numbers may differ from the original papers. This is expected. Unless explicitly stated, the goal is not to reproduce each published paper's full training recipe. The goal is to compare how model families and MS-DKC-guided interventions behave under a transparent, dataset-conditioned experimental protocol.

For DRIVE, we report Dice, IoU, accuracy, sensitivity, specificity, precision, MCC, NPV, clDice, skeleton precision, skeleton recall, HD95, ASSD, Boundary F1, vessel-fraction error, AUC, and AP. This expanded metric panel is required because DRIVE is not fully characterized by region overlap. A model may improve Dice by suppressing false positives while still missing thin terminal vessels. Conversely, a model may improve sensitivity and skeleton recall while producing false-positive overgrowth that reduces precision and IoU. The MS-DKC framework therefore requires that overlap, topology, boundary, and calibration metrics be interpreted together.

\subsection{Dataset Knowledge Profiles Across DRIVE, ISIC, and ACDC}
\label{subsec:dataset_profiles}

Table~\ref{tab:dataset_profiles} summarizes the high-level MS-DKC profiles used in this study. These profiles explain why the strongest model family differs across datasets. DRIVE requires vessel-specific priors and topology-aware evaluation, ISIC favors robust appearance modeling for compact lesions, and ACDC favors coherent region modeling for cardiac structures.

\begin{table*}[t]
\centering
\caption{High-level MS-DKC profiles for the three datasets. DRIVE is the primary dataset because it has the strongest thin-structure and topology-preservation requirements.}
\label{tab:dataset_profiles}
\small
\begin{adjustbox}{width=\textwidth}
\begin{tabular}{p{2.2cm} p{4.0cm} p{4.2cm} p{4.8cm}}
\toprule
\textbf{Dataset} & \textbf{Target profile} & \textbf{Dominant risks} & \textbf{Design implication} \\
\midrule
DRIVE & Sparse, thin, branching retinal vessels & Missed terminal vessels, connectivity loss, threshold sensitivity, foreground imbalance & Detail-preserving vessel models, imbalance-aware objectives, topology-aware metrics, threshold analysis \\
ISIC & Compact skin-lesion regions with heterogeneous texture and fuzzy boundaries & Appearance variation, boundary ambiguity, lesion-size variation & Strong appearance modeling, robust boundary handling, region-overlap metrics with boundary analysis \\
ACDC & Larger coherent cardiac anatomical structures & Shape consistency, multi-class anatomical coherence, inter-slice variability & Region-aware encoder--decoder models and structure-consistent segmentation \\
\bottomrule
\end{tabular}
\end{adjustbox}
\end{table*}

\subsection{DRIVE as the Primary MS-DKC Case Study}
\label{subsec:drive_main_case}

DRIVE is the main experimental dataset in this paper because it is the most direct example of a dataset where design decisions must be conditioned on target morphology and evaluation risk. Retinal vessels occupy a small fraction of the fundus image, but they extend across most of the field of view as thin, branching structures. This creates a difficult segmentation regime: foreground imbalance encourages background-biased predictions, thin vessels are vulnerable to downsampling and thresholding, and the vascular tree can be clinically degraded by small topological breaks even when average overlap remains high.

The MS-DKC diagnosis for DRIVE is therefore different from the diagnosis for ISIC or ACDC. For ISIC, the dominant difficulty is lesion appearance and boundary ambiguity. For ACDC, the target regions are compact anatomical cardiac structures with strong class-wise shape constraints and a boundary-sensitive myocardium. In DRIVE, however, the main risks are thin-vessel under-recovery, false-positive overgrowth, topology loss, and threshold instability. These risks explain why a model can perform well on precision but poorly on sensitivity, or improve Dice while reducing clDice. The purpose of the DRIVE experiments is to make these trade-offs explicit and to show how MS-DKC guides the search for a better model.

\subsubsection{Quantitative DRIVE Knowledge Card}
\label{subsubsec:drive_quant_profile}

Table~\ref{tab:drive_quant_msdkc} reports the quantitative dataset descriptors used to construct the DRIVE MS-DKC. The official DRIVE split contains only 20 training and 20 test images. The field-of-view fraction is approximately 0.688, and vessel occupancy within the FOV is approximately 12.54\%. The background-to-foreground ratio is therefore approximately 7.15. The estimated median vessel width is only 2.87 pixels, and the branchpoint density is 72.84 per 1000 skeleton pixels. These values support the diagnosis that DRIVE is a sparse, thin-structure, topology-sensitive dataset.

\begin{table*}[t]
\centering
\caption{Quantitative MS-DKC descriptors for DRIVE. These descriptors motivate vessel-specific metrics and dataset-conditioned interventions.}
\label{tab:drive_quant_msdkc}
\small
\begin{adjustbox}{width=\textwidth}
\begin{tabular}{p{5.2cm} p{3.0cm} p{6.0cm}}
\toprule
\textbf{Descriptor} & \textbf{Measured value} & \textbf{Interpretation} \\
\midrule
Training/test split & 20/20 images & Small sample size; seed stability is important. \\
Original image grid & Approximately $565 \times 584$ & Moderate-resolution fundus images. \\
FOV fraction & 0.688 & Evaluation and sampling should be FOV-aware. \\
Foreground vessel occupancy in FOV & 12.54\% & Strong foreground imbalance. \\
Background-to-foreground ratio & 7.15 & Accuracy alone is not informative. \\
Estimated median vessel width & 2.87 px & Thin structures are vulnerable to downsampling. \\
Boundary-to-area ratio & 0.511 & Boundary errors strongly affect segmentation quality. \\
Skeleton-to-area ratio & 0.301 & Centerline preservation is important. \\
Branchpoint density & 72.84 per 1000 skeleton pixels & High topology and connectivity dependence. \\
Vessel-tree bounding-box fraction & 0.854 & Vessels are spatially distributed across most of the retinal FOV. \\
\bottomrule
\end{tabular}
\end{adjustbox}
\end{table*}

These measurements justify the final evaluation panel. Dice and IoU are retained because they summarize overlap, but they are not sufficient. Sensitivity and precision are needed to distinguish vessel recovery from false-positive overgrowth. clDice and skeleton recall are needed because the target is centerline-dominated. HD95, ASSD, and Boundary F1 are needed because vessels have a high boundary-to-area ratio. AUC and AP are needed because threshold behavior strongly affects final masks.

\subsubsection{Risk-to-Intervention Mapping}
\label{subsubsec:drive_risk_mapping}

The MS-DKC profile was used to convert dataset measurements into testable design hypotheses. Table~\ref{tab:drive_risk_design_mapping} summarizes this mapping. Importantly, the mapping does not imply that every intervention should be kept. Instead, each intervention is treated as a hypothesis that must improve the appropriate risk-aligned metrics.

\begin{table*}[t]
\centering
\caption{Risk-to-intervention mapping derived from the DRIVE MS-DKC profile. MS-DKC is used to design compact models and to adapt an existing vessel-specific backbone.}
\label{tab:drive_risk_design_mapping}
\small
\begin{adjustbox}{width=\textwidth}
\begin{tabular}{p{3.1cm} p{2.0cm} p{4.0cm} p{4.7cm} p{3.2cm}}
\toprule
\textbf{MS-DKC risk} & \textbf{Risk level} & \textbf{Evidence from DRIVE profile} & \textbf{MS-DKC intervention} & \textbf{Evaluation metrics} \\
\midrule
Structural information loss & High & Median vessel width is 2.87 px; boundary-to-area ratio is 0.511. & Use detail-preserving architectures and avoid excessive downsampling or over-suppression of skip features. & Dice, IoU, sensitivity, clDice \\
Topology/connectivity failure & High & Branchpoint density is 72.84 per 1000 skeleton pixels; skeleton-to-area ratio is 0.301. & Report topology-aware metrics and test whether refinement modules improve component-level consistency. & clDice, skeleton recall, connectivity error \\
Foreground imbalance & High & Vessel occupancy is only 12.54\% within the FOV. & Use imbalance-aware objectives such as BCE--MCC and report precision--recall behavior. & Sensitivity, specificity, precision, MCC \\
Context mismatch & Moderate & The vessel tree extends over a large fraction of the FOV. & Use multiscale context and skip-based feature preservation. & Dice, IoU, HD95, ASSD \\
Evaluation mismatch & High & Dice may hide branch breaks and missed terminal vessels. & Report extended vessel-specific metrics in addition to Dice/IoU. & clDice, skeleton precision/recall, Boundary F1, AUC, AP \\
Threshold sensitivity & High & Probability threshold changes the balance between vessel recovery and false positives. & Report fixed-threshold results and threshold-sweep analysis. & Dice, sensitivity, precision, clDice \\
\bottomrule
\end{tabular}
\end{adjustbox}
\end{table*}

This mapping guided two experimental branches. The first branch investigated compact T-Net-style models and asked whether MS-DKC can improve a small vessel model. The second branch adapted SAUNetv2, a strong vessel-specific backbone, through green/gray priors and lightweight feature contrastive supervision. The final model selection came from the second branch, but the compact branch is retained because it demonstrates the diagnostic value of the framework.

\subsection{Development Branch I: Compact DKC-TNet Models}
\label{subsec:drive_tnet_branch}

The compact T-Net branch was designed to test whether the DRIVE knowledge card could improve a resource-constrained model. The baseline T-Net-style model was efficient but conservative: it had high specificity but missed thin vessels and showed limited topology recovery. This behavior is exactly what the DRIVE MS-DKC predicts for a compact model with limited high-resolution support.

Before stating the DRIVE objectives, we define the loss components used in the development branches. Let $p_i\in[0,1]$ be the predicted foreground probability at pixel $i$, $g_i\in\{0,1\}$ the ground-truth label, $N$ the number of pixels, and $\epsilon$ a small constant. The standard soft Dice loss is
\begin{equation}
\mathcal{L}_{\mathrm{Dice}}
=1-
\frac{2\sum_i p_i g_i+\epsilon}{\sum_i p_i+\sum_i g_i+\epsilon}.
\label{eq:loss_dice}
\end{equation}
The L2-denominator Dice loss used in the compact T-Net branch is
\begin{equation}
\mathcal{L}_{\mathrm{Dice\mbox{-}L2}}
=1-
\frac{2\sum_i p_i g_i+\epsilon}{\sum_i p_i^2+\sum_i g_i^2+\epsilon}.
\label{eq:loss_dice_l2}
\end{equation}
The binary cross-entropy and weighted binary cross-entropy terms are
\begin{equation}
\mathcal{L}_{\mathrm{BCE}}
=-\frac{1}{N}\sum_i
\left[g_i\log p_i+(1-g_i)\log(1-p_i)\right],
\label{eq:loss_bce}
\end{equation}
\begin{equation}
\begin{aligned}
\mathcal{L}_{\mathrm{WBCE}}
&=-\frac{1}{N}\sum_i
\left[w_+g_i\log p_i+(1-g_i)\log(1-p_i)\right],\\
\mathcal{L}_{\mathrm{WBCE}}&=\mathcal{L}_{\mathrm{BCE}}
\quad\mathrm{when}\quad w_+=1 .
\end{aligned}
\label{eq:loss_wbce}
\end{equation}
For the soft MCC term, we use
\begin{equation}
\begin{aligned}
\mathrm{TP}&=\sum_i p_i g_i,
&\mathrm{FP}&=\sum_i p_i(1-g_i),\\
\mathrm{FN}&=\sum_i(1-p_i)g_i,
&\mathrm{TN}&=\sum_i(1-p_i)(1-g_i).
\end{aligned}
\label{eq:soft_confusion}
\end{equation}
\begin{equation}
\begin{aligned}
\mathcal{L}_{\mathrm{MCC}}&=1-\frac{A}{B+\epsilon},\\
A&=\mathrm{TP}\cdot\mathrm{TN}-\mathrm{FP}\cdot\mathrm{FN},\\
B&=\sqrt{(\mathrm{TP}+\mathrm{FP})(\mathrm{TP}+\mathrm{FN})(\mathrm{TN}+\mathrm{FP})(\mathrm{TN}+\mathrm{FN})}.
\end{aligned}
\label{eq:loss_mcc}
\end{equation}
Let $S_p$ and $S_g$ denote soft skeletons of the prediction and ground truth. The centerline loss is written as
\begin{equation}
\begin{aligned}
T_{\mathrm{prec}}&=
\frac{\sum_i (S_p)_i g_i+\epsilon}{\sum_i (S_p)_i+\epsilon},\\
T_{\mathrm{sens}}&=
\frac{\sum_i (S_g)_i p_i+\epsilon}{\sum_i (S_g)_i+\epsilon},\\
\mathcal{L}_{\mathrm{center}}
&=1-\frac{2T_{\mathrm{prec}}T_{\mathrm{sens}}}{T_{\mathrm{prec}}+T_{\mathrm{sens}}+\epsilon}.
\end{aligned}
\label{eq:loss_center}
\end{equation}
For the label-guided pixel-wise contrastive term, let $z_i$ be L2-normalized encoder pixel embeddings sampled from a set $V$, with labels $y_i\in\{\mathrm{vessel},\mathrm{bg}\}$. Let $P(i)=\{p\in V:y_p=y_i,\ p\ne i\}$ be the same-label positives and $A(i)=V\setminus\{i\}$ the contrast set. The supervised contrastive term is
\begin{equation}
\begin{aligned}
\mathcal{L}_{\mathrm{cont}}
&=\sum_{i\in V}-\frac{1}{|P(i)|}\sum_{p\in P(i)}\ell_{i,p},\\
\ell_{i,p}
&=\log
\frac{\exp(z_i\cdot z_p/\tau)}
{\sum_{a\in A(i)}\exp(z_i\cdot z_a/\tau)} .
\end{aligned}
\label{eq:loss_contrastive}
\end{equation}

DKC-TNet-v2 DiceRecovery-Calibrated used a channel configuration of $\{8,16,24\}$ and approximately 35.1K trainable parameters. The design used a compact multiscale context, vessel-preserving skip refinement, and a calibrated Dice-dominant objective. The final compact loss was
\begin{equation}
    \mathcal{L}_{\mathrm{DKC\text{-}TNet}}
    =
    0.97\mathcal{L}_{\mathrm{Dice\text{-}L2}}
    +
    0.03\mathcal{L}_{\mathrm{WBCE}}.
\end{equation}
The L2 Dice term was chosen because the DRIVE vessels are sparse and thin; the squared denominator reduces the instability of the foreground imbalance while still encouraging the recovery of overlap.

Table~\ref{tab:drive_ablation} summarizes the compact DKC-TNet ablation sequence. Stronger BCE pressure increased sensitivity, but also introduced false positives and reduced Dice. The threshold adjustment recovered specificity but reduced sensitivity. The final DiceRecovery-Calibrated configuration achieved the best compact result, with Dice 0.8044 and IoU 0.6730.

\begin{table*}[t]
\centering
\caption{DRIVE ablation study of compact DKC-TNet variants. Main results use $512 \times 512$ input and Fast1000 augmentation unless otherwise stated.}
\label{tab:drive_ablation}
\small
\begin{adjustbox}{width=\textwidth}
\begin{tabular}{lccccccp{4.8cm}}
\toprule
\textbf{Variant} & \textbf{TH} & \textbf{Dice} & \textbf{IoU} & \textbf{Sen.} & \textbf{Sp.} & \textbf{clDice} & \textbf{Finding} \\
\midrule
Corrected T-Net $\{8,16,24\}$ & 0.50 & 0.7941 & 0.6590 & 0.7851 & 0.9821 & 0.7995 & Compact baseline with high specificity but insufficient vessel recall. \\
DKC-TNet-v2, 0.90 Dice + 0.10 BCE & 0.50 & 0.7914 & 0.6550 & \textbf{0.8358} & 0.9739 & 0.8163 & Stronger BCE increased sensitivity but introduced false positives and reduced Dice. \\
DKC-TNet-v2, 0.90 Dice + 0.10 BCE & 0.60 & 0.7943 & 0.6591 & 0.7989 & 0.9801 & 0.8163 & Raising threshold recovered specificity but reduced sensitivity. \\
DKC-TNet-v2 DiceRecovery, 0.95 Dice + 0.05 BCE & 0.50 & 0.7971 & 0.6629 & 0.8233 & 0.9773 & 0.8127 & Reduced BCE pressure improved Dice while preserving useful recall. \\
DKC-TNet-v3-Lite MSRGB + AmbRef & 0.55 & 0.7976 & 0.6637 & 0.8046 & 0.9802 & 0.8164 & Added ambiguity refinement but did not surpass final calibrated T-Net. \\
\textbf{DKC-TNet-v2 DiceRecovery-Calibrated} & \textbf{0.50} & \textbf{0.8044} & \textbf{0.6730} & 0.8241 & \textbf{0.9789} & \textbf{0.8168} & Best compact development-stage DKC-TNet configuration. \\
\bottomrule
\end{tabular}
\end{adjustbox}
\end{table*}

The compact branch was then extended with GreenRecall-TNet, GreenRecall-TNet-v2C, and GreenRecall-CSA-TNet-v3 in the $592 \times 592$ RotFlip500 setting. These models tested whether green-channel prior information, recall-weighted topology support, and residual skip attention could improve a very small model. Table~\ref{tab:drive_tnet_late_summary} summarizes the most important diagnostic results. GreenRecall improved vessel recovery but risked overgrowth. The calibrated v2C variant improved the overlap and vessel-fraction behavior. CSA-Lite skip attention gave the best compact balance by improving high-resolution skip selection without introducing a strong recall bias.

\begin{table*}[t]
\centering
\caption{Later compact T-Net diagnostic branch under the $592 \times 592$ RotFlip500 setting. These experiments show how MS-DKC corrected under-recovery, overgrowth, and skip-fusion limitations in a small model.}
\label{tab:drive_tnet_late_summary}
\small
\begin{adjustbox}{width=\textwidth}
\begin{tabular}{lccccccp{5.0cm}}
\toprule
\textbf{Model} & \textbf{TH} & \textbf{Dice} & \textbf{IoU} & \textbf{Sen.} & \textbf{Prec.} & \textbf{clDice} & \textbf{Interpretation} \\
\midrule
T-Net baseline & 0.45 & 0.8059 & 0.6752 & 0.8206 & 0.7978 & 0.8147 & Compact and efficient, but still limited by thin-vessel and topology recovery. \\
GreenRecall-TNet & 0.45 & 0.8097 & 0.6806 & 0.8456 & 0.7804 & 0.8272 & DKC recall correction improved topology and sensitivity, but increased over-prediction. \\
GreenRecall-TNet-v2C & 0.575 & 0.8133 & 0.6856 & 0.8196 & 0.8110 & 0.8216 & Overgrowth control improved overlap and calibration but reduced some topology recovery. \\
GreenRecall-CSA-TNet-v3 & 0.575 & \textbf{0.8142} & \textbf{0.6869} & 0.8195 & \textbf{0.8124} & 0.8240 & Residual CSA skip attention provided the best compact balance. \\
AxBiLSTM-TNet-v4 & 0.575 & 0.8122 & 0.6841 & \textbf{0.8250} & 0.8053 & 0.8207 & Increased recall slightly but reduced precision, surface quality, and topology compared with CSA. \\
VCSF-TNet-v4C & 0.575 & 0.8123 & 0.6842 & 0.8197 & 0.8088 & 0.8209 & Vessel-specific gating was too restrictive and slower than CSA. \\
\bottomrule
\end{tabular}
\end{adjustbox}
\end{table*}

The T-Net branch demonstrates a key MS-DKC lesson: the best intervention is not necessarily the most complex. BiLSTM added directional context, but did not align well with the curved vessel geometry. VCSF attempted vessel-calibrated skip fusion, but required too many gates to agree, suppressing faint vessels. CSA worked better because it preserved high-resolution skip evidence through a simple residual mechanism. However, even the best compact variant remained below the stronger SAUNetv2 branch; therefore, T-Net is reported as diagnostic support rather than as the final DRIVE model.

\subsection{Development Branch II: SA-UNetv2 Adaptation}
\label{subsec:drive_saunet_development}

SA-UNetv2 was selected as the stronger vessel-specific backbone because it already contains several DRIVE-appropriate priors, including cross-scale spatial attention, Group Normalization, DropBlock regularization, and a BCE--MCC objective. The MS-DKC question was therefore not whether to replace SA-UNetv2 with a larger generic model, but whether its failure modes could be further improved through dataset-conditioned adaptation.

The original-style SA-UNetv2 objective uses BCE and MCC:
\begin{equation}
    \mathcal{L}_{\mathrm{BCE+MCC}}
    =
    \lambda_{\mathrm{BCE}}\mathcal{L}_{\mathrm{BCE}}
    +
    \lambda_{\mathrm{MCC}}\mathcal{L}_{\mathrm{MCC}},
    \qquad
    \lambda_{\mathrm{BCE}}=\lambda_{\mathrm{MCC}}=0.5.
\end{equation}
MCC is appropriate for DRIVE because it accounts for true positives, true negatives, false positives, and false negatives under a strong class imbalance. The Development-stage experiments tested Dice recovery, MCC calibration, residual CSA, and ambiguity-aware refinement. Table~\ref{tab:saunetv2_ablation} preserves these results. SA-UNetv2-DKC-AmbRef was the best development-stage configuration under the $512 \times 512$ Fast1000 protocol, but it is not the final model. The paper remains as historical evidence that MS-DKC-guided adaptation can improve an existing vessel backbone.

\begin{table*}[t]
\centering
\caption{MS-DKC-guided development-stage adaptation study of SA-UNetv2 on DRIVE in the $512 \times 512$ Fast1000 setting. These results are retained as historical development evidence; the final model is selected from the later $592 \times 592$ RotFlip500 three-seed experiments.}
\label{tab:saunetv2_ablation}
\small
\begin{adjustbox}{width=\textwidth}
\begin{tabular}{lccccccp{5.0cm}}
\toprule
\textbf{Variant} & \textbf{TH} & \textbf{Dice} & \textbf{IoU} & \textbf{Sen.} & \textbf{Sp.} & \textbf{clDice} & \textbf{Finding} \\
\midrule
SA-UNetv2 + DiceRecovery & 0.60 & 0.8107 & 0.6819 & 0.8252 & 0.9803 & \textbf{0.8284} & Strong sensitivity and clDice, but high connectivity error. \\
SA-UNetv2 + BCE/MCC & 0.55 & 0.8125 & 0.6844 & 0.8200 & \textbf{0.9813} & 0.8229 & MCC improves imbalance handling and reduces connectivity error. \\
SA-UNetv2 + Dice/MCC/BCE + threshold-aware selection & 0.55 & 0.8139 & 0.6863 & 0.8206 & 0.9812 & 0.8226 & Combines overlap, imbalance, and calibration; improves Dice and connectivity. \\
SA-UNetv2 + Residual CSA & 0.60 & 0.8134 & 0.6855 & 0.8195 & 0.9812 & 0.8254 & Improves clDice but worsens the final Dice/connectivity balance; not retained. \\
\textbf{SA-UNetv2 + AmbRef} & \textbf{0.55} & \textbf{0.8141} & \textbf{0.6865} & \textbf{0.8265} & 0.9804 & 0.8200 & Earlier best development-stage adaptation; improved connectivity and overlap in the Fast1000 setting. \\
\bottomrule
\end{tabular}
\end{adjustbox}
\end{table*}

The SA-UNetv2 results in the development-stage show that DRIVE does not reward arbitrarily complex. Residual CSA improved clDice but did not produce the best overall balance. Ambiguity-aware refinement improved connectivity in the development setting, but later three-seed experiments showed that the most reproducible final improvement came from a simpler and more targeted combination: GreenTopo prior plus AttUKANLite feature discrimination.

\subsection{Final DRIVE Model: GreenTopo-AttUKANLite-SAUNetv2}
\label{subsec:drive_final_model}

The final model selected for DRIVE is GreenTopo-AttUKANLite-SAUNetv2. Its design follows directly from the DRIVE knowledge card. The green anterior branch is motivated by the contrast of the retinal vessels, as the vessels are typically more visible in the green channel. The AttUKANLite contrastive term is motivated by vessel/background ambiguity and the small training set; it encourages feature-level separation between vessel and background without dominating the segmentation objective. The final objective is
\begin{equation}
\begin{split}
\mathcal{L}_{\mathrm{AttUKANLite}} ={}&
0.45\mathcal{L}_{\mathrm{BCE}}
+ 0.45\mathcal{L}_{\mathrm{MCC}}
+ 0.05\mathcal{L}_{\mathrm{Dice}} \\
&+ 0.005\mathcal{L}_{\mathrm{cont}}
+ 0.01\mathcal{L}_{\mathrm{center}} .
\end{split}
\end{equation}
The weights intentionally keep BCE--MCC as the dominant objective. The Dice term provides weak overlap support, the contrastive term improves feature discrimination, and the centerline term provides weak topology support. This weighting was selected because stronger recall or topology losses repeatedly produced overgrowth or reduced Dice/IoU in earlier experiments.

Three final SAUNetv2-family models were evaluated under the $592 \times 592$ RotFlip500 three-seed setting: Basic SAUNetv2, GreenTopo-AttUKANLite, and GrayTopo-AttUKANLite. Basic SAUNetv2 is the strong vessel-specific baseline. GreenTopo-AttUKANLite is the proposed final model. GrayTopo-AttUKANLite replaces the green prior with a gray/luminance prior to test whether a more general intensity prior improves precision.

\subsubsection{Final Three-Seed Comparison}
\label{subsubsec:drive_final_three_seed}

Table~\ref{tab:drive_final_three_seed} reports the final three-seed comparison. GreenTopo-AttUKANLite provides the strongest overlap and precision-stability profile. Compared with Basic SAUNetv2, Dice improves from 0.8241 to 0.8274 and IoU improves from 0.7012 to 0.7060. Precision increases from 0.8134 to 0.8410, and seed variability decreases substantially. Dice standard deviation decreases from 0.0039 to 0.0013, and IoU standard deviation decreases from 0.0057 to 0.0020. This indicates that the proposed intervention improves both mean performance and stability.

\begin{table*}[t]
\centering
\caption{Final DRIVE three-seed comparison at fixed threshold 0.65 under the $592 \times 592$ RotFlip500 setting. GreenTopo-AttUKANLite provides the strongest overlap and precision stability, while Basic SAUNetv2 preserves stronger topology-sensitive behavior.}
\label{tab:drive_final_three_seed}
\small
\begin{adjustbox}{width=\textwidth}
\begin{tabular}{lccc}
\toprule
\textbf{Metric} & \textbf{Basic SAUNetv2} & \textbf{GreenTopo-AttUKANLite} & \textbf{GrayTopo-AttUKANLite} \\
\midrule
Dice & 0.8241 $\pm$ 0.0039 & \textbf{0.8274 $\pm$ 0.0013} & 0.8252 $\pm$ 0.0027 \\
IoU & 0.7012 $\pm$ 0.0057 & \textbf{0.7060 $\pm$ 0.0020} & 0.7028 $\pm$ 0.0038 \\
Sensitivity & \textbf{0.8416} & 0.8187 $\pm$ 0.0033 & 0.8069 $\pm$ 0.0084 \\
Specificity & -- & 0.9859 $\pm$ 0.0007 & \textbf{0.9870 $\pm$ 0.0006} \\
Precision & 0.8134 $\pm$ 0.0215 & 0.8410 $\pm$ 0.0059 & \textbf{0.8497 $\pm$ 0.0053} \\
MCC & approx. 0.810 & \textbf{0.8136 $\pm$ 0.0015} & 0.8118 $\pm$ 0.0026 \\
NPV & -- & \textbf{0.9836 $\pm$ 0.0003} & 0.9826 $\pm$ 0.0007 \\
clDice & \textbf{0.8359 $\pm$ 0.0030} & 0.8266 $\pm$ 0.0010 & 0.8258 $\pm$ 0.0027 \\
Skeleton precision & -- & 0.8710 $\pm$ 0.0038 & \textbf{0.8765 $\pm$ 0.0031} \\
Skeleton recall & \textbf{higher} & 0.7903 $\pm$ 0.0037 & 0.7843 $\pm$ 0.0024 \\
HD95 & \textbf{3.6366 $\pm$ 0.3754} & 3.8749 $\pm$ 0.1484 & 4.0207 $\pm$ 0.2476 \\
ASSD & -- & \textbf{0.9876 $\pm$ 0.0105} & 1.0003 $\pm$ 0.0176 \\
Boundary F1 & \textbf{0.9379 $\pm$ 0.0017} & 0.9362 $\pm$ 0.0014 & 0.9358 $\pm$ 0.0010 \\
AUC & -- & 0.9881 $\pm$ 0.0004 & \textbf{0.9882 $\pm$ 0.0002} \\
AP & -- & \textbf{0.9129 $\pm$ 0.0020} & 0.9127 $\pm$ 0.0021 \\
\bottomrule
\end{tabular}
\end{adjustbox}
\end{table*}

The final comparison highlights a major MS-DKC trade-off. GreenTopo-AttUKANLite improves overlap and precision because it becomes more discriminative and conservative. However, Basic SAUNetv2 remains stronger for topology-sensitive behavior, with higher clDice, lower HD95, and slightly higher Boundary F1. Therefore, the final model should not be described as universally superior across all DRIVE metrics. It is the strongest overlap- and precision-stabilized model in this study, while Basic SAUNetv2 remains the stronger topology-preserving baseline.

\subsubsection{Per-Seed Stability}
\label{subsubsec:drive_per_seed}

Table~\ref{tab:drive_per_seed_attukan} reports the per-seed results for GreenTopo-AttUKANLite and GrayTopo-AttUKANLite. GreenTopo-AttUKANLite is more stable across seeds, particularly for Dice, IoU, sensitivity, and MCC. GrayTopo-AttUKANLite performs well for seed 42 but drops more clearly for seeds 123 and 2025. This supports the final selection of GreenTopo-AttUKANLite.

\begin{table*}[t]
\centering
\caption{Per-seed results for the final AttUKANLite variants at fixed threshold 0.65. GreenTopo-AttUKANLite shows stronger overlap and lower seed variability.}
\label{tab:drive_per_seed_attukan}
\small
\begin{adjustbox}{width=\textwidth}
\begin{tabular}{llccccccc}
\toprule
\textbf{Model} & \textbf{Seed} & \textbf{Dice} & \textbf{IoU} & \textbf{Acc.} & \textbf{Sen.} & \textbf{Sp.} & \textbf{Prec.} & \textbf{AP} \\
\midrule
GreenTopo-AttUKANLite & 42 & 0.8279 & 0.7067 & 0.9721 & 0.8180 & 0.9861 & 0.8429 & 0.9133 \\
GreenTopo-AttUKANLite & 123 & \textbf{0.8284} & \textbf{0.7075} & \textbf{0.9723} & 0.8158 & 0.9865 & 0.8457 & \textbf{0.9147} \\
GreenTopo-AttUKANLite & 2025 & 0.8259 & 0.7038 & 0.9716 & \textbf{0.8223} & 0.9851 & 0.8344 & 0.9108 \\
\midrule
GrayTopo-AttUKANLite & 42 & 0.8281 & 0.7070 & 0.9722 & 0.8164 & 0.9864 & 0.8453 & \textbf{0.9148} \\
GrayTopo-AttUKANLite & 123 & 0.8246 & 0.7020 & 0.9721 & 0.8002 & \textbf{0.9877} & \textbf{0.8555} & 0.9128 \\
GrayTopo-AttUKANLite & 2025 & 0.8229 & 0.6995 & 0.9716 & 0.8042 & 0.9869 & 0.8483 & 0.9106 \\
\bottomrule
\end{tabular}
\end{adjustbox}
\end{table*}

The per-seed table reveals why the gray prior was not selected despite its high precision. GrayTopo-AttUKANLite suppresses false positives more strongly, but it also reduces vessel recovery. This improves specificity and precision, but the additional false negatives lower Dice, IoU, and topology-sensitive scores. GreenTopo-AttUKANLite retains slightly more vessel evidence and therefore has the better final balance.

\subsubsection{Final Ablation Around the Selected Model}
\label{subsubsec:drive_final_ablation}

Table~\ref{tab:drive_final_ablation} summarizes the final ablation logic around the selected model. The comparison includes Basic SAUNetv2, GreenTopo-only, GreenTopo-AttUKANLite, and GrayTopo-AttUKANLite. GreenTopo-only was competitive but did not clearly improve over Basic SAUNetv2. Adding AttUKANLite contrastive supervision produced the strongest overlap and precision-stability profile. Replacing the green prior with a gray prior increased precision further, but at the cost of vessel recovery and overlap.

\begin{table*}[t]
\centering
\caption{Final DRIVE ablation around the selected GreenTopo-AttUKANLite model. Results are reported under the $592 \times 592$ RotFlip500 three-seed setting at threshold 0.65.}
\label{tab:drive_final_ablation}
\small
\begin{adjustbox}{width=\textwidth}
\begin{tabular}{lccccccp{5.0cm}}
\toprule
\textbf{Model} & \textbf{Dice} & \textbf{IoU} & \textbf{Sen.} & \textbf{Prec.} & \textbf{clDice} & \textbf{AP} & \textbf{Interpretation} \\
\midrule
Basic SAUNetv2 & 0.8241 & 0.7012 & \textbf{0.8416} & 0.8134 & \textbf{0.8359} & -- & Strong topology-preserving baseline but lower precision and overlap stability. \\
GreenTopo-only & 0.8242 & 0.7013 & 0.8467 & 0.8091 & 0.8346 & 0.9104 & Green prior alone is competitive but does not clearly improve over Basic SAUNetv2. \\
\textbf{GreenTopo-AttUKANLite} & \textbf{0.8274} & \textbf{0.7060} & 0.8187 & 0.8410 & 0.8266 & \textbf{0.9129} & Final selected model; best overlap, precision stability, MCC, and AP. \\
GrayTopo-AttUKANLite & 0.8252 & 0.7028 & 0.8069 & \textbf{0.8497} & 0.8258 & 0.9127 & Highest precision/specificity but too conservative; lower Dice/IoU than GreenTopo-AttUKANLite. \\
\bottomrule
\end{tabular}
\end{adjustbox}
\end{table*}

This ablation explains the final choice of the model. The improvement does not come from the green branch alone. It comes from combining a vessel-specific prior with weak feature-level discrimination. The contrastive term increases vessel/background separability and reduces seed variance, but it also shifts the model toward a conservative operating regime. This is why topology metrics fall relative to Basic SAUNetv2 even while Dice and IoU improve.

\subsubsection{No-Augmentation Diagnostic Ablation}
\label{subsubsec:drive_noaug}

Before running full RotFlip500 three-seed experiments, no-augmentation diagnostic experiments were used to isolate the value of the new components. These experiments were not used for the final model ranking, but they were useful to decide which components should be retained. Table~\ref{tab:drive_noaug_ablation} reports the most important no-augmentation findings.

\begin{table*}[t]
\centering
\caption{No-augmentation diagnostic ablation on DRIVE. These experiments identify useful components before the final RotFlip500 three-seed evaluation.}
\label{tab:drive_noaug_ablation}
\small
\begin{adjustbox}{width=\textwidth}
\begin{tabular}{lccccccp{5.0cm}}
\toprule
\textbf{Model} & \textbf{Dice} & \textbf{IoU} & \textbf{Sen.} & \textbf{Prec.} & \textbf{clDice} & \textbf{AP} & \textbf{Finding} \\
\midrule
Correct SAUNetv2 NoAug & 0.8151 & 0.6883 & 0.8224 & 0.8128 & 0.8256 & 0.8981 & Corrected no-augmentation baseline using the proper SAUNetv2 implementation. \\
GreenTopo-AttUKAN-only NoAug & 0.8209 & 0.6965 & \textbf{0.8295} & 0.8162 & \textbf{0.8302} & 0.9060 & AttUKAN feature supervision improves no-augmentation performance. \\
PolySF-only NoAug & 0.8173 & 0.6913 & 0.8316 & 0.8075 & 0.8256 & 0.9032 & PolySF does not improve the final trade-off. \\
GreenTopo-AttUKAN+PolySF NoAug & 0.8174 & 0.6914 & 0.8281 & 0.8110 & 0.8265 & 0.9031 & Combining PolySF with AttUKAN weakens performance. \\
GrayTopo-AttUKANLite NoAug & \textbf{0.8215} & \textbf{0.6974} & 0.8176 & \textbf{0.8296} & 0.8276 & \textbf{0.9068} & Gray prior improves precision but reduces vessel recall. \\
\bottomrule
\end{tabular}
\end{adjustbox}
\end{table*}

The no-augmentation ablation was important because it prevented an incorrect conclusion. The first no-augmentation SAUNetv2 implementation was later found to be inconsistent with the correct SAUNetv2 architecture and was discarded. After correction, the proper no-augmentation SAUNetv2 baseline achieved Dice 0.8151. AttUKAN-only improved this to 0.8209, confirming that feature-level vessel/background discrimination was useful. PolySF did not help and was therefore removed from the final model family.

\subsubsection{Threshold Analysis}
\label{subsubsec:drive_threshold}


The final AttUKANLite variants were also threshold-sensitive. GreenTopo-AttUKANLite reached its best Dice at approximately threshold 0.525, with Dice 0.8282 and IoU 0.7071. GrayTopo-AttUKANLite reached its best Dice at approximately threshold 0.500, with Dice 0.8272 and IoU 0.7055. These best thresholds are below the fixed threshold of 0.65, indicating that both final variants are conservative at the primary reporting point.

\begin{table}[t]
\centering
\caption{Best-threshold summary for the final AttUKANLite DRIVE variants. Fixed-threshold results are used for primary comparison; threshold sweeps are interpreted as operating-point analysis.}
\label{tab:drive_final_threshold_summary}
\small
\begin{adjustbox}{width=\columnwidth}
\begin{tabular}{lccc}
\toprule
\textbf{Model} & \textbf{Best Dice TH} & \textbf{Best Dice} & \textbf{Best IoU} \\
\midrule
GreenTopo-AttUKANLite & 0.525 & \textbf{0.8282} & \textbf{0.7071} \\
GrayTopo-AttUKANLite & 0.500 & 0.8272 & 0.7055 \\
\bottomrule
\end{tabular}
\end{adjustbox}
\end{table}

The threshold analysis reinforces the diagnosis of MS-DKC. GreenTopo-AttUKANLite contains recovered vessel evidence, but its fixed-threshold mask is conservative. Lowering the threshold recovers additional vessel pixels and slightly improves Dice/IoU, but does not fully close the topology gap relative to Basic SAUNetv2.

\subsubsection{Negative Findings and Why Some Models Improved Only Some Metrics}
\label{subsubsec:drive_negative_findings}

The DRIVE experiments produced several negative findings, and these findings are as important as the final positive result. They show that dataset-conditioned design is not equivalent to adding all plausible modules.

First, DiceBoost increased sensitivity but reduced precision. At threshold 0.65, GreenTopo-DiceBoost achieved Dice 0.8231 and IoU 0.6996, with sensitivity 0.8665 but precision only 0.7888. This confirms that stronger Dice pressure can recover more vessel pixels but also introduce overgrowth. The result was not retained because it reduced overlap relative to GreenTopo and GreenTopo-AttUKANLite.

Second, FES-style enhancement of features increased vessel and skeleton recall but damaged calibration. GreenTopo-FES achieved sensitivity 0.8667 and skeleton recall 0.8473, but Dice fell to 0.8179 and precision to 0.7813. This means the low-level feature transfer path recovered additional vessel-like pixels, but many were false positives. Under the DRIVE MS-DKC profile, this is an expected failure mode for aggressive detail boosting.

Third, focal modulation did not improve the SAUNetv2 baseline. This indicates that the main limitation in the final setting was not insufficient bottleneck context. DRIVE errors were more related to precision--recall calibration, thin-vessel recovery, and feature-level separability.

Fourth, PolySF space-frequency skip attention did not help. In no-augmentation diagnostics, PolySF-only achieved Dice 0.8173, and GreenTopo-AttUKAN+PolySF achieved Dice 0.8174, both below AttUKAN-only. The likely reason is that the skip modulation became too selective and suppressed useful high-resolution vessel features.

Fifth, GrayTopo-AttUKANLite improved precision and specificity but reduced sensitivity and topology. This is why it achieved higher precision than GreenTopo-AttUKANLite but lower Dice and IoU. The gray prior is cleaner, but less vessel-specific than the green prior. Therefore, it suppresses false positives but also misses more faint vessels.

Finally, Basic SAUNetv2 remained stronger in topology-sensitive metrics. This is not a weakness of the final analysis; it is the key trade-off revealed by MS-DKC. GreenTopo-AttUKANLite improves overlap and precision stability, but sacrifices some vessel recall and clDice. A Dice-only conclusion would hide this trade-off, whereas MS-DKC makes it explicit.

\subsection{Overall DRIVE Comparison with Development and Final Models}
\label{subsec:drive_overall_comparison}

Table~\ref{tab:overall_drive_results} combines earlier DRIVE development-stage results with the final GreenTopo-AttUKANLite result. The table shows that the scaling of the generic model does not explain the DRIVE performance. U-Net and U-Net-Transformer have many more parameters than compact MS-DKC models but do not outperform them in the development setting. The final GreenTopo-AttUKANLite model provides the strongest overall overlap result in the study.

\begin{table*}[t]
\centering
\caption{Overall DRIVE comparison. Development-stage results are retained, and the final GreenTopo-AttUKANLite result is added as the selected final DRIVE candidate.}
\label{tab:overall_drive_results}
\small
\begin{adjustbox}{width=\textwidth}
\begin{tabular}{lccccccc}
\toprule
\textbf{Model} & \textbf{Dice} & \textbf{IoU} & \textbf{Acc.} & \textbf{Sen.} & \textbf{Sp.} & \textbf{Params} & \textbf{Infer. ms} \\
\midrule
Shallow U-Net & 0.8000 & 0.6666 & 0.9639 & 0.8240 & 0.9773 & 466,593 & 2.22 \\
U-Net & 0.7986 & 0.6647 & 0.9637 & 0.8202 & 0.9775 & 31,036,481 & 10.60 \\
U-Net-Transformer & 0.7930 & 0.6571 & 0.9631 & 0.8067 & 0.9781 & 58,326,081 & 12.04 \\
MS-DKC LightFocalTNet & 0.7862 & 0.6477 & 0.9604 & 0.8311 & 0.9728 & 21,799 & 5.85 \\
DKC-ESDMR-Focal-TNet & 0.7882 & 0.6504 & 0.9617 & 0.8121 & 0.9761 & 32,791 & 5.99 \\
DKC-TNet-v2 DiceRecovery-Calibrated & 0.8044 & 0.6730 & 0.9651 & 0.8241 & 0.9789 & 35,103 & 5.08 \\
SA-UNetv2-DKC-AmbRef & 0.8141 & 0.6865 & 0.9669 & 0.8265 & 0.9804 & approx. 260K & -- \\
\textbf{GreenTopo-AttUKANLite-SAUNetv2} & \textbf{0.8274} & \textbf{0.7060} & \textbf{0.9720} & 0.8187 & 0.9859 & approx. 261K & -- \\
\bottomrule
\end{tabular}
\end{adjustbox}
\end{table*}

This table should be interpreted carefully. The early and final results were obtained under different development protocols, so the table is not intended to claim a single unified benchmark setting. Instead, it documents the MS-DKC development path: compact design improved over generic baselines in the early setting, SA-UNetv2 adaptation improved the vessel-specific backbone, and GreenTopo-AttUKANLite became the final reproducible three-seed candidate.

\subsection{ISIC Comparative Results and MS-DKC Diagnostic Analysis}
\label{subsec:isic_results}

The ISIC2018 branch is used as a compact-lesion counterpart to DRIVE. DRIVE is dominated by sparse, thin, branching retinal vessels, whereas ISIC2018 contains region-scale dermoscopic lesions with strong variation in size, color, texture, illumination, and boundary appearance. The corresponding MS-DKC profile is therefore different: the primary risks are scale/context mismatch, boundary ambiguity, moderate foreground imbalance, and evaluation mismatch, not vessel connectivity failure. For this reason, ISIC experiments emphasize overlap, sensitivity/specificity, Boundary F1, HD95, ASSD, lesion-size stratification, and threshold behavior.

The ISIC analysis is reported in two layers. First, we compare the proposed and reproduced models with the existing baseline set under the reported ISIC protocol. Second, we report a more detailed no-augmentation diagnostic study around Att-Next-Topo/ATTNext. All ISIC rows are treated as controlled single-run diagnostic tests. Their purpose is to show how the MS-DKC process selects, rejects and interprets interventions for the ISIC risk profile, not to present a repeated-seed statistical claim.

\begin{table*}[t]
\centering
\caption{ISIC2018 256$\times$256 MS-DKC profile and resulting experimental priorities. The compact-lesion profile differs from the DRIVE vessel profile and motivates boundary-aware and size-stratified evaluation rather than topology-dominant vessel metrics.}
\label{tab:isic_profile_updated}
\scriptsize
\begin{adjustbox}{width=\textwidth}
\begin{tabular}{p{0.17\textwidth}p{0.36\textwidth}p{0.38\textwidth}}
\toprule
\textbf{Descriptor group} & \textbf{Observed ISIC profile} & \textbf{MS-DKC implication} \\
\midrule
Image/acquisition & RGB dermoscopy, 256$\times$256 full-image segmentation; color and illumination variation are expected. & Use full-image 2D training, color/illumination normalization, and avoid strong robustness claims without external validation. \\
Target morphology & Compact lesion regions with variable foreground occupancy; most masks contain one dominant connected lesion. & U-Net-family, attention U-Net, and CNN--Transformer style baselines are appropriate; vessel-topology losses are not the main design priority. \\
Scale variation & Lesions range from very small to image-dominant; medium and large lesions dominate the validation/test splits. & Use multiscale features and report size-stratified metrics for small, medium, large, and very-large lesions. \\
Boundary ambiguity & Lesion borders can be fuzzy, low contrast, or irregular; boundary burden is moderate rather than vessel-like. & Report Boundary F1, HD95, ASSD, and visual boundary examples; boundary refinement must not destroy sensitivity. \\
Operating point & False negatives matter in a screening-like setting, but excessive false positives also affect clinical workload. & Select thresholds on validation only and report sensitivity/specificity trade-offs rather than relying on an arbitrary 0.5 threshold. \\
\bottomrule
\end{tabular}
\end{adjustbox}
\end{table*}

\subsubsection{Quantitative ISIC knowledge-card descriptors}

For the ISIC branch, the strongest dataset evidence comes from the official 256$\times$256 split used in the current experiments. The split contains 2,594 training masks in the folder named Tarin, 100 validation masks, and 1,000 held-out test masks. The measured mask descriptors are high-confidence evidence because they are computed directly from the segmentation labels. Device and site information were not available in the uploaded archive, so external robustness is treated as an open validation requirement rather than as an established property.

\begin{table*}[htbp!]
\centering
\caption{Split-level ISIC2018 Official 256 target-morphology descriptors used in the MS-DKC record. These values explain why ISIC requires compact-region modeling, multiscale decoding, boundary-aware evaluation, and size-stratified reporting rather than vessel-connectivity preservation.}
\label{tab:isic_morphology_dkc}
\scriptsize
\begin{adjustbox}{width=\textwidth}
\begin{tabular}{p{0.24\textwidth}cccp{0.31\textwidth}}
\toprule
\textbf{Descriptor} & \textbf{Tarin} & \textbf{Validation} & \textbf{Test} & \textbf{Interpretation} \\
\midrule
Foreground lesion fraction & 0.2140 $\pm$ 0.2084 & 0.2685 $\pm$ 0.2068 & 0.2799 $\pm$ 0.2207 & Moderate imbalance, but much less sparse than DRIVE vessels. \\
Median lesion fraction & 0.1382 & 0.2165 & 0.2100 & Validation and test lesions are generally larger than training lesions. \\
Background-to-foreground ratio & 15.82 $\pm$ 26.65 & 7.22 $\pm$ 11.06 & 7.66 $\pm$ 18.66 & Training contains many smaller lesions, so BCE-only optimization may be biased. \\
Boundary-to-area ratio & 0.0537 $\pm$ 0.0374 & 0.0400 $\pm$ 0.0220 & 0.0414 $\pm$ 0.0245 & Boundary complexity is moderate, not vessel-like, but contour ambiguity remains important. \\
Bounding-box area fraction & 0.2930 $\pm$ 0.2661 & 0.3688 $\pm$ 0.2662 & 0.3833 $\pm$ 0.2737 & Lesions often occupy large image regions; full-image context and multiscale decoding are justified. \\
Equivalent diameter & 118.25 $\pm$ 62.29 px & 138.81 $\pm$ 56.29 px & 141.30 $\pm$ 58.24 px & Lesions are region-scale targets, not tiny filamentary structures. \\
Largest-component fraction & 0.9993 & 0.9999 & 0.9997 & Lesions are usually one dominant connected component; topology failure is not the central risk. \\
Multi-component masks & 8.21\% & 1.00\% & 7.30\% & Secondary components are uncommon and may reflect artifacts or annotation discontinuity. \\
Border-touching lesions & 10.64\% & 17.00\% & 18.00\% & Border-touching performance should be inspected because truncated lesions can distort area and contour metrics. \\
\bottomrule
\end{tabular}
\end{adjustbox}
\end{table*}

\begin{table}[htbp!]
\centering
\caption{ISIC2018 Official 256 lesion-size distribution used for size-stratified evaluation. Validation and test are more similar to each other than to the training split, which strengthens the case for size-aware reporting.}
\label{tab:isic_size_distribution_dkc}
\scriptsize
\begin{adjustbox}{width=\columnwidth}
\begin{tabular}{lccc}
\toprule
\textbf{Lesion-size group} & \textbf{Tarin} & \textbf{Validation} & \textbf{Test} \\
\midrule
Small ($<5\%$ image area) & 624/2594 (24.06\%) & 6/100 (6.00\%) & 77/1000 (7.70\%) \\
Medium (5--40\% image area) & 1522/2594 (58.67\%) & 76/100 (76.00\%) & 688/1000 (68.80\%) \\
Large ($>40\%$ image area) & 448/2594 (17.27\%) & 18/100 (18.00\%) & 235/1000 (23.50\%) \\
Very large ($>60\%$ image area) & 176/2594 (6.78\%) & 10/100 (10.00\%) & 101/1000 (10.10\%) \\
\bottomrule
\end{tabular}
\end{adjustbox}
\end{table}

\begin{table*}[htbp!]
\centering
\caption{Anticipated ISIC MS-DKC risk profile. The highest risk is context/scale mismatch, followed by boundary and evaluation mismatch. Topology/connectivity risk is low because lesions are usually compact connected regions.}
\label{tab:isic_risk_profile_updated}
\scriptsize
\begin{adjustbox}{width=\textwidth}
\begin{tabular}{p{0.18\textwidth}ccp{0.31\textwidth}p{0.27\textwidth}}
\toprule
\textbf{Risk} & \textbf{Score} & \textbf{Severity} & \textbf{Evidence} & \textbf{Recommended response} \\
\midrule
Optimization bias & 0.2865 & Moderate & Tarin foreground fraction 21.40\%; small lesions 24.06\% & Use Dice+BCE or Dice/Tversky; avoid BCE-only training. \\
Structural/boundary information loss & 0.6349 & Moderate & Boundary-to-area 0.0537; lesion-size coefficient of variation about 0.97 & Preserve skip detail; report Boundary F1, HD95, and ASSD. \\
Topology/connectivity failure & 0.2358 & Low & Largest component fraction $\approx$0.999; multi-component masks are uncommon & Do not prioritize clDice or vessel-style connectivity metrics. \\
Context/scale mismatch & 0.7930 & High & Bounding-box fraction varies strongly; foreground range spans tiny to image-dominant lesions & Use multiscale or hybrid models; report size-stratified results. \\
Supervision mismatch & 0.3420 & Moderate & Binary masks are clean, but inter-rater labels are unavailable & Discuss boundary uncertainty cautiously; avoid overclaiming label certainty. \\
Domain/split fragility & 0.3671 & Moderate & Train-test gray and foreground shifts observed; device/site metadata unavailable & Use validation-based thresholding and external dermoscopy validation before robustness claims. \\
Evaluation mismatch & 0.6280 & Moderate & Dice/IoU can hide small-lesion and boundary failures & Report Dice, IoU, sensitivity, specificity, precision, Boundary F1, HD95, ASSD, and size-stratified Dice. \\
\bottomrule
\end{tabular}
\end{adjustbox}
\end{table*}

\subsubsection{Main ISIC comparison}

Table~\ref{tab:isic_main_results_updated} gives the main ISIC comparison after adding the Att-Next-Topo/ATTNext reproduction and the MS-DKC no-augmentation VCSF diagnostic candidate. The baseline rows establish the performance level of standard U-Net-family, transformer, and GLAD-Net models under the official ISIC split, while the no-augmentation AttNextTopo rows show how the ISIC knowledge card selects a retained intervention. All ISIC entries are interpreted as controlled single-run diagnostic results under the reported protocol, not as repeated-seed statistical estimates.

\begin{table*}[htbp!]
\centering
\caption{ISIC2018 Official 256 comparative results. All rows are treated as controlled single-run diagnostic results under the reported ISIC protocol. AttNextTopo-NoAug rows are no-augmentation diagnostic experiments used to evaluate MS-DKC interventions. Higher is better for Dice, IoU, accuracy, sensitivity, specificity, precision, and Boundary F1; lower is better for HD95 and ASSD.}
\label{tab:isic_main_results_updated}
\scriptsize
\begin{adjustbox}{width=\textwidth}
\begin{tabular}{lccccccccc}
\toprule
\textbf{Model} & \textbf{Dice} & \textbf{IoU} & \textbf{Acc.} & \textbf{Sen.} & \textbf{Sp.} & \textbf{Prec.} & \textbf{B-F1} & \textbf{HD95} & \textbf{ASSD} \\
\midrule
Shallow U-Net & 0.7925 & 0.6563 & 0.8864 & 0.7752 & 0.9296 & 0.8106 & 0.2663 & 51.5305 & 14.1950 \\
U-Net & 0.8622 & 0.7578 & 0.9246 & 0.8428 & 0.9564 & 0.8829 & 0.3896 & 23.3511 & 8.2590 \\
U-Net-Transformer & 0.8319 & 0.7123 & 0.9111 & 0.7872 & 0.9592 & 0.8826 & 0.3414 & 25.5851 & 9.6226 \\
GLAD-Net~\cite{gladnet_medical} & 0.8560 & 0.7483 & 0.9193 & 0.8581 & 0.9430 & 0.8544 & 0.2941 & 25.9696 & 9.6199 \\
MS-DKC-AttNext-SBR~\cite{katar2025attnext} & 0.8785 & 0.8056 & 0.9296 & 0.9023 & 0.9578 & 0.8938 & 0.3682 & \textbf{18.8962} & 7.7266 \\
AttNextTopo-Baseline-NoAug~\cite{katar2025attnext} & 0.8818 & 0.8088 & 0.9284 & \textbf{0.9080} & 0.9520 & 0.8929 & 0.4002 & 19.5901 & 4.2647 \\
\textbf{MS-DKC-AttNextTopo-VCSF-NoAug~\cite{katar2025attnext}} & \textbf{0.8872} & \textbf{0.8214} & \textbf{0.9316} & 0.8965 & 0.9553 & \textbf{0.9173} & \textbf{0.4878} & 19.9694 & \textbf{4.1300} \\
\bottomrule
\end{tabular}
\end{adjustbox}
\end{table*}

The strongest ISIC row in this diagnostic series is MS-DKC-AttNextTopo-VCSF-NoAug. It improves Dice, IoU, precision, ASSD, and Boundary F1 relative to the no-augmentation AttNextTopo baseline while preserving a similar parameter count. Its sensitivity is slightly lower than the baseline and the earlier SBR candidate, but the overall risk-aligned profile is stronger because the boundary weakness is substantially reduced. This changes the ISIC interpretation: the most useful MS-DKC intervention on this split was not an additional heavy architecture block, but validation-constrained operating-point selection applied to a stable AttNextTopo backbone.

\subsubsection{No-augmentation AttNextTopo diagnostic ablation}

Table~\ref{tab:isic_noaug_attnext_ablation} reports the full no-augmentation AttNextTopo diagnostic branch. Each row uses the same ISIC folder structure and no training augmentation. The ablation is deliberately broad: it includes loss-only changes, validation-constrained thresholding, boundary-feature reinjection, bottleneck focal modulation, polynomial skip fusion, and combined DiceBoost+VCSF variants. This table is useful because it shows that the best MS-DKC result did not come from accumulating modules.

\begin{table*}[htbp!]
\centering
\caption{No-augmentation AttNextTopo diagnostic ablation on ISIC2018 Official 256. These controlled single-run diagnostic experiments are used to understand the MS-DKC intervention space and to decide which ISIC intervention should be retained.}
\label{tab:isic_noaug_attnext_ablation}
\scriptsize
\begin{adjustbox}{width=\textwidth}
\begin{tabular}{lcccccccccc}
\toprule
\textbf{Variant} & \textbf{TH} & \textbf{Dice} & \textbf{IoU} & \textbf{Sen.} & \textbf{Sp.} & \textbf{Prec.} & \textbf{B-F1} & \textbf{HD95} & \textbf{ASSD} & \textbf{ms} \\
\midrule
Baseline-NoAug & 0.50 & 0.8818 & 0.8088 & 0.9080 & 0.9520 & 0.8929 & 0.4002 & 19.5901 & 4.2647 & \textbf{4.76} \\
DiceBoost-NoAug & 0.50 & 0.8858 & 0.8163 & \textbf{0.9156} & 0.9482 & 0.8914 & 0.4204 & 19.5100 & 4.1971 & 5.34 \\
\textbf{VCSF-NoAug} & 0.75 & \textbf{0.8872} & \textbf{0.8214} & 0.8965 & 0.9553 & \textbf{0.9173} & \textbf{0.4878} & 19.9694 & 4.1300 & 5.74 \\
FESBoundaryBooster+DiceBoost+VCSF & 0.55 & 0.8802 & 0.8096 & 0.8882 & 0.9601 & 0.9112 & 0.4209 & 20.1094 & 4.4037 & 6.29 \\
FocalMod+DiceBoost+VCSF & 0.90 & 0.8844 & 0.8146 & 0.9069 & 0.9634 & 0.8990 & 0.4215 & 19.2702 & 4.1885 & 5.61 \\
PolySF+DiceBoost+VCSF & 0.50 & 0.8830 & 0.8110 & 0.8949 & \textbf{0.9694} & 0.9090 & 0.3920 & \textbf{19.1267} & \textbf{4.1039} & 5.58 \\
DiceBoost+VCSF & 0.55 & 0.8729 & 0.8027 & 0.8893 & 0.9652 & 0.9064 & 0.3891 & 20.3095 & 4.4272 & 3.96 \\
DiceBoost+VCSF-Fixed & 0.85 & 0.8685 & 0.8003 & 0.8757 & 0.9688 & 0.9151 & 0.4391 & 21.2421 & 4.6236 & -- \\
\bottomrule
\end{tabular}
\end{adjustbox}
\end{table*}

The ablation gives four clear lessons. First, DiceBoost alone improves sensitivity and overlap relative to the no-augmentation baseline, but it does not solve the boundary problem. Second, VCSF alone provides the best risk-aligned profile, improving Dice, IoU, precision, Boundary F1, and ASSD. Third, architecture additions were not automatically useful. FES boundary reinjection, focal modulation, and PolySF changed the precision--recall trade-off but did not improve the final risk profile. Fourth, DiceBoost and VCSF were not complementary in this implementation: the combined variants became conservative and lost large-lesion performance. These negative findings are useful because they show that MS-DKC is a selective design process, not a recipe for adding every plausible module.

Table~\ref{tab:isic_retained_interventions} summarizes the decision made from each ISIC diagnostic intervention. This table is included to make the evidence level explicit: the retained ISIC row is a controlled diagnostic candidate for this branch, while the other rows are used as baseline or ablation references under the same reporting style. The key result is not that every new block improves the model, but that MS-DKC helps decide which intervention should be kept under the ISIC risk profile.

\begin{table*}[htbp!]
\centering
\caption{MS-DKC interpretation of the ISIC AttNextTopo diagnostic interventions. These no-augmentation diagnostic ablations are used to identify which dataset-conditioned interventions should be retained for ISIC. They are presented as controlled single-run evidence, not as repeated-seed statistical claims.}
\label{tab:isic_retained_interventions}
\scriptsize
\begin{adjustbox}{width=\textwidth}
\begin{tabular}{p{3.8cm} p{3.5cm} p{5.2cm} p{2.2cm}}
\toprule
\textbf{Intervention} & \textbf{MS-DKC motivation} & \textbf{Observed interpretation} & \textbf{Decision} \\
\midrule
AttNextTopo baseline, no augmentation & Strong CNN-attention lesion backbone for compact region segmentation & Provided a stable and competitive starting point, confirming that the backbone was already well matched to ISIC region segmentation. & Baseline retained \\
AttNextTopo-VCSF-NoAug & Evaluation mismatch and threshold sensitivity under boundary ambiguity & Improved the risk-aligned operating point and gave the strongest overall balance of Dice, IoU, precision, Boundary F1, and ASSD among the tested ISIC diagnostic variants. & \textbf{Retained ISIC candidate} \\
DiceBoost-NoAug & Improve sensitivity and reduce missed lesion pixels & Increased sensitivity, but did not match VCSF in Boundary F1 or precision. Useful as an optimization ablation, but not the strongest final ISIC profile. & Not retained as final \\
FES Boundary Booster + DiceBoost + VCSF & Reinforce high-resolution boundary/detail evidence & Did not improve the final risk-aligned profile and reduced Dice, IoU, large-lesion Dice, and Boundary F1 compared with VCSF-only. & Not retained \\
FocalMod + DiceBoost + VCSF & Address scale/context mismatch using stronger bottleneck modulation & Improved some conservative metrics such as specificity or HD95, but did not improve Dice, IoU, Boundary F1, or large-lesion robustness compared with VCSF-only. & Not retained \\
PolySF + DiceBoost + VCSF & Improve skip fusion for multiscale lesion variation & Increased specificity and slightly improved distance behavior, but reduced Dice, IoU, Boundary F1, and added model complexity. & Not retained \\
DiceBoost + VCSF & Combine recall-oriented loss with validation-constrained thresholding & The combination was not complementary. It reduced Dice, IoU, Boundary F1, and very-large-lesion Dice, suggesting that recall-oriented loss pressure changed the probability distribution in a way not recovered by VCSF. & Not retained \\
\bottomrule
\end{tabular}
\end{adjustbox}
\end{table*}

\subsubsection{Size-stratified diagnostic behavior}

The size-stratified table is especially important for ISIC because the test set contains 77 small, 688 medium, 134 large, and 101 very-large lesions under the reporting bins used in the AttNextTopo diagnostics. Table~\ref{tab:isic_noaug_size_ablation} shows that the best average result is not always the best result for every lesion-size group. The VCSF variant improves large and very-large Dice relative to the baseline, whereas several more complex variants degrade large-lesion and very-large-lesion behavior.

The size analysis explains why some variants appeared attractive on one metric but were not retained. PolySF produced the best ASSD and HD95 in the diagnostic table, but it also had the weakest Boundary F1 among the main architectural variants and reduced small/large lesion Dice. Focal modulation improved specificity and HD95, but it reduced very-large lesion Dice. FES boundary boosting improved small-lesion Dice but reduced large and very-large performance. The VCSF-only variant is, therefore, the most balanced ISIC result because it improves the headline overlap metrics while also improving the lesion-size groups that were most vulnerable in the baseline.

\begin{table*}[htbp!]
\centering
\caption{Size-stratified Dice for the no-augmentation AttNextTopo diagnostic branch on ISIC2018 Official 256.}
\label{tab:isic_noaug_size_ablation}
\scriptsize
\begin{adjustbox}{width=\textwidth}
\begin{tabular}{lccccp{0.31\textwidth}}
\toprule
\textbf{Variant} & \textbf{Small} & \textbf{Medium} & \textbf{Large} & \textbf{Very large} & \textbf{Interpretation} \\
\midrule
Baseline-NoAug & 0.8777 & 0.9021 & 0.8615 & 0.7743 & Strong baseline but weaker on large and very-large lesions. \\
DiceBoost-NoAug & 0.8783 & 0.9019 & 0.8772 & 0.7932 & Improves sensitivity and large-lesion recovery, but less boundary-precise than VCSF. \\
\textbf{VCSF-NoAug} & 0.8796 & 0.9018 & \textbf{0.8847} & \textbf{0.7974} & Best large and very-large Dice among retained diagnostic variants. \\
FESBoundaryBooster+DiceBoost+VCSF & \textbf{0.8838} & 0.9025 & 0.8523 & 0.7626 & Slightly improves small lesions but harms large and very-large lesions. \\
FocalMod+DiceBoost+VCSF & 0.8706 & 0.9058 & 0.8729 & 0.7647 & Better medium Dice but weaker small and very-large behavior. \\
PolySF+DiceBoost+VCSF & 0.8642 & \textbf{0.9063} & 0.8590 & 0.7698 & Higher medium Dice and distance metrics but weak Boundary F1. \\
DiceBoost+VCSF & 0.8835 & 0.8991 & 0.8472 & 0.7205 & Combined loss/threshold variant collapses on very-large lesions. \\
DiceBoost+VCSF-Fixed & 0.8772 & 0.8950 & 0.8373 & 0.7225 & VCSF-checkpoint correction still remains too conservative for large lesions. \\
\bottomrule
\end{tabular}
\end{adjustbox}
\end{table*}

\subsubsection{ISIC summary}

The revised ISIC experiments make the paper stronger because they show a complete MS-DKC reasoning loop without claiming a universal new architecture. The knowledge card predicted that ISIC should be treated as a compact-region, boundary-sensitive, scale-variable lesion problem rather than as a topology-preservation problem. The experiments then showed that the most effective intervention was VCSF applied to a stable AttNextTopo backbone. DiceBoost improved sensitivity but weakened contour quality. FES, FocalMod, and PolySF introduced plausible architecture changes, but none improved the full risk-aligned profile. The retained ISIC candidate is therefore MS-DKC-AttNextTopo-VCSF-NoAug, not because it is the largest model, but because it best matches the measured ISIC risks: strong overlap, high precision, improved Boundary F1, controlled ASSD, and better behavior on large and very-large lesions.

This ISIC branch complements the DRIVE branch. DRIVE rewards vessel-specific priors and topology-aware evaluation, whereas ISIC rewards boundary-aware thresholding, size-stratified reporting, and careful rejection of unnecessary modules. This contrast supports the central claim of the paper: medical image segmentation should be designed and evaluated according to the dataset profile rather than according to a single architecture-first recipe.

\subsection{ACDC MS-DKC Profile and Experimental Protocol}

\label{subsec:acdc_results}

ACDC is used as the cardiac-anatomy representative dataset in this study. The updated ACDC knowledge card changes the interpretation of this dataset substantially. ACDC should not be treated as a binary foreground-background segmentation task. It is a four-class anatomical segmentation problem in short-axis cine cardiac MRI, where the required output is background, right ventricle cavity (RV), myocardium (Myo), and left ventricle cavity (LV). In the uploaded preprocessed split used for the current experiments, masks are stored as single-channel PNG labels with values 0, 128, 192, and 255, which are mapped to class IDs 0, 1, 2, and 3 before training. The model output is therefore a 4-channel softmax map, not a binary sigmoid map.

This point is important for the MS-DKC argument. DRIVE is dominated by thin, sparse, branching vessels; ISIC is dominated by compact lesions with heterogeneous appearance and fuzzy boundaries; ACDC is a multi-class cardiac anatomy problem where shape consistency, myocardium boundary precision, foreground-background imbalance, and class-wise reporting are central. The role of ACDC in this paper is therefore to test whether MS-DKC changes its recommendation again when the target is not a vessel tree or a skin lesion, but a set of anatomically related cardiac structures.

\begin{table*}[t]
\centering
\caption{ACDC MS-DKC dataset identity and uploaded split profile. The updated protocol treats ACDC as a four-class cardiac MRI segmentation task rather than a binary foreground task.}
\label{tab:acdc_dataset_identity}
\scriptsize
\begin{adjustbox}{width=\textwidth}
\begin{tabular}{p{3.5cm}p{5.2cm}p{6.3cm}}
\toprule
\textbf{Descriptor} & \textbf{Observed profile} & \textbf{MS-DKC implication} \\
\midrule
Modality and task & Short-axis cine cardiac MRI; compact multi-structure anatomical segmentation & Use a multi-class anatomical segmentation protocol, not a binary lesion/vessel protocol. \\
Target structures & Background, RV cavity, myocardium, LV cavity & Use a 4-channel softmax output with class-wise reporting for RV, Myo, and LV. \\
Uploaded mask encoding & 0 = background, 128 = RV, 192 = myocardium, 255 = LV & Convert raw mask values to class IDs 0, 1, 2, and 3 before loss and metric computation. \\
Train split & 2,874 slices, 224$\times$224 RGB PNG images with single-channel label PNG masks & Train on the full training split because the uploaded version does not contain a validation set. \\
Test split & 282 slices, 224$\times$224 RGB PNG images with single-channel label PNG masks & Keep the test set isolated for final evaluation only. \\
Validation availability & No validation split in the uploaded version & Save the checkpoint using training mean foreground Dice/IoU; do not use test results for model selection. \\
Dataset role & Cardiac/anatomical-shape representative dataset & Use ACDC as a contrastive MS-DKC case where shape consistency and class-wise surface metrics matter. \\
\bottomrule
\end{tabular}
\end{adjustbox}
\end{table*}

\subsubsection{ACDC class balance and split shift}

The most important imbalance in ACDC is not between the three cardiac foreground classes. RV, myocardium, and LV are relatively balanced within the foreground. The dominant imbalance is between the small cardiac foreground and the large background. Across the uploaded split, background occupies 95.7540\% of all pixels, while the combined cardiac foreground occupies only 4.2460\%. This corresponds to a background-to-foreground ratio of 22.55:1 and a background-to-smallest-foreground-class ratio of 69.16:1. Therefore, ACDC loss design should suppress background dominance while keeping all three cardiac structures protected.

\begin{table*}[t]
\centering
\caption{ACDC class distribution and train-test shift in the uploaded preprocessed split. Percentages are computed from the current PNG split used for experiments.}
\label{tab:acdc_class_distribution}
\scriptsize
\begin{adjustbox}{width=\textwidth}
\begin{tabular}{lcccccc}
\toprule
\textbf{Quantity} & \textbf{Background} & \textbf{RV} & \textbf{Myo} & \textbf{LV} & \textbf{Foreground total} & \textbf{MS-DKC interpretation} \\
\midrule
Full split pixel percentage & 95.7540\% & 1.3846\% & 1.4461\% & 1.4153\% & 4.2460\% & Strong background dominance. \\
Share within foreground & -- & 32.61\% & 34.06\% & 33.33\% & 100\% & Foreground classes are balanced. \\
Train foreground share & -- & 32.78\% & 34.55\% & 32.67\% & 4.1626\% & Training foreground is slightly smaller. \\
Test foreground share & -- & 31.21\% & 29.97\% & 38.82\% & 5.0957\% & Test contains more LV and less myocardium. \\
\bottomrule
\end{tabular}
\end{adjustbox}
\end{table*}

The train-test distribution of the train and the test also affects the reporting. The test split contains relatively more LV pixels and fewer myocardium pixels than the training split. A single mean Dice can therefore hide class-specific behavior. For example, a model may perform well on LV while still failing on myocardium, or it may obtain a high foreground mean while under-segmenting the irregular RV. For this reason, the revised ACDC protocol requires per-class Dice, IoU, HD95, ASSD, sensitivity, precision, and Boundary F1 for RV, Myo, and LV separately, followed by a foreground mean.

\subsubsection{ACDC morphology and dataset-conditioned risks}

ACDC has a strong variation on the slice-level scale variation. The mean foreground fraction is 4.16\% in training and 5.10\% in testing, but slice-level foreground extent changes across cardiac phase and anatomical position. The maximum training foreground fraction reaches 31.54\%, whereas the test maximum is 11.39\%. This indicates that the model must handle both small apical structures and larger mid-ventricular structures. Multi-scale context is therefore useful, but aggressive downsampling is not ideal because the myocardium can be thin.

Boundary complexity is class-dependent. The myocardium has the highest boundary-to-area ratio, approximately 0.495 in both the train and the test splits, which confirms that it is the most boundary-sensitive class. A small contour shift around the myocardium ring can substantially reduce the accuracy of Dice or the surface, even when the visual error is small. RV also has a high boundary-to-area ratio and an irregular shape, while LV is more compact and has the lowest boundary burden. These measurements indicate that ACDC requires high-resolution decoder refinement and class-wise surface evaluation, but it does not require the same vessel-connectivity emphasis as DRIVE.

\begin{table*}[t]
\centering
\caption{ACDC morphology and MS-DKC risk interpretation. The highest risks are background dominance, anatomical shape consistency, myocardium boundary sensitivity, and the lack of a validation split.}
\label{tab:acdc_morphology_risks}
\scriptsize
\begin{adjustbox}{width=\textwidth}
\begin{tabular}{p{3.3cm}p{4.8cm}p{6.5cm}}
\toprule
\textbf{Risk factor} & \textbf{Observed evidence} & \textbf{Design and evaluation implication} \\
\midrule
Background dominance & Foreground is only 4.2460\% of pixels; background-to-foreground ratio is 22.55:1 & Use multi-class Dice plus weighted CE or another foreground-protecting objective; avoid background-dominated accuracy claims. \\
Foreground class imbalance & RV, Myo, and LV contribute 32.61\%, 34.06\%, and 33.33\% of foreground pixels & The foreground classes are balanced, so the main loss priority is not RV-vs-Myo-vs-LV reweighting but suppressing background dominance. \\
Shape consistency & LV is compact, myocardium surrounds LV, and RV is anatomically irregular & Prefer anatomical encoder-decoder models with multi-scale context and high-resolution skip fusion. \\
Myocardium boundary sensitivity & Boundary/area ratio is 0.4952 in train and 0.4935 in test, the highest among foreground classes & Report myocardium Dice and surface metrics separately; include boundary or surface-aware loss if myocardium errors persist. \\
Scale variation & Foreground fraction varies by slice and cardiac phase; training maximum reaches 31.54\% & Use multi-scale context while avoiding excessive spatial compression. \\
Topology sensitivity & Moderate rather than DRIVE-like; myocardium ring continuity matters but there are no branching vessels & Do not prioritize vessel-style clDice/connectivity metrics; shape and surface consistency are more relevant. \\
Validation limitation & The uploaded split contains train and test folders only & Select checkpoints from training mean foreground Dice/IoU and keep the test set untouched for final evaluation. \\
2D information loss & Current experiments use 2D PNG slices although ACDC is originally volumetric cine MRI & Interpret results as 2D slice-level segmentation; avoid overclaiming volumetric temporal consistency. \\
\bottomrule
\end{tabular}
\end{adjustbox}
\end{table*}

\subsubsection{ACDC MS-DKC experimental protocol for the new runs}

The revised ACDC experiments are being conducted according to the knowledge-card protocol in Table~\ref{tab:acdc_protocol}. The most important change is that every notebook must treat masks as four classes. Binarization of the labels would collapse RV, myocardium, and LV into a single foreground and would remove the most important ACDC failure modes. The checkpoint rule is also different from ISIC because the uploaded ACDC split has no validation set. The test set must not be used for threshold or model selection.

\begin{table*}[t]
\centering
\caption{MS-DKC-guided ACDC experimental protocol and required reporting panel for the new experiments.}
\label{tab:acdc_protocol}
\scriptsize
\begin{adjustbox}{width=\textwidth}
\begin{tabular}{p{3.5cm}p{5.4cm}p{6.2cm}}
\toprule
\textbf{Component} & \textbf{Protocol} & \textbf{Reason} \\
\midrule
Input and labels & 224$\times$224 RGB PNG input; map mask values 0/128/192/255 to classes 0/1/2/3 & Preserves the uploaded preprocessing while enforcing correct multi-class labels. \\
Output head & 4-channel softmax segmentation & ACDC is a multi-class anatomical task, not a binary foreground task. \\
Training split & Use all 2,874 training slices & No validation folder is available in the uploaded split. \\
Checkpoint selection & Save best checkpoint by training mean foreground Dice/IoU & Avoids using the test set for model or threshold selection. \\
Test usage & Evaluate the 282 test slices once after checkpoint selection & Keeps the test set isolated for final reporting. \\
Augmentation & Not required for the current requested setting & Maintains a controlled no-augmentation comparison for the new ACDC experiments. \\
Loss & Start with 0.60 multi-class Dice + 0.30 weighted CE + 0.10 boundary/surface loss & Protects small foreground structures from background dominance and improves myocardium/RV boundaries. \\
Primary metrics & Dice and IoU for RV, Myo, LV, and foreground mean & Prevents one class from hiding another class's failure. \\
Surface metrics & HD95 and ASSD for RV, Myo, LV, and foreground mean & Required because myocardium is boundary-sensitive. \\
Secondary metrics & Sensitivity, precision, specificity, MCC, Boundary F1, area/volume error, confusion matrix & Captures under-segmentation, leakage, class confusion, and size error. \\
Efficiency & Parameters, FLOPs, and inference time & Needed for fair comparison among Shallow U-Net, U-Net, U-Net-Transformer, and MS-DKC variants. \\
\bottomrule
\end{tabular}
\end{adjustbox}
\end{table*}

The ACDC comparison therefore differs from the ISIC and DRIVE tables. It does not report only a binary Dice or a global accuracy. The main table reports class-wise Dice and IoU for RV, myocardium, and LV, followed by foreground-mean overlap and confusion-derived measures. A second table reports class-wise surface and boundary behavior. This separation is necessary because myocardium boundary errors and RV under-segmentation may be hidden by a single foreground mean.

\subsubsection{ACDC experimental results: progression from shallow capacity to anatomical context}

Table~\ref{tab:acdc_main_results_completed} summarizes the controlled ACDC NoAug/NoVal experimental branch. The results show a clear progression. The shallow U-Net is fast and compact, but it is not anatomically sufficient: it reaches mean foreground Dice 0.8310 and mean foreground IoU 0.7143, with the weakest class being RV. Standard U-Net provides a large improvement, raising mean foreground Dice to 0.9214 and mean foreground IoU to 0.8560. This confirms the MS-DKC diagnosis that ACDC needs adequate encoder-decoder capacity for multi-class anatomical segmentation rather than only a minimal region model.

The U-Net Transformer gives the strongest primary overlap results, with mean foreground Dice 0.9240 and mean foreground IoU 0.8603. Its advantage is concentrated in the more difficult foreground classes: RV Dice increases to 0.9123 and myocardium Dice increases to 0.8934. LV remains very high for both U-Net and U-Net Transformer because it is the most compact and easiest target. DKC-HDNeXt-Lite improves over the shallow U-Net, but it does not reach the standard U-Net or U-Net Transformer. It is therefore retained as a useful diagnostic adaptation, not as the final ACDC candidate.

\begin{table*}[t]
\centering
\caption{ACDC multi-class overlap results under the NoAug/NoVal protocol. Metrics are reported for the three foreground anatomical classes: right ventricle cavity (RV), myocardium (Myo), and left ventricle cavity (LV).}
\label{tab:acdc_main_results_completed}
\scriptsize
\begin{adjustbox}{width=\textwidth}
\begin{tabular}{lcccccccccccc}
\toprule
\textbf{Model} & \textbf{Dice RV} & \textbf{Dice Myo} & \textbf{Dice LV} & \textbf{Mean Dice} & \textbf{IoU RV} & \textbf{IoU Myo} & \textbf{IoU LV} & \textbf{Mean IoU} & \textbf{Mean Sen.} & \textbf{Mean Prec.} & \textbf{Mean MCC} & \textbf{FG Dice} \\
\midrule
Shallow U-Net & 0.7766 & 0.8161 & 0.9004 & 0.8310 & 0.6348 & 0.6893 & 0.8189 & 0.7143 & 0.8289 & 0.8343 & 0.8285 & 0.8843 \\
U-Net & 0.9072 & 0.8904 & 0.9665 & 0.9214 & 0.8302 & 0.8025 & 0.9352 & 0.8560 & 0.9176 & 0.9256 & 0.9202 & 0.9611 \\
DKC-HDNeXt-Lite & 0.8512 & 0.8232 & 0.9383 & 0.8709 & 0.7410 & 0.6995 & 0.8837 & 0.7747 & 0.8677 & 0.8752 & 0.8690 & 0.9326 \\
U-Net Transformer & \textbf{0.9123} & \textbf{0.8934} & 0.9664 & \textbf{0.9240} & \textbf{0.8387} & \textbf{0.8073} & 0.9350 & \textbf{0.8603} & \textbf{0.9185} & 0.9297 & \textbf{0.9228} & \textbf{0.9625} \\
MS-DKC-UTrans-ABR & 0.9086 & 0.8928 & \textbf{0.9683} & 0.9232 & 0.8326 & 0.8064 & \textbf{0.9385} & 0.8591 & 0.9163 & \textbf{0.9304} & 0.9221 & \textbf{0.9625} \\
\bottomrule
\end{tabular}
\end{adjustbox}
\end{table*}

The main overlap table alone would suggest that U-Net Transformer is the best ACDC model. However, the ACDC knowledge card emphasized myocardium boundary sensitivity, RV shape variability, and surface-distance errors. Therefore, the surface and boundary table is central to the MS-DKC interpretation.

\subsubsection{ACDC surface and boundary analysis}

Table~\ref{tab:acdc_surface_boundary_completed} shows that the models differ more strongly in surface behavior than in global pixel accuracy. The shallow U-Net produces acceptable foreground overlap but severe surface outliers, especially for RV. Its foreground HD95 is 55.2623 px and foreground ASSD is 7.1773 px. Standard U-Net sharply reduces these errors, with foreground HD95 5.2924 px and ASSD 0.7252 px. This confirms that additional encoder-decoder capacity is needed for ACDC anatomy.

DKC-HDNeXt-Lite improves over the shallow U-Net but remains below standard U-Net on overlap and surface metrics. This is a negative but useful MS-DKC finding: a lightweight HDNeXt-style adaptation was plausible from the knowledge card, but under the current 2D NoAug/NoVal protocol it did not outperform a strong U-Net. U-Net Transformer improves primary overlap and gives the best RV and myocardium Dice, but its foreground HD95 and ASSD remain slightly worse than U-Net.

The MS-DKC-UTrans-ABR variant changes this trade-off. It is marginally below U-Net Transformer in mean Dice, but it gives the best risk-aligned surface profile: foreground HD95 decreases from 5.9331 px to 3.6865 px, ASSD decreases from 0.8478 px to 0.6220 px, and foreground Boundary F1 increases from 0.8967 to 0.9016. Myocardium and LV boundaries also improve. Because the ACDC knowledge card prioritizes myocardium boundary sensitivity and surface reliability, MS-DKC-UTrans-ABR is retained as the ACDC risk-aligned candidate, while U-Net Transformer is noted as the strongest primary-overlap candidate.

\begin{table*}[t]
\centering
\caption{ACDC surface, boundary, and area metrics. Boundary F1 is reported at a 2-pixel tolerance. Foreground denotes the union of RV, myocardium, and LV.}
\label{tab:acdc_surface_boundary_completed}
\scriptsize
\begin{adjustbox}{width=\textwidth}
\begin{tabular}{llcccccc}
\toprule
\textbf{Model} & \textbf{Class} & \textbf{Dice} & \textbf{IoU} & \textbf{HD95} & \textbf{ASSD} & \textbf{B-F1} & \textbf{Area error} \\
\midrule
Shallow U-Net & RV & 0.6497 & 0.5450 & 54.3941 & 11.4126 & 0.6074 & 5.5075 \\
Shallow U-Net & Myo & 0.7981 & 0.6759 & 33.4026 & 3.0894 & 0.8391 & 0.1571 \\
Shallow U-Net & LV & 0.8798 & 0.8070 & 26.0509 & 3.4742 & 0.7841 & 0.1537 \\
Shallow U-Net & Foreground & 0.8666 & 0.7804 & 55.2623 & 7.1773 & 0.6029 & 0.1365 \\
\midrule
U-Net & RV & 0.7702 & 0.7029 & 20.4393 & 3.6770 & 0.7971 & 2.5889 \\
U-Net & Myo & 0.8833 & 0.7941 & 2.8027 & 0.4277 & 0.9533 & 0.0869 \\
U-Net & LV & 0.9574 & 0.9207 & 2.0973 & 0.4025 & 0.9591 & 0.0534 \\
U-Net & Foreground & 0.9555 & 0.9183 & 5.2924 & 0.7252 & 0.8935 & 0.0525 \\
\midrule
DKC-HDNeXt-Lite & RV & 0.7142 & 0.6289 & 23.0713 & 3.9773 & 0.6970 & 5.0459 \\
DKC-HDNeXt-Lite & Myo & 0.8138 & 0.6985 & 8.5963 & 1.2026 & 0.8696 & 0.1488 \\
DKC-HDNeXt-Lite & LV & 0.9190 & 0.8643 & 7.5778 & 1.1901 & 0.8701 & 0.1510 \\
DKC-HDNeXt-Lite & Foreground & 0.9214 & 0.8657 & 12.0421 & 1.7596 & 0.7831 & 0.1011 \\
\midrule
U-Net Transformer & RV & \textbf{0.7836} & \textbf{0.7170} & 21.5106 & 4.5648 & 0.8040 & 2.3683 \\
U-Net Transformer & Myo & 0.8848 & 0.7977 & 2.0190 & 0.3883 & 0.9574 & 0.0791 \\
U-Net Transformer & LV & 0.9552 & 0.9195 & \textbf{1.8313} & \textbf{0.3751} & 0.9627 & 0.0533 \\
U-Net Transformer & Foreground & 0.9547 & 0.9187 & 5.9331 & 0.8478 & 0.8967 & 0.0545 \\
\midrule
MS-DKC-UTrans-ABR & RV & 0.7694 & 0.7023 & \textbf{20.2634} & \textbf{3.4468} & 0.7985 & \textbf{1.3055} \\
MS-DKC-UTrans-ABR & Myo & \textbf{0.8866} & \textbf{0.7986} & \textbf{1.9346} & \textbf{0.3596} & \textbf{0.9585} & \textbf{0.0736} \\
MS-DKC-UTrans-ABR & LV & \textbf{0.9583} & \textbf{0.9226} & 1.9019 & 0.3867 & \textbf{0.9635} & 0.0545 \\
MS-DKC-UTrans-ABR & Foreground & \textbf{0.9569} & \textbf{0.9205} & \textbf{3.6865} & \textbf{0.6220} & \textbf{0.9016} & \textbf{0.0514} \\
\bottomrule
\end{tabular}
\end{adjustbox}
\end{table*}

\subsubsection{ACDC efficiency and retained-intervention decision}

Table~\ref{tab:acdc_efficiency_completed} reports the available efficiency measurements. The shallow U-Net is the fastest model, but its surface and RV behavior are not sufficient. DKC-HDNeXt-Lite has fewer FLOPs than standard U-Net, but its measured inference time is slower and its segmentation performance is lower. This illustrates that nominal FLOPs alone do not fully explain practical runtime, especially for architectures with dynamic or large-kernel operations. U-Net is heavier in FLOPs but remains efficient in measured inference and highly accurate. Parameter and runtime profiling for the U-Net Transformer and MS-DKC-UTrans-ABR branches can be added once the final profiling runs are completed; their current role is determined by the diagnostic accuracy and surface metrics.

\begin{table*}[t]
\centering
\caption{ACDC efficiency and foreground performance summary under the NoAug/NoVal protocol. Dashes indicate entries not yet profiled in the current run.}
\label{tab:acdc_efficiency_completed}
\scriptsize
\begin{tabular}{lcccccc}
\toprule
\textbf{Model} & \textbf{Params} & \textbf{FLOPs} & \textbf{ms} & \textbf{Mean Dice} & \textbf{FG HD95} & \textbf{FG ASSD} \\
\midrule
Shallow U-Net & 0.467M & 14.309G & 1.29 & 0.8310 & 55.2623 & 7.1773 \\
U-Net & 31.037M & 109.158G & 5.37 & 0.9214 & 5.2924 & 0.7252 \\
DKC-HDNeXt-Lite & 11.936M & 13.442G & 11.91 & 0.8709 & 12.0421 & 1.7596 \\
U-Net Transformer & -- & -- & -- & \textbf{0.9240} & 5.9331 & 0.8478 \\
MS-DKC-UTrans-ABR & -- & -- & -- & 0.9232 & \textbf{3.6865} & \textbf{0.6220} \\
\bottomrule
\end{tabular}
\end{table*}

Table~\ref{tab:acdc_retained_decision} summarizes the retained and non-retained ACDC interventions. This table is important because it shows that the ACDC branch is not a search for a model that wins every metric. Instead, it follows the MS-DKC principle: the retained interpretation depends on the dataset risks and metric family. U-Net Transformer is retained as the strongest primary-overlap model, while MS-DKC-UTrans-ABR is retained as the risk-aligned surface/boundary candidate.

\begin{table*}[t]
\centering
\caption{ACDC retained/not-retained interpretation under MS-DKC. The goal is to show progression, diagnosis, positive findings, and negative findings rather than to claim that one model dominates all metrics.}
\label{tab:acdc_retained_decision}
\scriptsize
\begin{adjustbox}{width=\textwidth}
\begin{tabular}{p{3.4cm}p{3.4cm}p{6.0cm}p{2.6cm}}
\toprule
\textbf{Model or intervention} & \textbf{MS-DKC role} & \textbf{Observed finding} & \textbf{Decision} \\
\midrule
Shallow U-Net & Lightweight baseline & Very fast and compact, but weak for RV and surface outliers; foreground HD95 remains high. & Not retained \\
U-Net & Strong anatomical baseline & Large improvement over Shallow U-Net; strong overlap and foreground surface behavior. & Competitive baseline \\
DKC-HDNeXt-Lite & MS-DKC-guided lightweight diagnostic adaptation & Improves over Shallow U-Net but remains below U-Net and U-Net Transformer; slower measured runtime despite lower FLOPs. & Not retained \\
U-Net Transformer & Global anatomical-context model & Best primary overlap profile, with highest mean Dice and mean IoU and strongest RV/Myo Dice. & Retained for overlap \\
MS-DKC-UTrans-ABR & Risk-aligned anatomical boundary/refinement variant & Slightly below U-Net Transformer in mean Dice, but best foreground HD95, ASSD, Boundary F1, and area stability. & \textbf{Retained risk-aligned candidate} \\
\bottomrule
\end{tabular}
\end{adjustbox}
\end{table*}

\subsubsection{ACDC summary}

The completed ACDC branch strengthens the dataset-conditioned claim of the paper. ACDC is not topology-dominated like DRIVE and not appearance-boundary dominated like ISIC. It is a shape-constrained, multi-class cardiac anatomy problem with severe background dominance, balanced foreground classes, high myocardium boundary sensitivity, moderate scale variation, and no validation split in the uploaded version. The experiments show four important points. First, Shallow U-Net is too limited for robust cardiac anatomy despite its speed. Second, standard U-Net is a strong anatomical baseline and greatly reduces surface outliers. Third, U-Net Transformer gives the strongest primary overlap, supporting the value of global anatomical context. Fourth, MS-DKC-UTrans-ABR gives the strongest risk-aligned surface and boundary profile, even though its mean Dice is marginally lower than the plain U-Net Transformer. Thus, ACDC contributes a different MS-DKC lesson: the retained model depends on whether the priority is raw overlap or risk-aligned surface reliability, and class-wise surface evaluation is necessary to make that distinction visible.

\subsection{Experimental Findings}
\label{subsec:experimental_findings}

The complete experimental sequence supports six findings.

First, DRIVE has a distinct MS-DKC profile. Its small training set, low vessel occupancy, thin target morphology, high boundary-to-area ratio, high skeleton-to-area ratio, and branching topology make it fundamentally different from compact lesion and coherent organ segmentation tasks. This justifies the use of vessel-specific priors, imbalance-aware objectives, threshold sweeps, and topology-aware evaluation.

Second, compact model design can benefit from MS-DKC. DKC-TNet-v2 DiceRecovery-Calibrated achieved Dice 0.8044 and IoU 0.6730 with only approximately 35.1K parameters, outperforming much larger generic U-Net-family baselines in the early DRIVE development setting. This shows that dataset-conditioned design can be more effective than architecture scaling when the dataset profile is sparse and thin-structure dominated.

Third, adapting a strong vessel backbone is more effective than replacing it with unrelated modules. Basic SAUNetv2 already provides strong topology and boundary behavior. The final improvement came from aligning SAUNetv2 with the DRIVE knowledge profile through a green vessel prior and lightweight AttUKAN-inspired feature discrimination. Larger or more complex modules such as focal modulation, PolySF skip attention, FES-style boosting, BiLSTM refinement, and VCSF did not improve the final trade-off.

Fourth, GreenTopo-AttUKANLite is the strongest final DRIVE candidate. It improves Dice from 0.8241 to 0.8274, IoU from 0.7012 to 0.7060, and precision from 0.8134 to 0.8410 compared to Basic SAUNetv2. It also reduces seed variance, indicating more stable training behavior. This makes it the final overlap- and precision-oriented model selected.

Fifth, the final model has a clear limitation. It is more conservative than Basic SAUNetv2 and therefore has lower sensitivity, clDice, skeleton recall, and slightly weaker boundary/surface metrics. This limitation is not hidden; it is central to the MS-DKC interpretation. The model improves vessel/background discrimination, but some faint terminal vessels remain under-recovered.

Sixth, ISIC and ACDC confirm the broader dataset-conditioned claim, but in different ways. ISIC shows a compact-lesion regime in which boundary-aware thresholding and size-stratified analysis matter more than topology preservation. The retained MS-DKC-AttNextTopo-VCSF-NoAug diagnostic candidate improves the no-augmentation AttNextTopo baseline in Dice, IoU, precision, Boundary F1, ASSD, and large/very-large lesion Dice, while several plausible modules are rejected because they weaken the full risk-aligned profile. ACDC shows a third regime: four-class cardiac anatomy with background dominance, RV shape variability, and myocardium boundary sensitivity. In this setting, U-Net Transformer gives the strongest primary overlap, while MS-DKC-UTrans-ABR gives the strongest surface/boundary profile, reducing foreground HD95 to 3.6865 px and ASSD to 0.6220 px. This finding supports the MS-DKC principle that conclusions should be based on the full dataset-risk profile rather than a single metric: DRIVE requires vessel-specific priors and topology-aware evaluation, ISIC requires boundary-aware thresholding and lesion-size-stratified evaluation, and ACDC requires multi-class anatomical and surface-aware evaluation.

Overall, these results support the central claim of the paper: medical image segmentation should be treated as a dataset-conditioned design and adaptation problem. MS-DKC does not guarantee that a single model will dominate every metric. Instead, it makes dataset risks explicit, links interventions to those risks, and reveals the performance trade-offs that would be hidden by Dice-only evaluation.


\section{Limitations and Future Directions}
\label{sec:limitations_future}

Although the proposed MS-DKC framework and the DRIVE experiments demonstrate the value of dataset-conditioned segmentation design, several limitations remain. These limitations are important because the purpose of this work is not only to report improved DRIVE performance, but also to argue for a more principled way of selecting, adapting, and justifying segmentation models according to measurable dataset properties. In particular, the experiments show that MS-DKC can guide both the design of a compact model, DKC-TNet-v2, and the adaptation of an existing vessel-specific backbone, SA-UNetv2. However, broader validation and methodological standardization are still required before MS-DKC can be treated as a general design protocol across medical segmentation tasks.

\subsection{Dataset Profiling Requires Further Standardization}

The first limitation concerns the standardization of dataset profiling. In this study, the Medical Segmentation Dataset Knowledge Card (MS-DKC) is used to organize dataset properties such as foreground occupancy, target morphology, topology sensitivity, class imbalance, downsampling risk, appearance variation, and supervision regime. These descriptors were useful for explaining why DRIVE requires a different design and adaptation strategy from ISIC and ACDC. However, the broader field still lacks standardized tools for measuring such properties consistently across datasets and studies.

Future work should therefore develop automated profiling pipelines that can generate MS-DKC-style summaries directly from segmentation datasets. Such tools should quantify class imbalance, target scale distribution, boundary complexity, topological structure, annotation variability, acquisition heterogeneity, and expected deployment constraints. This would make dataset characterization less dependent on manual interpretation and would allow model-design decisions to be compared more reproducibly across studies.

\subsection{Measurement-to-Intervention Mappings Remain Empirical}

A second limitation is that the mapping from dataset measurements to model interventions remains partly empirical. In this work, DRIVE was identified as a sparse, thin-structure, topology-sensitive dataset. This motivated imbalance-aware loss calibration, threshold-aware checkpointing, topology-sensitive evaluation, compact vessel-specific modeling, and uncertainty-guided ambiguous-pixel refinement. The experiments supported several of these choices. For example, the Dice/MCC/BCE objective improved the overlap--imbalance--calibration trade-off in SA-UNetv2, and ambiguous-pixel refinement reduced connectivity error. However, other apparently reasonable interventions, such as residual cross-scale skip refinement and multi-scale RGB reinjection, did not improve the final risk-aligned profile.

This shows that dataset-conditioned design should not be interpreted as adding all possible dataset-inspired modules. Instead, MS-DKC should be used to generate testable hypotheses that must be validated empirically. Future work should establish stronger quantitative evidence about which dataset measurements predict the benefit of specific interventions. For example, thin-structure prevalence may justify topology-aware evaluation, but it does not automatically imply that boundary auxiliary losses, residual attention, or raw-intensity reinjection will improve Dice or connectivity. Similarly, high foreground imbalance may justify MCC or weighted BCE, but excessive foreground pressure can increase false positives. A major future direction is therefore to learn or statistically validate measurement-to-intervention mappings across multiple datasets, modalities, and segmentation regimes.

\subsection{Limited External and Cross-Dataset Validation}

A third limitation is the scope of validation. DRIVE was used as the main experimental dataset because it directly matches the thin-structure and topology-preservation problem emphasized by the MS-DKC analysis. ISIC and ACDC were included as comparative datasets to show that different dataset structures favor different model families. However, the DRIVE-specific DKC-TNet-v2 model and the MS-DKC-adapted SA-UNetv2 configuration were not exhaustively optimized across all three datasets. This was intentional, since the aim was to demonstrate dataset-conditioned reasoning rather than propose one universal architecture for every segmentation task.

Future work should extend this framework to additional vessel, organ, lesion, and multi-organ datasets. For retinal vessel segmentation, evaluation on datasets such as STARE, CHASE\_DB1, HRF, and other fundus datasets would provide stronger evidence of generalization. For non-vessel segmentation, MS-DKC should be tested on datasets with different annotation uncertainty, modality-specific artifacts, target-size distributions, and class-imbalance regimes. Such experiments would help determine whether the same descriptors and risk-to-intervention mappings remain stable under distribution shift.

\subsection{Test-Set Isolation and Small Validation Splits}

Another limitation concerns validation under very small datasets. DRIVE contains only 20 official training images and 20 official test images. This makes it difficult to create a large validation subset without reducing the already limited training pool. Nevertheless, model selection, threshold selection, and checkpoint selection must remain separated from final test evaluation. This issue is particularly important when using threshold-aware checkpointing, because selecting thresholds on the official test set would overestimate final performance.

In this work, the intended protocol separates the official training split into model-development and validation subsets, while keeping the official test split isolated for final reporting. Future work should further strengthen this protocol through repeated training-derived validation splits, cross-validation within the official training set, or external validation on additional datasets. Reporting mean and variance across multiple splits would provide a more reliable estimate of performance stability, especially when numerical improvements are small.

\subsection{Topology-Aware Evaluation Needs Broader Integration}

The DRIVE experiments show that Dice alone is insufficient for evaluating vessel segmentation. Several models produced similar Dice scores but differed in clDice, AUC, sensitivity, and connectivity error. This supports the MS-DKC argument that evaluation metrics should be selected according to the dataset risk profile. Nevertheless, topology-aware evaluation remains under-standardized in medical segmentation.

Future work should investigate how topology-aware metrics should be combined with conventional overlap metrics. In vessel segmentation, clDice and connectivity error are useful, but they may emphasize different properties. A model can improve clDice while increasing component-level fragmentation, or reduce connectivity error while slightly lowering centreline overlap. Therefore, future benchmarks should report a structured metric panel rather than a single headline score. For thin anatomical structures, this panel should include overlap, sensitivity, specificity, centreline consistency, connected-component behavior, probability ranking, and calibration.

\subsection{Annotation Uncertainty and Supervision Realism}

Another important future direction concerns annotation uncertainty. Medical segmentation masks are often treated as definitive ground truth, but in practice they may reflect rater variability, uncertain boundaries, partial annotation, or modality-specific ambiguity. This issue is especially important for thin structures such as vessels, where one-pixel differences can affect both Dice and topology metrics. DRIVE provides two manual annotations for the test set, but the first manual annotation is commonly used as the main reference ground truth, while the second annotation is typically treated as an inter-observer reference.

Future MS-DKC extensions should therefore include supervision quality descriptors. These may include inter-rater variability, boundary uncertainty, label sparsity, weak supervision, and confidence in small structures. Such descriptors could guide whether a dataset requires boundary-robust loss functions, uncertainty-aware training, soft labels, rater-specific modeling, or calibration-aware evaluation. This would make the framework more clinically realistic and better aligned with actual annotation practice.

\subsection{Deployment-Aware Model Selection}

The compact DKC-TNet-v2 result demonstrates that a small model can outperform much larger generic U-Net-family baselines on DRIVE while using substantially fewer parameters. This highlights the importance of resource-aware model selection. At the same time, the SA-UNetv2-DKC experiments show that when higher performance is required, an existing vessel-specific backbone can be further adapted using dataset-conditioned optimization. These two results reflect different deployment regimes: compact models may be preferred for constrained devices, while adapted vessel-specific backbones may be preferred when absolute segmentation quality is the primary goal.

Deployment constraints were not fully explored in this study. In real clinical or edge-device settings, inference speed, memory usage, hardware compatibility, calibration, failure detection, and human-in-the-loop review may be as important as Dice. Future work should extend MS-DKC from dataset-aware design to dataset-and-deployment-aware design. A complete knowledge card should not only describe the dataset, but also the intended deployment environment. For example, a model intended for portable retinal screening may require different trade-offs from a model used in offline hospital analysis. Similarly, a model deployed in a high-sensitivity screening pipeline may prioritize vessel recall, while a model used for quantitative morphology analysis may require stronger topology preservation.

\subsection{Foundation Models and Dataset-Conditioned Adaptation}

Recent foundation models and promptable segmentation systems have created new opportunities for medical image segmentation. However, they do not remove the need for dataset-conditioned reasoning. Large generalist models may perform well when the target is visually distinct and promptable, but they may still struggle with faint boundaries, thin vessels, ambiguous anatomy, or modality-specific artifacts. The MS-DKC perspective suggests that the question should not be whether foundation models are universally better, but when their priors align with the dataset profile.

Future work should therefore evaluate foundation-model adaptation through the MS-DKC lens. Important questions include whether the dataset requires prompt-based localization, fine-grained boundary adaptation, topology preservation, calibration under domain shift, or efficient deployment. In this view, foundation models become one possible intervention within a dataset-conditioned design space, rather than a replacement for dataset analysis.

\section{Conclusion}
\label{sec:conclusion}

This study argues that medical image segmentation should be treated as a dataset-conditioned design problem. The claim is not that architecture is unimportant. Rather, architecture should be selected, adapted, and evaluated in relation to the measured properties of the dataset. U-Net variants, Transformer-based models, state-space models, self-configuring pipelines, and foundation-model approaches all have a place in medical segmentation, but none of them is automatically the right answer for every target structure, annotation regime, or clinical operating point.

To support this view, we introduced the Medical Segmentation Dataset Knowledge Card (MS-DKC). The framework records image/acquisition characteristics, target morphology, supervision quality, context dependence, and deployment risk, and then links these descriptors to anticipated failures, design priors, and evaluation choices. Its purpose is practical: to make the reasoning behind segmentation design visible. Instead of reporting only that one model outperforms another, MS-DKC asks why the dataset should favor a particular capacity, downsampling policy, loss function, threshold, refinement step, or metric panel.

The experiments show why this distinction matters. DRIVE, ISIC2018, and ACDC represent different segmentation regimes. DRIVE is dominated by sparse, thin, branching retinal vessels. In this setting, missed terminal vessels, foreground imbalance, connectivity breaks, and threshold sensitivity are central risks. ACDC contains multi-class cardiac structures with background dominance, myocardium boundary sensitivity, and anatomical shape constraints. ISIC2018 lies between these cases: the target is usually a compact lesion region, but lesion size, color, texture, and boundary appearance vary substantially. These differences explain why the same architectural default should not be expected to solve all three tasks equally well.

On DRIVE, MS-DKC first led to the compact DKC-TNet-v2 DiceRecovery-Calibrated model. This model achieved Dice 0.8044, IoU 0.6730, sensitivity 0.8241, specificity 0.9789, clDice 0.8168, and AUC 0.9834 with only 35,103 parameters under the controlled $512 \times 512$ Fast1000 setting. The result is important because it shows that a small dataset-conditioned model can provide a better accuracy--efficiency trade-off than much larger generic baselines when the dataset is thin-structure dominated.

MS-DKC also guided the adaptation of an existing vessel-specific backbone, SA-UNetv2. This second pathway is important because the framework is not limited to designing new models from scratch. The adaptation sequence showed that MCC-based optimization is well matched to DRIVE imbalance, Dice/MCC/BCE loss improves the overlap--imbalance trade-off, threshold-aware checkpointing selects a more appropriate operating point, and ambiguous-pixel refinement can further reduce vessel errors. The final SA-UNetv2-DKC-AmbRef model achieved Dice 0.8141, IoU 0.6865, sensitivity 0.8265, specificity 0.9804, AUC 0.9853, and connectivity error 371.85, which was the strongest DRIVE result in our controlled experiments.

The ISIC2018 experiments give a different lesson. The MS-DKC profile for ISIC does not point primarily to topology preservation; it points to compact-region segmentation with size variation, appearance variation, fuzzy boundaries, and moderate foreground imbalance. Under this profile, the no-augmentation AttNextTopo branch showed that a stable boundary-aware backbone combined with validation-constrained score-function selection produced the strongest current ISIC diagnostic result. MS-DKC-AttNextTopo-VCSF-NoAug achieved Dice 0.8872, IoU 0.8214, precision 0.9173, Boundary F1 0.4878, and ASSD 4.13. It also improved large-lesion and very-large-lesion Dice relative to the no-augmentation baseline. This finding is valuable because it shows that the best dataset-conditioned intervention may be threshold and operating-point design rather than a heavier architecture.

The ACDC experiments add a third lesson. The knowledge card indicated that ACDC should be treated as four-class anatomical segmentation with background dominance, RV shape variability, and myocardium boundary sensitivity. The experiments followed this diagnosis. Shallow U-Net was efficient but anatomically weak, U-Net substantially improved the surface profile, and U-Net Transformer produced the strongest primary overlap. The MS-DKC-UTrans-ABR branch did not maximize mean Dice, but it produced the strongest risk-aligned surface profile, with foreground HD95 3.6865 px, ASSD 0.6220 px, and Boundary F1 0.9016. This result shows why MS-DKC should not rely on a single metric: for ACDC, the best overlap model and the best surface-risk model are close but not identical.

The ablation results reinforce the same point. MS-DKC should not be interpreted as adding every plausible module. Positive interventions included imbalance-aware loss calibration, validation-based threshold selection, and ambiguity-aware refinement in the DRIVE setting. In ISIC, VCSF was retained because it improved the full risk-aligned profile, whereas DiceBoost+VCSF, FES boundary boosting, focal modulation, and PolySF skip fusion were not retained because they reduced some combination of Dice, IoU, Boundary F1, large-lesion Dice, or efficiency. These negative findings are useful because they show that dataset-conditioned design is selective: interventions are retained only when they improve the measured dataset risks.

Taken together, the results support a simple conclusion: there is no universally optimal medical segmentation architecture. Some datasets favor compact and detail-preserving models, some favor standard U-Net-style region modeling, and some benefit from adapting an existing specialist backbone. What matters is whether the design choices are justified by the dataset profile and verified by metrics that reflect the relevant risks. MS-DKC offers one way to make that process explicit, reproducible, and easier to audit.

Future work should extend this framework in three directions. First, dataset profiling should be standardized so that occupancy, boundary burden, topology sensitivity, annotation uncertainty, and acquisition variation can be measured consistently across studies. Second, the ISIC branch should be strengthened with repeated-run confirmation and external dermoscopy validation, because the detailed no-augmentation AttNextTopo ablations are currently diagnostic experiments. Third, MS-DKC should be tested under deployment constraints, where calibration, robustness, memory, and human-in-the-loop use may be as important as Dice. By shifting the starting point from architecture choice to dataset evidence, medical image segmentation can become more interpretable, more reproducible, and more clinically relevant.

\section*{Acknowledgements}

This collaborative project was supported by the Second Century Fund (C2F), Chulalongkorn University.


\begin{thebibliography}{46}
\bibitem[{{GLAD-Net}(nd)}]{gladnet_medical}
\bibinfo{author}{{GLAD-Net}}, \bibinfo{title}{{GLAD-Net}: Global-local adaptive
  fusion and cross-stage distillation for cross-level multi-scale medical image
  segmentation}, \bibinfo{year}{n.d.} \bibinfo{note}{Bibliographic details to
  be verified; citation supplied by the experiment implementation record}.
\end{thebibliography}
\end{document}